\documentclass{article}

\PassOptionsToPackage{authoryear,round}{natbib}
\usepackage[main, final]{neurips_2026}

\usepackage[utf8]{inputenc} 
\usepackage[T1]{fontenc}    
\usepackage{hyperref}       
\usepackage{url}            
\usepackage{booktabs}       
\usepackage{amsfonts}       
\usepackage{nicefrac}       
\usepackage{microtype}      
\usepackage[table,dvipsnames]{xcolor}         
\usepackage{algorithm}
\usepackage{algorithmic}
\usepackage{amsmath}
\usepackage{amsthm}
\usepackage{multirow}
\usepackage{xspace}
\usepackage{graphicx}
\usepackage{pifont}
\usepackage{bbding}
\usepackage{array}
\usepackage{tabularx}
\usepackage{makecell}
\usepackage{wrapfig}
\usepackage[most]{tcolorbox}
\usepackage{fontawesome5}
\usepackage{placeins}
\usepackage{fvextra}
\makeatletter\g@addto@macro\UrlBreaks{\do\-}\makeatother
\newcommand{\hficon}{\raisebox{-0.2ex}{\includegraphics[height=1.05em]{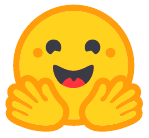}}}
\usepackage{multirow}

\newcommand{\goskills}{\textsc{GoSkills}\xspace}
\newcommand{\goskillsfull}{\textsc{Group of Skills}\xspace}
\definecolor{tablehead}{RGB}{255,255,255}
\definecolor{tableheadtext}{RGB}{0,0,0}
\definecolor{tableborder}{RGB}{0,0,0}
\definecolor{tableband}{RGB}{255,255,255}
\definecolor{methodrow}{RGB}{241,228,251}
\definecolor{vanillacell}{RGB}{214,228,248}
\definecolor{vectorcell}{RGB}{207,235,233}
\definecolor{graphcell}{RGB}{250,219,194}
\definecolor{ablhead}{RGB}{255,255,255}
\definecolor{ablheadtext}{RGB}{0,0,0}
\definecolor{ablborder}{RGB}{0,0,0}
\definecolor{ablband}{RGB}{255,255,255}
\definecolor{ablstripe}{RGB}{255,255,255}
\definecolor{ablours}{RGB}{241,228,251}
\definecolor{goodgreen}{RGB}{0,158,115}
\definecolor{badred}{RGB}{213,94,0}
\definecolor{partamber}{RGB}{190,130,18}
\definecolor{roadhead}{HTML}{FFFFFF}
\definecolor{roadband}{HTML}{FFFFFF}
\definecolor{roadstripe}{HTML}{FFFFFF}
\definecolor{roadborder}{HTML}{000000}

\newcommand{\accesssym}[1]{\makebox[0.95em][c]{#1}}
\newcommand{\cmark}{\accesssym{\textcolor{goodgreen}{\Checkmark}}}
\newcommand{\xmark}{\accesssym{\textcolor{badred}{\XSolidBrush}}}
\newcommand{\pmark}{\accesssym{{\large\textcolor{partamber}{{\ding{51}}{\small{\kern-0.7em\ding{55}}}}}}}
\newcommand{\betterup}[1]{\textcolor{goodgreen}{\scriptsize\,\(\uparrow #1\)}}
\newcommand{\betterdown}[1]{\textcolor{goodgreen}{\scriptsize\,\(\downarrow #1\)}}
\newcommand{\worseup}[1]{\textcolor{badred}{\scriptsize\,\(\uparrow #1\)}}
\newcommand{\worsedown}[1]{\textcolor{badred}{\scriptsize\,\(\downarrow #1\)}}
\newcommand{\unchanged}[1]{\textcolor{black}{\scriptsize\,\(= #1\)}}
\newcommand{\bandcell}[1]{\cellcolor{tableband}{#1}}
\newcommand{\ourscell}[1]{\cellcolor{methodrow}{#1}}
\newcommand{\tblhead}[1]{\textcolor{tableheadtext}{\textbf{#1}}}

\definecolor{appendixframe}{RGB}{61,101,124}
\definecolor{appendixcallout}{RGB}{246,250,252}
\newcommand{\apdxhdr}[1]{\textcolor{tableheadtext}{\textbf{#1}}}

\newcolumntype{L}[1]{>{\raggedright\arraybackslash}p{#1}}
\newsavebox{\apdxtabbox}
\newsavebox{\apdxcalloutbox}
\newenvironment{ApdxTabFrame}[1][\linewidth]{%
  \begingroup
  \begin{lrbox}{\apdxtabbox}%
  \begin{minipage}{#1}%
  \centering
  \arrayrulecolor{black}%
  \setlength{\heavyrulewidth}{1.1pt}%
  \setlength{\lightrulewidth}{0.5pt}%
}{%
  \end{minipage}%
  \end{lrbox}%
  \noindent\usebox{\apdxtabbox}%
  \endgroup
}

\usepackage{pdfrender}
\definecolor{contractaccent}{RGB}{91,108,214}
\definecolor{contractbg}{RGB}{247,248,254}
\definecolor{contracttitle}{RGB}{28,40,110}
\definecolor{contractchip}{RGB}{232,235,242}
\definecolor{contractslot}{RGB}{107,63,212}
\newcommand{\fieldtag}[1]{{\setlength{\fboxsep}{1.3pt}\colorbox{contractchip}{\ttfamily\footnotesize\vphantom{Ag[}\textpdfrender{TextRenderingMode=FillStroke,LineWidth=0.22pt}{#1}}}}
\newcommand{\cslot}[1]{\textcolor{contractslot}{[#1]}}
\newcommand{\cfield}[1]{\par\vspace{0.5em}\noindent\fieldtag{#1}\par\vspace{0.2em}\noindent\ignorespaces}
\newtcolorbox{contractbox}[1]{enhanced, colback=contractbg, colframe=contractaccent,
  boxrule=0.6pt, leftrule=2.6pt, arc=2.2pt, outer arc=2.2pt,
  attach boxed title to top left={xshift=4mm, yshift*=-\tcboxedtitleheight/2},
  boxed title style={colback=white, colframe=contractaccent, boxrule=0.6pt, arc=2.2pt},
  coltitle=contracttitle, fonttitle=\sffamily\bfseries\small, title={#1},
  top=3.6mm, bottom=2.4mm, left=3.2mm, right=3.2mm,
  fontupper=\ttfamily\fontsize{8pt}{11.4pt}\selectfont\raggedright\frenchspacing}
\newtcolorbox{instancebox}[1]{enhanced, colback=contractbg, colframe=contractaccent,
  boxrule=0.6pt, leftrule=2.6pt, arc=2.2pt, outer arc=2.2pt,
  attach boxed title to top left={xshift=4mm, yshift*=-\tcboxedtitleheight/2},
  boxed title style={colback=white, colframe=contractaccent, boxrule=0.6pt, arc=2.2pt},
  coltitle=contracttitle, fonttitle=\sffamily\bfseries\small, title={#1},
  top=3.6mm, bottom=2.4mm, left=3.2mm, right=3.2mm,
  fontupper=\small\raggedright}
\newtcolorbox{promptbox}[1]{enhanced, breakable, colback=appendixcallout, colframe=appendixframe,
  boxrule=0.5pt, leftrule=2.6pt, arc=2.2pt, outer arc=2.2pt,
  attach boxed title to top left={xshift=4mm, yshift*=-\tcboxedtitleheight/2},
  boxed title style={colback=white, colframe=appendixframe, boxrule=0.5pt, arc=2.2pt},
  coltitle=appendixframe, fonttitle=\sffamily\bfseries\small, title={#1},
  top=3.4mm, bottom=1.6mm, left=2.6mm, right=2.6mm, before skip=10pt, after skip=10pt}
\newtcolorbox{notebox}[1]{enhanced, colback=appendixcallout, colframe=appendixframe,
  boxrule=0.5pt, leftrule=2.6pt, arc=2.2pt, outer arc=2.2pt,
  attach boxed title to top left={xshift=4mm, yshift*=-\tcboxedtitleheight/2},
  boxed title style={colback=white, colframe=appendixframe, boxrule=0.5pt, arc=2.2pt},
  coltitle=appendixframe, fonttitle=\sffamily\bfseries\small, title={#1},
  top=3.4mm, bottom=2.2mm, left=3.2mm, right=3.2mm,
  before skip=10pt, after skip=10pt, fontupper=\small}
\newcommand{\modelicon}[1]{\raisebox{-0.22ex}{\includegraphics[height=1.12em]{#1}}}
\newcommand{\modelcell}[2]{#1\,\textbf{#2}}
\newcommand{\geminiicon}{\modelicon{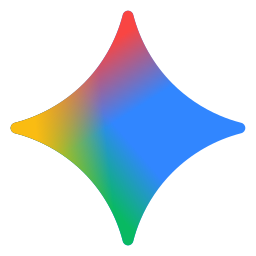}}
\newcommand{\minimaxicon}{\modelicon{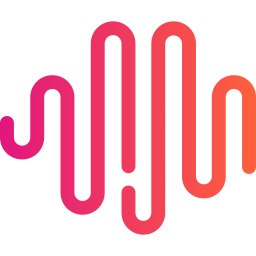}}
\newcommand{\gpticon}{\modelicon{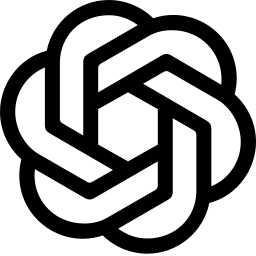}}
\newcommand{\claudeicon}{\modelicon{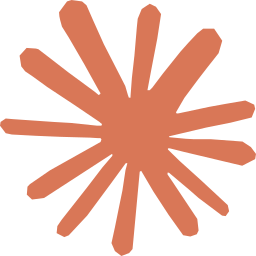}}
\newcommand{\qwenicon}{\modelicon{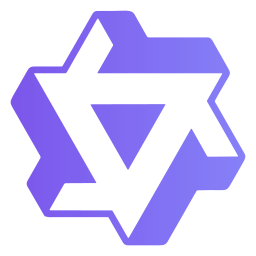}}
\newcommand{\kimiicon}{\modelicon{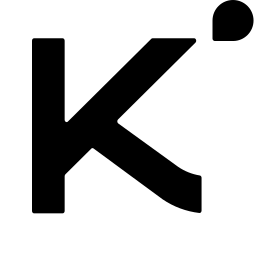}}

\title{Group of Skills: Group-Structured Skill Retrieval for Agent Skill Libraries}

\author{
  \textbf{Kun Zeng\textsuperscript{$\spadesuit$}\thanks{Equal contribution}},
  \textbf{Yu Huo\textsuperscript{$\spadesuit$}\footnotemark[1]},
  \textbf{Siyu Zhang\textsuperscript{$\heartsuit\spadesuit$}},
  \textbf{Haoyue Liu\textsuperscript{$\spadesuit$}},  \textbf{Yuecheng Zhuo\textsuperscript{$\diamondsuit$}},\\
  \textbf{Zi Ye\textsuperscript{$\clubsuit$}},
  \textbf{Yuquan Lu\textsuperscript{$\clubsuit$}},
  \textbf{Junhao Wen\textsuperscript{$\clubsuit$}},
  \textbf{Xiaoying Tang\textsuperscript{$\spadesuit$}\thanks{Corresponding author}}
  \\
  \textsuperscript{$\spadesuit$}School of Science and Engineering, The Chinese University of Hong Kong, Shenzhen\\
  \textsuperscript{$\clubsuit$}Sun Yat-sen University\quad
  \textsuperscript{$\heartsuit$}UC San Diego\quad
  \textsuperscript{$\diamondsuit$}Taiyuan University of Technology\\[0.2em]
  {\normalfont\normalsize
   \faGithub\ \href{https://github.com/yuhuo03/Group-of-Skills}{\texttt{yuhuo03/Group-of-Skills}}
   \quad
   \hficon\ \href{https://huggingface.co/datasets/yuhuo03/group-of-skills-data}{\texttt{yuhuo03/group-of-skills-data}}}
}

\begin{document}

\maketitle
\vspace{-1.5em}

\begin{abstract}
Skill-augmented agents increasingly rely on large reusable skill libraries, but retrieving
relevant skills is not the same as presenting usable context. Existing methods typically
return atomic skills or dependency-aware bundles whose internal roles remain implicit,
leaving the agent to infer the execution entry point, support skills, visible requirements,
and failure-avoidance guidance. We introduce \textbf{G}roup \textbf{o}f
\textbf{Skills} (\textsc{GoSkills}), an inference-time group-structured retrieval method
that changes the agent-facing retrieval object from a flat skill list to a compact,
role-labeled execution context. \goskills builds anchor-centered skill groups from a typed
skill graph, expands support groups through a group graph, bottlenecks the selected group
plan into a bounded set of atomic skill payloads, and renders a fixed execution contract
with \textsc{Start}, \textsc{Support}, \textsc{Check}, and \textsc{Avoid} fields, without
changing the downstream agent, skill payloads, or execution environment. Across
SkillsBench, ALFWorld, and Terminal-Bench 2 with six backbones, \goskills attains the
highest mean reward or accuracy among the evaluated methods in every benchmark--backbone
setting, and matched controls show higher reward and fewer agent turns with group
selection than with fixed-contract or post-hoc role-annotation alternatives.
\end{abstract}

\begin{figure}[H]
    \centering
    \vspace{-1.2em}
    \includegraphics[width=1\textwidth]{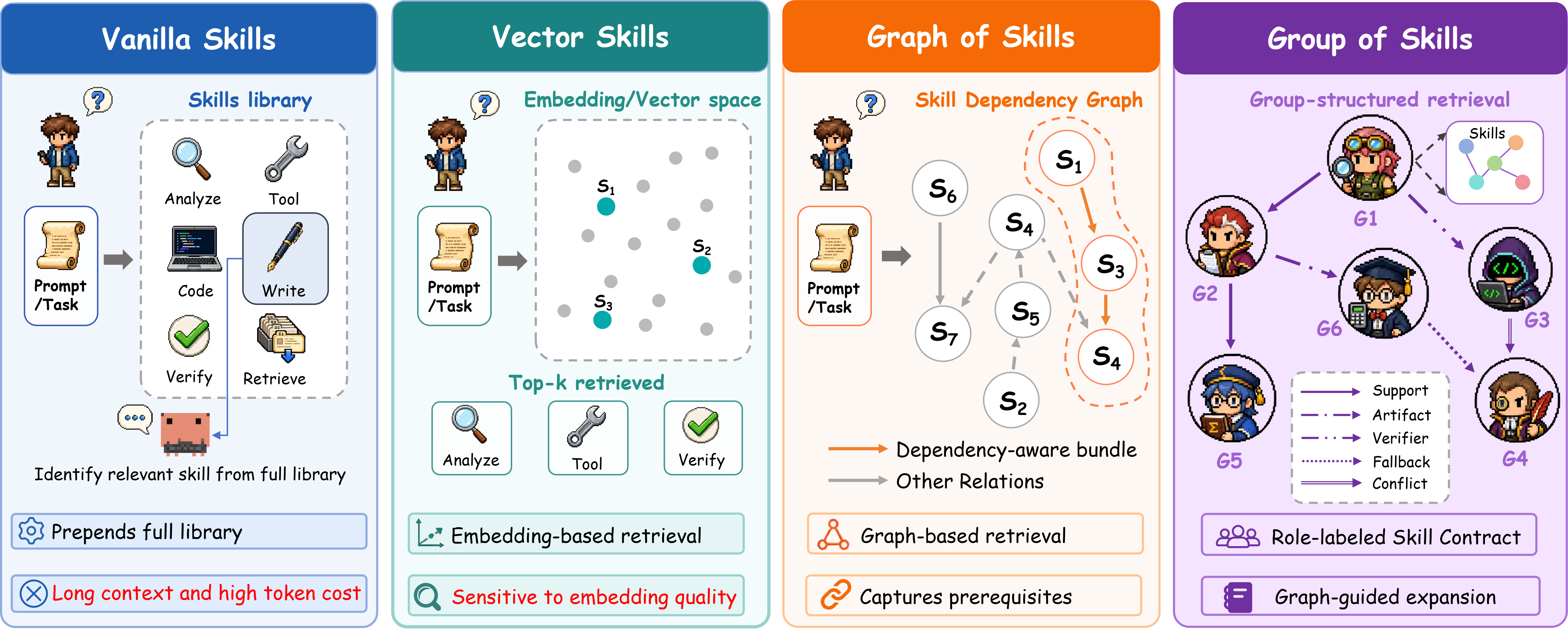}
    \vspace{-1.4em}
    \caption{
    From individual skill retrieval to group-structured retrieval: Vanilla Skills expose
    the library directly, Vector Skills retrieve top-$k$ similar skills, Graph of Skills
    hydrates dependency-aware bundles, and \goskillsfull{} scores anchor-centered groups
    and expands them into a group plan.
    }
    \label{fig:skill_retrieval_comparison}
    \vspace{-0.8em}
\end{figure}

\section{Introduction}

Skill-augmented LLM agents increasingly rely on external skill libraries: reusable
snippets, procedural templates, tool instructions, checkers, and task-specific conventions
that are too numerous to place in the prompt at once~\citep{jiang2026sok,wang2026skillx,han2025legomem}.
This reliance on reusable external knowledge aligns with broader efforts to empower
LLMs through knowledge integration at inference time~\citep{wu2025knowledge}.
As these libraries grow, the bottleneck shifts from whether an agent can access skills to
how retrieved skills should be organized under a limited context budget. Full-library prompting preserves recall but is expensive; flat semantic retrieval is
cheaper but can miss functionally required skills; and graph-based retrieval improves
recall by modeling relations among skills~\citep{li2026graph,xia2026grasp}. Yet even a relevant retrieved bundle
may still leave the downstream agent to infer the execution entry point, supporting skills,
visible requirements, and failure-avoidance guidance. Figure~\ref{fig:skill_retrieval_comparison}
summarizes this progression from individual skills and hydrated bundles toward role-aware
skill groups.

This motivates a different question: what unit should a skill retriever expose to the
agent? Existing interfaces
usually decide which atomic skills or hydrated bundles to include, while leaving the
agent-facing roles among them implicit~\citep{qu2024towards,shi2025toolret}. In
verifier-sensitive coding and interaction tasks, this missing organization can matter as
much as recall: a setup utility may need to precede a checker, a parser may only be useful
as support for a formatter, and visible requirements such as output formats or public tests
must remain explicit after context compression~\citep{NEURIPS2021_CodingChallenge,geng2023grammar,kang2025acon}.
We use ``visible requirements'' only for information available before execution, excluding
hidden tests, evaluator internals, and previous failure traces.

We introduce \emph{Group of Skills} (\goskills), an inference-time group-level retrieval and
contextualization method for agent skill libraries. The key idea is to change the
agent-facing retrieval object from a flat list of atomic skills to a compact, role-labeled
execution context. A useful group is not an arbitrary semantic cluster: it is an
anchor-centered local pattern whose support members add complementary roles, artifact
coverage, visible checks, or failure-avoidance cues. Offline, \goskills constructs such
bounded groups from typed skill neighborhoods and links related groups through a group graph.
At inference time, it retrieves an anchor group, expands support groups, bottlenecks the
selected group plan into a small set of atomic skill payloads, and renders a fixed execution
contract. Because the downstream agent, skill payloads, and execution environment are unchanged,
we isolate the intervention to retrieval-time context organization rather than
agent training, tool execution, or environment modification.

We evaluate \goskills on \textbf{SkillsBench}~\citep{li2026skillsbench},
which tests technical skill selection with visible, checkable requirements;
\textbf{ALFWorld}~\citep{shridhar2021alfworld}, which tests multi-turn
interactive decision-making; and \textbf{Terminal-Bench 2}~\citep{merrill2026terminal},
which tests transfer to terminal tasks. Together, these benchmarks examine
whether group-structured context helps agents use retrieved skills across
technical problem solving and interactive execution under constrained skill
budgets. A matched control study shows higher mean reward and fewer agent turns
with group selection than with fixed-contract or post-hoc role-annotation
controls, which is consistent with reduced organizational effort.

Our contributions are:
\begin{itemize}
    \item We formulate \emph{group-structured skill retrieval} for agent skill libraries:
    retrieval selects anchor-centered skill groups and expands them under a context budget,
    rather than exposing isolated skills or post-hoc bundles.

    \item We propose \goskills, a deterministic inference-time method that decomposes
skill-context construction into anchor selection, support expansion, and payload exposure.
This decomposition makes retrieved skills directly usable by rendering the resulting group
plan as an execution contract with coverage debt over visible requirements.

    \item We evaluate \goskills on SkillsBench, ALFWorld, and Terminal-Bench 2.
Results show higher mean reward or accuracy across all evaluated backbones, with
benchmark-dependent token and runtime tradeoffs, supported by matched controls,
task-level paired confidence intervals, and coverage diagnostics.
\end{itemize}

\section{Related Work}

\paragraph{Tool and skill retrieval.}
Tool-augmented language models, API-retrieval systems, and tool-use benchmarks show that
external tools can expand agent capabilities, while large tool or skill collections make
retrieval necessary~\citep{
schickS2023toolformer,mialon2023augmented,NEURIPS2024_Gorilla,
qin2024toolllm,shi2025toolret,huo2026reposhapley,yao2023react,zhuang2023toolqa,li2023apibank}. 
Machine memory research studies how external information can be encoded into
computable representations that support learning and reasoning~\citep{zheng2025machine}.
Prior skill repositories and benchmarks emphasize packaging, discovering, and evaluating
reusable skills across diverse agent tasks and library settings~\citep{wang2023voyager,zhang2026memskill,li2026skillsbench,liang2026skillnet,li2026organizing}. However,
retrieving relevant skills is not equivalent to presenting usable context: a retrieved
bundle may still leave the execution entry point, support roles, visible constraints, and
failure-avoidance guidance implicit. Our work targets this interface layer by using role structure during group selection
and expansion, before deciding which payloads to expose to the agent.

\paragraph{Structured retrieval.}
Graph-structured retrieval has been studied for documents, memory, and tool access, where
relations help retrieval move beyond independent nearest-neighbor matching~\citep{
edge2024local,gutierrez2024hipporag,liu2024controlllm,liu2024toolnet}.
Hypergraph computation provides a related perspective on representing higher-order
relations among multiple entities~\citep{gao2024hypergraph}. In skill settings,
such structure can recover prerequisites, setup utilities, preprocessors, and formatters
that may not be lexically salient. \goskills builds on this structural view but shifts the
decision unit from individual skills to small role-aware groups, then expands an anchor
group into support groups before bottlenecking the final payloads for the agent.

\paragraph{Concurrent work.}
Concurrent work uses structure at different stages of skill use. Graph of Skills retrieves
dependency-aware execution bundles through graph construction, seeding, diffusion,
reranking, and hydration~\citep{li2026graph}, while GRASP studies structured skill
composition and execution-time repair~\citep{xia2026grasp}. In contrast, \goskills uses
structure before execution: it performs group-level retrieval and expansion, renders a
role-labeled context, and leaves the downstream execution loop unchanged.
\section{Methodology}
\label{sec:methodology}

\begin{figure*}[t]
    \centering
    \includegraphics[width=\textwidth]{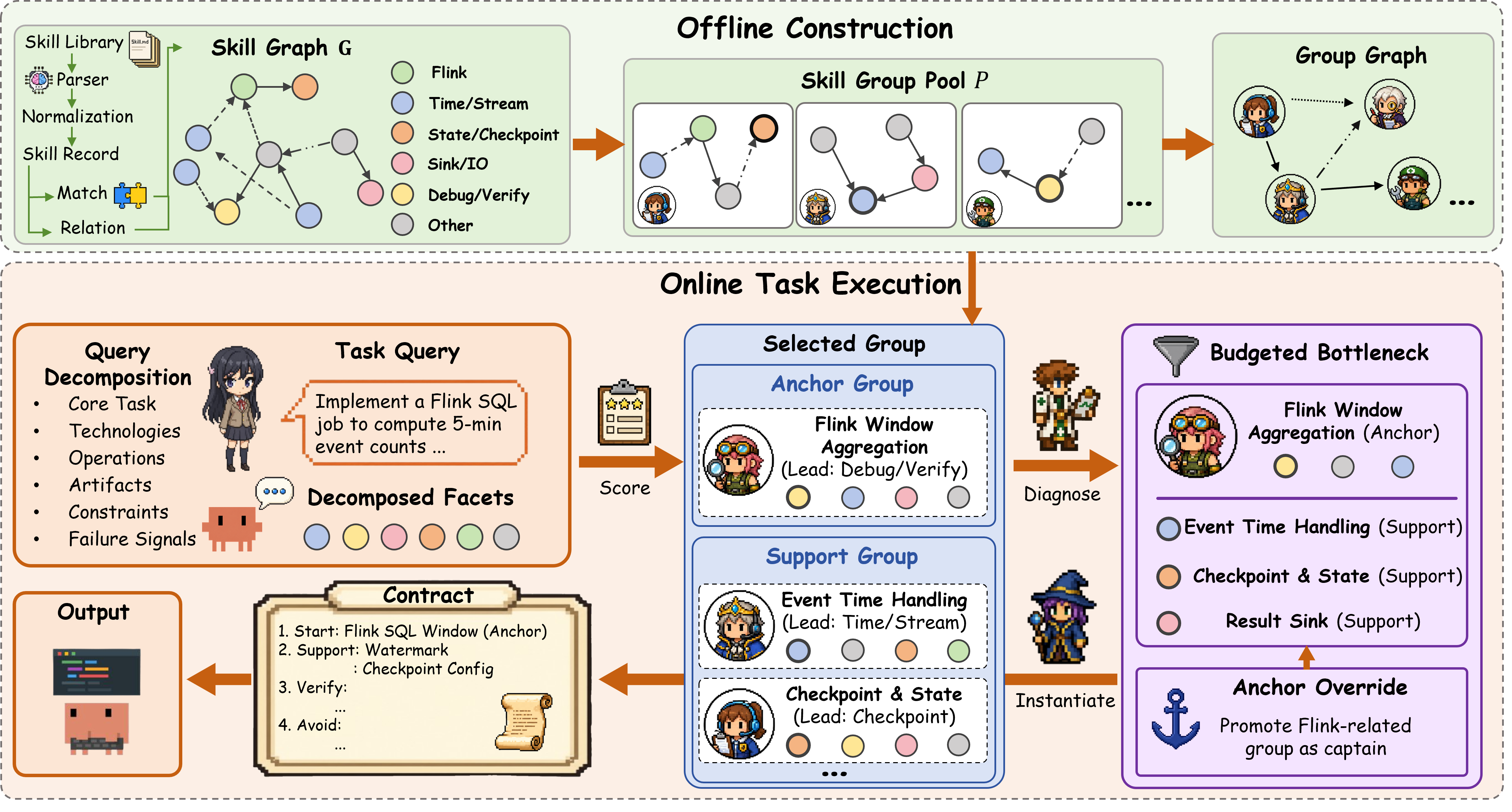}
    \caption{
    Overview of \goskills. 
    The offline stage constructs a skill graph from the skill library, extracts
    anchor-centered skill groups, and stores reusable group templates. 
    At inference time, \goskills decomposes the task query, retrieves and scores candidate
    groups, selects an anchor group, expands support groups, bottlenecks the selected group
    plan into a bounded set of atomic skill payloads, and renders a compact role-labeled
    execution contract for the downstream agent.
    }
    \label{fig:goskills_overview}
\end{figure*}

\goskillsfull{} (\goskills) is an inference-time group-level retrieval and
contextualization method for skill-augmented agents. Its goal is to change the
agent-facing retrieval object from a flat list of atomic skills to a compact, role-labeled
execution context. Unlike methods that stop after ranking or hydrating atomic skills,
\goskills uses atomic retrieval only as evidence for activating group-level retrieval units.
As shown in Figure~\ref{fig:goskills_overview}, \goskills first builds reusable skill
groups offline. At inference time, it retrieves an anchor group, expands it with support
groups, bottlenecks the selected groups into a bounded set of atomic skill payloads, and
renders them as an execution contract. Thus, any change in downstream behavior comes from
how retrieved skills are organized and exposed, not from changing the agent, skill
implementations, or environment.

We distinguish four objects. A \emph{group} is an offline reusable local retrieval unit
centered on a lead skill. A \emph{group plan} \(\mathcal{P}(q)\) is an ordered
anchor-support structure selected for a query. \(B(q)\) is the budgeted set of atomic skill
payloads presented to the agent. \(D(q,B)\) is any uncovered visible-requirement debt, and
\(C(q)\) is the deterministic contract rendered from \((\mathcal{P},B,D)\). The contribution
of \goskills is the group-level retrieval, expansion, bottlenecking, and rendering policy.

\subsection{Offline: Skill Groups and Group Graph}
\label{subsec:offline_groups}

Let \(\mathcal{S}=\{s_1,\ldots,s_n\}\) be a library of atomic skills. Each skill \(s\) has
a hydrated payload and a normalized facet set \(F_s\) extracted from its name, metadata,
tags, and text. We assume a typed skill graph
\begin{equation}
\label{eq:skill_graph}
G_{\mathcal{S}}=(\mathcal{S},E,w,\phi),
\end{equation}
where edges encode dependency, workflow, semantic, artifact, or alternative relations.

Offline, \goskills constructs a pool \(\mathcal{G}\) of small anchor-centered groups.
Each group is represented as
\begin{equation}
\label{eq:group_tuple}
g=
\left\langle
s_g^{\mathrm{lead}},M_g,R_g,F_g^+,F_g^{\mathrm{opt}},F_g^-,
A_g,V_g,T_g,\pi_g
\right\rangle ,
\end{equation}
where \(s_g^{\mathrm{lead}}\) is the lead skill, \(M_g\) are at most two support members,
\(R_g\) assigns member roles, \(F_g^+\), \(F_g^{\mathrm{opt}}\), and \(F_g^-\) encode
required, optional, and negative applicability facets, \(A_g\) and \(V_g\) store artifact
and visible-requirement cues, \(T_g\) records the local topology, and \(\pi_g\) is a fixed
group prior. A group is not an executable hidden program; it is a bounded role template
whose members may later be expanded, pruned, and rendered under a prompt budget.

\goskills also builds a typed group graph
\begin{equation}
\label{eq:group_graph}
\mathcal{H}=(\mathcal{G},E_{\mathcal{G}},\rho_{\mathcal{G}},\omega),
\end{equation}
where \(\rho_{\mathcal{G}}\) labels each group edge as support, artifact,
visible-check, fallback, or incompatibility evidence, and \(\omega\) assigns edge
weights. During online expansion, \goskills follows only positive
non-incompatibility edges:
\begin{equation}
\label{eq:positive_group_neighborhood}
\mathcal{N}_{\mathcal{H}}^{+}(\mathcal{P},q)
=
\left\{
g'\notin \mathcal{P}_{\mathrm{set}}:
\exists g\in\mathcal{P}_{\mathrm{set}},
(g,g')\in E_{\mathcal{G}},
\rho_{\mathcal{G}}(g,g')\neq \mathit{incompat},
\omega(g,g')>0
\right\}.
\end{equation}
This graph lifts atomic skill relations to reusable group-level relations. Instead of
repeatedly rediscovering support skills from the atomic graph at inference time, \goskills
can expand from an anchor group to nearby support groups that already encode role, artifact,
and visible-requirement structure. In this way, the group graph turns low-level skill
connectivity into a reusable retrieval substrate for agent-facing context organization.
Construction details are given in Appendix~\ref{app:offline_algorithm}.

\subsection{Online: Group Retrieval and Contextualized Exposure}
\label{subsec:online_contextualization}

Given a query \(q\), \goskills produces
\begin{equation}
\label{eq:online_output}
\Omega(q)=\big(\mathcal{P}(q),B(q),D(q,B),C(q)\big).
\end{equation}
The group plan is an ordered anchor-support object:
\begin{equation}
\label{eq:group_plan}
\begin{aligned}
\mathcal{P}(q)
&= \langle g_a(q),\mathcal{P}_{\mathrm{sup}}(q)\rangle,\\
\mathcal{P}_{\mathrm{set}}(q)
&= \{g_a(q)\}\cup\mathcal{P}_{\mathrm{sup}}(q).
\end{aligned}
\end{equation}
\(B(q)\subseteq\mathcal{S}\) is the presented atomic skill set, \(D(q,B)\) is remaining
coverage debt, and \(C(q)\) is the rendered execution contract. The plan is constrained by a
context budget \(\tau\); in our implementation, payload count is the primary budget and
character-level estimates act as a guard.

\paragraph{Query schema.}
The query is mapped to a deterministic schema
\begin{equation}
\label{eq:query_schema}
\psi(q)=
\left\langle
F_{\mathrm{core}},F_{\mathrm{tech}},F_{\mathrm{op}},
F_{\mathrm{artifact}},F_{\mathrm{constraint}},F_{\mathrm{failure}},F_{\mathrm{check}}
\right\rangle .
\end{equation}
Let
\begin{equation}
\label{eq:group_skill_set}
\mathcal{S}_g=\{s_g^{\mathrm{lead}}\}\cup M_g .
\end{equation}
The schema summarizes task goals, technology anchors, operations, artifacts, constraints,
failure cues, and visible-requirement semantics. Visible requirements are task properties
available before execution, such as exact output formats, required artifacts, deterministic
behavior, unit tests, or formal proof obligations. In the released implementation, schema
extraction uses the retrieval query (written by the agent or taken from the task instruction;
Appendix~\ref{app:query_formation}) and explicitly supplied public context. Skill metadata and
text provide separate retrieval features. Hidden tests and evaluator internals are outside
the query schema.

\paragraph{Group activation and scoring.}
Atomic retrieval evidence may be used, but it is not the final retrieved object. Let
\begin{equation}
\label{eq:seed_skills}
R_0(q)=\{s_{(1)},\ldots,s_{(M)}\}
\end{equation}
be an optional seed set returned by a vector or graph retriever. Candidate groups are
activated by direct query--group matches and by overlap with seed skills:
\begin{equation}
\label{eq:candidate_groups}
\mathcal{G}_q=
\textsc{DirectGroupMatches}(\psi(q),\mathcal{G})
\cup
\{g\in\mathcal{G}:\mathcal{S}_g\cap R_0(q)\neq\emptyset\}.
\end{equation}
Thus \(R_0(q)\) serves as evidence for group activation, while the final retrieval decision
is made over groups rather than individual skills.

The selection policy is fixed and deterministic rather than learned. It uses three linear
scores:
\begin{equation}
\label{eq:scoring_functions}
\begin{aligned}
U_{\mathrm{grp}}(g,q)
&= \boldsymbol{\beta}^{\top}\mathbf{x}(g,q)+\lambda_\pi\pi_g,\\
U_{\mathrm{sup}}(g\mid\mathcal{P},q)
&= \boldsymbol{\eta}^{\top}\mathbf{z}(g,\mathcal{P},q),\\
U_{\mathrm{bot}}(s\mid B,\mathcal{P},q)
&= \boldsymbol{\gamma}^{\top}\mathbf{h}(s,B,\mathcal{P},q).
\end{aligned}
\end{equation}
where \(U_{\mathrm{grp}}\) ranks candidate groups, \(U_{\mathrm{sup}}\) adds marginal support
groups, and \(U_{\mathrm{bot}}\) selects final atomic skills for presentation. The feature
families include retriever relevance, facet coverage, anchor match, visible-check support,
graph connectivity, redundancy, negative applicability, and cost. These feature
families define the scoring policy; Appendix~\ref{app:scoring_rules} specifies
the coefficients and rules of the released implementation.

\paragraph{Anchor and support-group expansion.}
Candidate groups are ranked by \(U_{\mathrm{grp}}\), and the top \(L(q)\) groups are retained
in a shortlist \(\widehat{\mathcal{G}}_q\), where \(L(q)\) is a capped size determined by query
complexity and retrieval ambiguity. Among admissible shortlisted groups
(Appendix~\ref{app:scoring_rules}), the anchor group is selected by
\begin{equation}
\label{eq:anchor_selection}
g_a=
\arg\max_{g\in\widehat{\mathcal{G}}_q}
\left[
U_{\mathrm{grp}}(g,q)+\lambda_a\mathrm{Anchor}(g,q)
\right],
\end{equation}
where \(\mathrm{Anchor}(g,q)\) promotes lead skills matching explicit technology or artifact
anchors and suppresses generic or incompatible leads. If no group is admissible, the plan is
empty; otherwise, starting from
\(\mathcal{P}_0=\langle g_a,\emptyset\rangle\), \goskills greedily adds support groups from
the retained shortlist and the group-graph neighborhood:
\begin{equation}
\label{eq:support_selection}
g^\star=
\arg\max_{
g\in
\left(
\widehat{\mathcal{G}}_q
\cup
\mathcal{N}_{\mathcal{H}}^{+}(\mathcal{P},q)
\right)
\setminus
\mathcal{P}_{\mathrm{set}}
}
U_{\mathrm{sup}}(g\mid\mathcal{P},q).
\end{equation}
Support expansion stops when the marginal gain, group budget, or context guard is
exhausted. The implementation caps the selected group plan at three groups.

\paragraph{Bottlenecking and coverage-safe backfill.}
The selected group plan is internal. The downstream agent receives only \(B(q)\) and the
contract. Let
\begin{equation}
\label{eq:plan_skill_union}
\mathcal{S}_{\mathcal{P}}=
\bigcup_{g\in\mathcal{P}_{\mathrm{set}}}\mathcal{S}_g .
\end{equation}
\goskills inserts lead skills first and then selects remaining presented skills with
\(U_{\mathrm{bot}}\), subject to \(\mathrm{cost}(B)\leq\tau\). This bottleneck uses group
structure for planning without exposing all group members.

To avoid losing explicit requirements, \goskills computes coverage debt
\begin{equation}
\label{eq:coverage_debt}
D(q,B)=
\mathcal{F}_{\mathrm{high}}(q)
\setminus
\bigcup_{s\in B}F_s^{\mathrm{vis}}(q),
\end{equation}
where \(\mathcal{F}_{\mathrm{high}}(q)\) denotes high-confidence query facets such as framework
names, file extensions, named APIs, output formats, and explicit constraints, and
\(F_s^{\mathrm{vis}}(q)\) contains facets of the presented skill name and body,
excluding hidden metadata and omitted text. Backfill uses only eligible activated skills that cover
current debt, pass negative-applicability checks, and fit the remaining budget.
Remaining debt is reported; accounting details are in Appendix~\ref{app:coverage_debt}.

\paragraph{Execution contract.}
The final output is a deterministic execution contract
\begin{equation}
\label{eq:execution_contract}
C(q)=
\left\langle
C_{\mathrm{start}},
C_{\mathrm{support}},
C_{\mathrm{check}},
C_{\mathrm{avoid}},
B(q),
D(q,B)
\right\rangle .
\end{equation}
The contract names the anchor skill, labels support skills, lists visible requirements,
states negative guidance, and includes the hydrated payloads for the presented atomic
skills. The template is fixed across tasks; task-specific content enters only through the
selected groups, presented skills, query schema, and coverage debt. \goskills does not bind
arguments, check preconditions, execute skills, or perform runtime graph repair.

\paragraph{Cost.}
Offline group construction runs once per skill library. Online, group activation uses
inverted member and facet indexes under bounded selection and payload budgets; total cost also
includes seed retrieval, which scans library documents per query for the default lexical seeder
or inherits an external retriever's cost (complexity terms in Appendix~\ref{app:complexity_details}).

\section{Experiments}
\label{sec:experiments}

\subsection{Setup}

We evaluate \goskills on \textbf{SkillsBench}, \textbf{ALFWorld} and \textbf{Terminal-Bench 2}, covering both technical skill selection
with deterministic checks and multi-turn interactive task execution. These benchmarks test
whether group-structured retrieval improves agent-facing skill use under a constrained
context budget. Within each benchmark--backbone comparison, methods share the
agent loop, environment, and source skill payloads. The main comparison uses
2,000-skill libraries for all three benchmarks. Retrieved-skill methods share
the exposure budget. Appendix~\ref{app:implementation_settings} specifies the settings.

\paragraph{Baselines.}
We compare against four skill-access settings: \emph{No Skills}, which provides no skill
context; \emph{Vanilla Skills}~\citep{agentskills2026}, which prepends task-provided
skills on SkillsBench and exposes the full library on ALFWorld and Terminal-Bench 2; \emph{Vector Skills}~\citep{openai_text_embedding_3_large}, which retrieves a flat
semantic top-$k$ list; and \emph{Graph of Skills}~\citep{li2026graph}, which hydrates
dependency-aware bundles from a typed skill graph. In contrast, \goskills retrieves
anchor-centered groups, expands support groups through a group graph, and renders the
selected context as a role-labeled execution contract. Details are shown in Appendix~\ref{app:baseline_spec}.

\paragraph{Models.}
We evaluate six backbones: GPT-5.4~\citep{openai2026gpt54}, Claude Sonnet 4.5~\citep{anthropic2025claudesonnet45}, Gemini 3 Pro~\citep{google2025gemini3}, MiniMax M2.7~\citep{minimax2026m27}, Qwen 3.5~\citep{qwen2026qwen35}, and Kimi K2.5~\citep{moonshot2026kimik25}. Each model--method--benchmark combination is run three times, and Table~\ref{tab:main_results} reports the mean over the three runs. Reward is averaged over tasks within each run and then averaged across runs. Reward is the primary outcome. Token usage reports mean cumulative input tokens over the agent loop, and runtime reports mean 
\begin{table*}[h]
\centering
\caption{
Aggregate downstream results by benchmark, method, and agent backbone.
\textbf{Reward} (SkillsBench, ALFWorld) and \textbf{Accuracy} (Terminal-Bench 2) are averaged
over tasks and reported in \%, \textbf{Tokens} denotes mean cumulative input tokens, and
\textbf{Runtime} denotes agent-only runtime (s).
For \textbf{Reward}/\textbf{Accuracy}, larger is better; for \textbf{Tokens} and \textbf{Runtime}, smaller is better.
Values are averaged over three repeats (sample SDs: Appendix~\ref{app:repeat_sd}). Symbol columns indicate skill-library access
(\textbf{Skill}), bounded retrieval (\textbf{Ret.}), and group-structured context
(\textbf{Group}); \pmark indicates dependency-graph structure without group roles.
Colored arrow deltas in skill-access rows are relative to No Skills under the same
benchmark and backbone; green/red indicates improvement/regression under the metric
direction. The best value for each metric within each benchmark--model block is
highlighted in \textbf{bold} and the second best is \underline{underlined}; No Skills is a
no-context reference and participates in metric highlighting.
}
\vspace{0.25em}
\label{tab:main_results}
\setlength{\tabcolsep}{4.1pt}
\renewcommand{\arraystretch}{1.12}
\arrayrulecolor{tableborder}

\resizebox{\linewidth}{!}{%
\begin{tabular}{ll|ccc|rrr|rrr|rrr}
\Xhline{1.1pt}

\rowcolor{tablehead}
& &
\multicolumn{3}{c|}{\tblhead{Access}} &
\multicolumn{3}{c|}{\tblhead{SkillsBench}} &
\multicolumn{3}{c|}{\tblhead{ALFWorld}} &
\multicolumn{3}{c}{\tblhead{Terminal-Bench 2}} \\

\rowcolor{tablehead}
\multicolumn{1}{c}{\tblhead{Model}} &
\multicolumn{1}{c|}{\tblhead{Method}} &
\tblhead{Skill} &
\tblhead{Ret.} &
\tblhead{Group} &
\tblhead{Reward $\uparrow$} &
\tblhead{Tokens $\downarrow$} &
\tblhead{Runtime $\downarrow$} &
\tblhead{Reward $\uparrow$} &
\tblhead{Tokens $\downarrow$} &
\tblhead{Runtime $\downarrow$} &
\tblhead{Accuracy $\uparrow$} &
\tblhead{Tokens $\downarrow$} &
\tblhead{Runtime $\downarrow$} \\
\Xhline{1.1pt}

\multirow{5}{*}{\modelcell{\gpticon}{GPT-5.4}}
 & No Skills & \xmark & \xmark & \xmark & 25.3 & \textbf{116,107} & 188.6 & 83.6 & \textbf{18,521} & \underline{35.3} & \underline{52.8} & \textbf{164,039} & \textbf{301.1} \\
 & \cellcolor{vanillacell}Vanilla Skills & \cmark & \xmark & \xmark & 28.9\betterup{3.6} & \underline{121,275}\worseup{5.2k} & \textbf{159.2}\betterdown{29.4} & \underline{92.5}\betterup{8.9} & 2,953,003\worseup{2934.5k} & 74.4\worseup{39.1} & 44.9\worsedown{7.9} & 2,071,755\worseup{1907.7k} & 445.7\worseup{144.6} \\
 & \cellcolor{vectorcell}Vector Skills & \cmark & \cmark & \xmark & 34.0\betterup{8.7} & 144,827\worseup{28.7k} & \underline{185.8}\betterdown{2.8} & \underline{92.5}\betterup{8.9} & 38,263\worseup{19.7k} & 38.5\worseup{3.2} & 48.3\worsedown{4.5} & 177,525\worseup{13.5k} & 357.7\worseup{56.6} \\
 & \cellcolor{graphcell}Graph of Skills & \bandcell{\cmark} & \bandcell{\cmark} & \bandcell{\pmark} & \bandcell{\underline{35.7}\betterup{10.4}} & \bandcell{136,054\worseup{19.9k}} & \bandcell{192.9\worseup{4.3}} & \bandcell{91.8\betterup{8.2}} & \bandcell{39,013\worseup{20.5k}} & \bandcell{40.2\worseup{4.9}} & \bandcell{46.1\worsedown{6.7}} & \bandcell{\underline{168,981}\worseup{4.9k}} & \bandcell{\underline{302.7}\worseup{1.6}} \\
 & \ourscell{\goskills} & \ourscell{\cmark} & \ourscell{\cmark} & \ourscell{\cmark} & \ourscell{\textbf{39.4}\betterup{14.1}} & \ourscell{156,696\worseup{40.6k}} & \ourscell{187.3\betterdown{1.3}} & \ourscell{\textbf{95.5}\betterup{11.9}} & \ourscell{\underline{35,042}\worseup{16.5k}} & \ourscell{\textbf{34.5}\betterdown{0.8}} & \ourscell{\textbf{60.7}\betterup{7.9}} & \ourscell{173,228\worseup{9.2k}} & \ourscell{427.6\worseup{126.5}} \\
\midrule
\multirow{5}{*}{\modelcell{\geminiicon}{Gemini 3 Pro}}
 & No Skills & \xmark & \xmark & \xmark & 20.1 & \textbf{138,914} & 161.7 & 82.1 & \textbf{19,447} & \textbf{31.1} & \underline{49.4} & \textbf{263,918} & \textbf{236.4} \\
 & \cellcolor{vanillacell}Vanilla Skills & \cmark & \xmark & \xmark & 22.4\betterup{2.3} & \underline{148,217}\worseup{9.3k} & \textbf{127.9}\betterdown{33.8} & 88.3\betterup{6.2} & 2,964,318\worseup{2944.9k} & 58.7\worseup{27.6} & 40.1\worsedown{9.3} & 1,934,487\worseup{1670.6k} & 370.8\worseup{134.4} \\
 & \cellcolor{vectorcell}Vector Skills & \cmark & \cmark & \xmark & 27.9\betterup{7.8} & 165,931\worseup{27.0k} & 168.4\worseup{6.7} & \underline{91.8}\betterup{9.7} & 43,016\worseup{23.6k} & 36.6\worseup{5.5} & 46.4\worsedown{3.0} & 300,841\worseup{36.9k} & \underline{257.3}\worseup{20.9} \\
 & \cellcolor{graphcell}Graph of Skills & \bandcell{\cmark} & \bandcell{\cmark} & \bandcell{\pmark} & \bandcell{\underline{30.6}\betterup{10.5}} & \bandcell{167,643\worseup{28.7k}} & \bandcell{174.1\worseup{12.4}} & \bandcell{91.0\betterup{8.9}} & \bandcell{37,491\worseup{18.0k}} & \bandcell{33.2\worseup{2.1}} & \bandcell{42.3\worsedown{7.1}} & \bandcell{\underline{279,723}\worseup{15.8k}} & \bandcell{265.1\worseup{28.7}} \\
 & \ourscell{\goskills} & \ourscell{\cmark} & \ourscell{\cmark} & \ourscell{\cmark} & \ourscell{\textbf{33.0}\betterup{12.9}} & \ourscell{178,762\worseup{39.8k}} & \ourscell{\underline{155.6}\betterdown{6.1}} & \ourscell{\textbf{94.8}\betterup{12.7}} & \ourscell{\underline{35,829}\worseup{16.4k}} & \ourscell{\underline{31.4}\worseup{0.3}} & \ourscell{\textbf{54.3}\betterup{4.9}} & \ourscell{292,977\worseup{29.1k}} & \ourscell{314.9\worseup{78.5}} \\
\midrule
\multirow{5}{*}{\modelcell{\minimaxicon}{MiniMax M2.7}}
 & No Skills & \xmark & \xmark & \xmark & 12.8 & \underline{404,872} & 418.3 & 41.5 & \textbf{44,587} & 74.3 & \underline{42.3} & \underline{741,520} & \underline{453.6} \\
 & \cellcolor{vanillacell}Vanilla Skills & \cmark & \xmark & \xmark & 14.2\betterup{1.4} & \textbf{401,377}\betterdown{3.5k} & \textbf{381.3}\betterdown{37.0} & 44.0\betterup{2.5} & 4,297,163\worseup{4252.6k} & 97.6\worseup{23.3} & 35.2\worsedown{7.1} & 3,013,804\worseup{2272.3k} & 608.3\worseup{154.7} \\
 & \cellcolor{vectorcell}Vector Skills & \cmark & \cmark & \xmark & 15.9\betterup{3.1} & 529,146\worseup{124.3k} & 442.6\worseup{24.3} & 50.5\betterup{9.0} & 67,814\worseup{23.2k} & \underline{68.7}\betterdown{5.6} & 36.3\worsedown{6.0} & 763,739\worseup{22.2k} & 562.1\worseup{108.5} \\
 & \cellcolor{graphcell}Graph of Skills & \bandcell{\cmark} & \bandcell{\cmark} & \bandcell{\pmark} & \bandcell{\underline{18.7}\betterup{5.9}} & \bandcell{462,039\worseup{57.2k}} & \bandcell{\underline{406.3}\betterdown{12.0}} & \bandcell{\underline{54.5}\betterup{13.0}} & \bandcell{\underline{61,903}\worseup{17.3k}} & \bandcell{73.1\betterdown{1.2}} & \bandcell{38.6\worsedown{3.7}} & \bandcell{\textbf{711,886}\betterdown{29.6k}} & \bandcell{\textbf{422.3}\betterdown{31.3}} \\
 & \ourscell{\goskills} & \ourscell{\cmark} & \ourscell{\cmark} & \ourscell{\cmark} & \ourscell{\textbf{20.3}\betterup{7.5}} & \ourscell{573,284\worseup{168.4k}} & \ourscell{438.7\worseup{20.4}} & \ourscell{\textbf{56.2}\betterup{14.7}} & \ourscell{65,587\worseup{21.0k}} & \ourscell{\textbf{62.9}\betterdown{11.4}} & \ourscell{\textbf{45.3}\betterup{3.0}} & \ourscell{830,471\worseup{89.0k}} & \ourscell{540.2\worseup{86.6}} \\
\midrule
\multirow{5}{*}{\modelcell{\claudeicon}{Claude Sonnet 4.5}}
 & No Skills & \xmark & \xmark & \xmark & 19.4 & \textbf{398,761} & 303.4 & 84.6 & \textbf{15,846} & \textbf{34.8} & \underline{37.5} & \textbf{571,286} & \textbf{392.6} \\
 & \cellcolor{vanillacell}Vanilla Skills & \cmark & \xmark & \xmark & 23.6\betterup{4.2} & \underline{431,096}\worseup{32.3k} & \textbf{243.1}\betterdown{60.3} & 91.3\betterup{6.7} & 3,206,972\worseup{3191.1k} & 56.1\worseup{21.3} & 31.1\worsedown{6.4} & 2,541,279\worseup{1970.0k} & 635.6\worseup{243.0} \\
 & \cellcolor{vectorcell}Vector Skills & \cmark & \cmark & \xmark & 26.4\betterup{7.0} & 485,712\worseup{87.0k} & 287.6\betterdown{15.8} & 92.8\betterup{8.2} & 31,408\worseup{15.6k} & \underline{36.5}\worseup{1.7} & 32.2\worsedown{5.3} & 622,735\worseup{51.4k} & 448.1\worseup{55.5} \\
 & \cellcolor{graphcell}Graph of Skills & \bandcell{\cmark} & \bandcell{\cmark} & \bandcell{\pmark} & \bandcell{\underline{27.1}\betterup{7.7}} & \bandcell{499,023\worseup{100.3k}} & \bandcell{325.2\worseup{21.8}} & \bandcell{\underline{96.3}\betterup{11.7}} & \bandcell{28,031\worseup{12.2k}} & \bandcell{44.6\worseup{9.8}} & \bandcell{33.0\worsedown{4.5}} & \bandcell{\underline{594,106}\worseup{22.8k}} & \bandcell{\underline{427.4}\worseup{34.8}} \\
 & \ourscell{\goskills} & \ourscell{\cmark} & \ourscell{\cmark} & \ourscell{\cmark} & \ourscell{\textbf{31.9}\betterup{12.5}} & \ourscell{521,658\worseup{122.9k}} & \ourscell{\underline{284.9}\betterdown{18.5}} & \ourscell{\textbf{97.0}\betterup{12.4}} & \ourscell{\underline{26,267}\worseup{10.4k}} & \ourscell{38.9\worseup{4.1}} & \ourscell{\textbf{44.2}\betterup{6.7}} & \ourscell{674,148\worseup{102.9k}} & \ourscell{573.7\worseup{181.1}} \\
\midrule
\multirow{5}{*}{\modelcell{\qwenicon}{Qwen 3.5}}
 & No Skills & \xmark & \xmark & \xmark & 17.2 & \textbf{431,209} & \underline{494.1} & 65.9 & \textbf{31,184} & \underline{89.2} & \underline{35.6} & \textbf{585,067} & \underline{756.4} \\
 & \cellcolor{vanillacell}Vanilla Skills & \cmark & \xmark & \xmark & 19.1\betterup{1.9} & \underline{440,257}\worseup{9.0k} & \textbf{430.4}\betterdown{63.7} & 76.4\betterup{10.5} & 3,588,241\worseup{3557.1k} & 181.4\worseup{92.2} & 25.5\worsedown{10.1} & 2,836,437\worseup{2251.4k} & 1067.1\worseup{310.7} \\
 & \cellcolor{vectorcell}Vector Skills & \cmark & \cmark & \xmark & \underline{25.3}\betterup{8.1} & 551,106\worseup{119.9k} & 498.3\worseup{4.2} & 74.4\betterup{8.5} & 58,037\worseup{26.9k} & 100.3\worseup{11.1} & 32.6\worsedown{3.0} & 655,241\worseup{70.2k} & 960.2\worseup{203.8} \\
 & \cellcolor{graphcell}Graph of Skills & \bandcell{\cmark} & \bandcell{\cmark} & \bandcell{\pmark} & \bandcell{25.0\betterup{7.8}} & \bandcell{505,338\worseup{74.1k}} & \bandcell{547.8\worseup{53.7}} & \bandcell{\underline{78.4}\betterup{12.5}} & \bandcell{64,509\worseup{33.3k}} & \bandcell{91.4\worseup{2.2}} & \bandcell{28.5\worsedown{7.1}} & \bandcell{\underline{590,953}\worseup{5.9k}} & \bandcell{\textbf{734.2}\betterdown{22.2}} \\
 & \ourscell{\goskills} & \ourscell{\cmark} & \ourscell{\cmark} & \ourscell{\cmark} & \ourscell{\textbf{29.6}\betterup{12.4}} & \ourscell{594,291\worseup{163.1k}} & \ourscell{503.6\worseup{9.5}} & \ourscell{\textbf{80.6}\betterup{14.7}} & \ourscell{\underline{53,912}\worseup{22.7k}} & \ourscell{\textbf{80.8}\betterdown{8.4}} & \ourscell{\textbf{38.6}\betterup{3.0}} & \ourscell{625,988\worseup{40.9k}} & \ourscell{938.4\worseup{182.0}} \\
\midrule
\multirow{5}{*}{\modelcell{\kimiicon}{Kimi K2.5}}
 & No Skills & \xmark & \xmark & \xmark & 18.6 & \textbf{410,538} & 361.2 & 78.1 & \textbf{26,183} & \textbf{61.3} & \underline{36.3} & \textbf{626,205} & \textbf{423.4} \\
 & \cellcolor{vanillacell}Vanilla Skills & \cmark & \xmark & \xmark & 22.8\betterup{4.2} & \underline{434,619}\worseup{24.1k} & \textbf{295.7}\betterdown{65.5} & 84.8\betterup{6.7} & 3,612,027\worseup{3585.8k} & 149.7\worseup{88.4} & 29.6\worsedown{6.7} & 2,668,963\worseup{2042.8k} & 648.2\worseup{224.8} \\
 & \cellcolor{vectorcell}Vector Skills & \cmark & \cmark & \xmark & 26.1\betterup{7.5} & 508,245\worseup{97.7k} & \underline{336.3}\betterdown{24.9} & \underline{88.6}\betterup{10.5} & 55,471\worseup{29.3k} & \underline{63.4}\worseup{2.1} & 30.7\worsedown{5.6} & 657,488\worseup{31.3k} & 469.6\worseup{46.2} \\
 & \cellcolor{graphcell}Graph of Skills & \bandcell{\cmark} & \bandcell{\cmark} & \bandcell{\pmark} & \bandcell{\underline{28.9}\betterup{10.3}} & \bandcell{448,172\worseup{37.6k}} & \bandcell{358.9\betterdown{2.3}} & \bandcell{87.8\betterup{9.7}} & \bandcell{\underline{50,836}\worseup{24.7k}} & \bandcell{73.8\worseup{12.5}} & \bandcell{32.2\worsedown{4.1}} & \bandcell{676,329\worseup{50.1k}} & \bandcell{\underline{436.6}\worseup{13.2}} \\
 & \ourscell{\goskills} & \ourscell{\cmark} & \ourscell{\cmark} & \ourscell{\cmark} & \ourscell{\textbf{31.2}\betterup{12.6}} & \ourscell{545,407\worseup{134.9k}} & \ourscell{351.1\betterdown{10.1}} & \ourscell{\textbf{92.5}\betterup{14.4}} & \ourscell{53,459\worseup{27.3k}} & \ourscell{65.3\worseup{4.0}} & \ourscell{\textbf{42.7}\betterup{6.4}} & \ourscell{\underline{632,436}\worseup{6.2k}} & \ourscell{583.8\worseup{160.4}} \\
\Xhline{1.1pt}
\end{tabular}%
}
\arrayrulecolor{black}
\vspace{-0.9em}
\end{table*}
agent-only task-processing time, excluding environment setup. Hosted-API runtime remains sensitive to provider load; agent-turn counts provide complementary evidence of organizational effort. Appendix~\ref{app:benchmark_protocol} describes the benchmark protocol, while
Appendix~\ref{app:run_provenance} reports valid-run denominators and aggregation
provenance for the entries in Table~\ref{tab:main_results}.

\subsection{Main Results}
\label{sec:main_results}

We present the main results in Table~\ref{tab:main_results} and
Figure~\ref{fig:model_method_reward_bars}; run provenance and valid-run denominators are
reported in Appendix~\ref{app:run_provenance}. Across SkillsBench, ALFWorld, and
Terminal-Bench 2, \goskills improves over the \emph{No Skills} setting for all
evaluated backbones, showing that the retrieved group-structured context provides useful
task guidance. It also exceeds \emph{Vanilla Skills} on every backbone and benchmark. Relative to
\emph{Graph of Skills}, its agent-only runtime is lower in five of six SkillsBench and all six
ALFWorld cells and higher in all six Terminal-Bench 2 cells.

\paragraph{SkillsBench.}
On SkillsBench, \goskills obtains the highest mean reward for every backbone. Under GPT-5.4, reward increases from
35.7 to 39.4 and runtime drops from 192.9s to 187.3s. Under Claude Sonnet 4.5,
reward increases from 27.1 to 31.9 and runtime drops from 325.2s to 284.9s.

These results suggest that the gain comes not only from retrieving relevant skills but also
from presenting them usably. SkillsBench tasks often carry explicit artifacts, output formats,
or public checks. A dependency-aware bundle may contain useful skills yet leave the entry
point and support roles implicit; the \textsc{Start}, \textsc{Support}, \textsc{Check}, and
\textsc{Avoid} fields make them explicit, reducing the burden of reorganizing context.


\paragraph{ALFWorld.}
\goskills obtains the highest reward across the evaluated backbones.
Compared with Graph of Skills, reward increases by 0.7--4.7 percentage points
and agent-only runtime decreases across all six backbones.
The smaller absolute margins reflect headroom: No Skills already reaches 65.9--84.6
on five of six backbones, where \goskills removes 43--81\% of the remaining error.
ALFWorld also exposes fewer visible technical requirements and no deterministic checks,
so the \textsc{Check}, \textsc{Avoid}, and debt fields carry less signal.

\begin{figure*}[t]
    \centering
    \includegraphics[width=\linewidth]{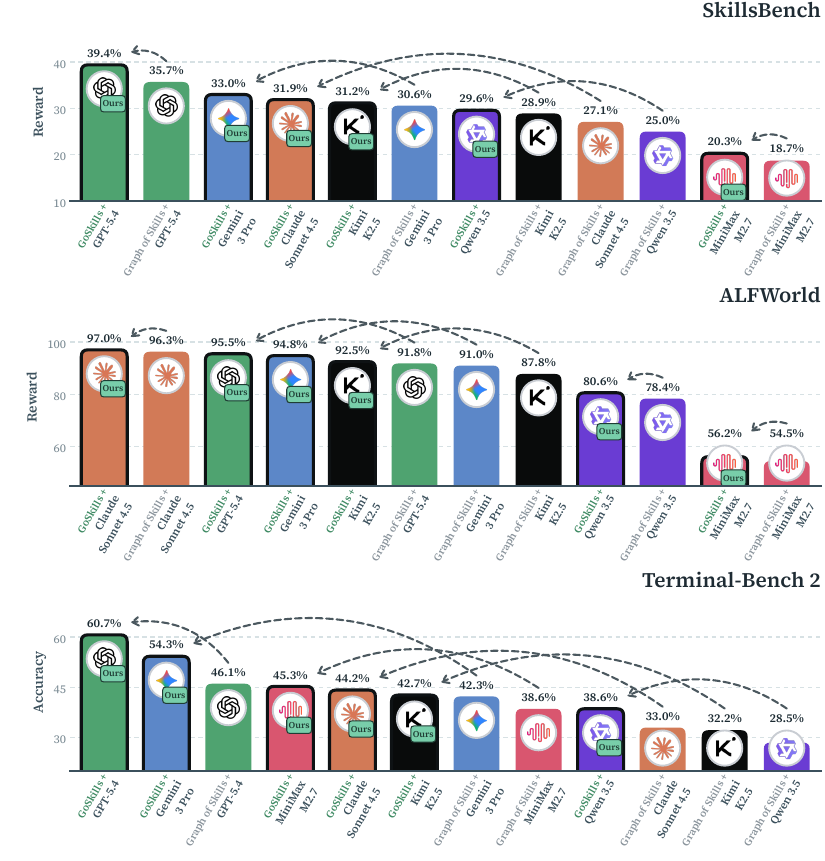}
    \caption{
    \goskills versus Graph of Skills under each agent backbone. Rows report
    \textbf{SkillsBench} reward, \textbf{ALFWorld} reward, and \textbf{Terminal-Bench 2}
    accuracy; bars within a row are sorted by score. Bar color and logo identify the
    backbone, outlined bars marked ``Ours'' are \goskills, and each dashed arrow links a
    backbone's Graph of Skills result to its \goskills result. The reward and accuracy axes do not start
    at zero. Results for all methods are in Table~\ref{tab:main_results}.
    }
    \label{fig:model_method_reward_bars}
    \vspace{-0.6em}
\end{figure*}

\paragraph{Terminal-Bench 2.}
\label{sec:terminalbench}
Terminal-Bench 2 requires multi-step terminal execution, debugging, and verification.
Across all six evaluated backbones, \goskills achieves the highest accuracy
and improves over \emph{Graph of Skills} by 6.7--14.6 percentage points.
Agent-only runtime increases across all six backbones, and cumulative input
tokens increase for five of six (Table~\ref{tab:main_results}). Explicit entry points, support roles, and checks
may help agents organize execution. However, the off-domain 2,000-skill library may lack
task-specific capabilities, and accuracies under our fixed agent loop are not comparable to the
public leaderboard (Appendix~\ref{app:transfer_scope}).

\paragraph{Contribution of group selection.}
Table~\ref{tab:expanded_controls} presents our primary comparison for assessing
group selection on all 87 SkillsBench tasks under GPT-5.4; all methods share the skill
payloads, agent, environment, and retrieval budgets. Relative to Graph of Skills with the
fixed contract, \goskills improves reward by 2.2 percentage points (paired 95\% CI
$[-3.9, 8.8]$) and uses 0.9 fewer turns (95\% CI $[-2.0, 0.2]$). The post-hoc (PH) controls
assign roles to the top retrieved skills with a rule-based (PH-Rule) or LLM-based (PH-LLM)
annotator and perform requirement backfill without constructing groups. Both have lower mean
reward and more turns than full \goskills, consistent with a benefit from group selection
beyond contract formatting and post-hoc role annotation (Appendix~\ref{app:expanded}).

\paragraph{Retrieval-gate diagnostic.}
The must-hit column uses 40 annotated visible-requirement items from the
10-task \texttt{tasks-extra} development subset, separately from the 87-task
end-to-end evaluation. \goskills achieves a 1.00 must-hit rate with fewer than
four presented skills on average, a development-set diagnostic of coverage under the
bottleneck rather than held-out evidence (both gate modes in Appendix~\ref{app:retrieval_gate_details}).

\section{Ablation Study}
\label{sec:ablation}
\begin{figure}[t]
    \centering
    \includegraphics[width=\linewidth]{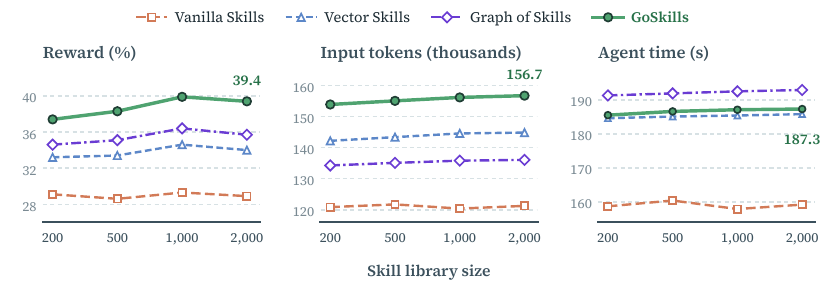}
    \caption{Sensitivity to library size on SkillsBench under GPT-5.4. Points show mean reward, cumulative input tokens, and agent-only time across three repeats at four library sizes. \goskills has the highest mean reward at each size, while its observed runtime remains relatively stable.}
    \label{fig:Data}
\end{figure}

\paragraph{Sensitivity to library size.}
Figure~\ref{fig:Data} compares libraries of 200 to 2,000 skills on SkillsBench
under GPT-5.4. \goskills has the highest reward at each size, peaking at 39.9
with 1,000 skills and reaching 39.4 with 2,000. Its agent time ranges from
185.5 to 187.3 seconds. Vanilla Skills uses task-provided skills on SkillsBench, so its variation across sizes reflects run-to-run noise.

\paragraph{Component ablations.}
Table~\ref{tab:ablation} reports component ablations and a contract-matched
control. The control keeps the Graph of Skills payloads fixed and renders them
with the \goskills{} contract. Reward rises from 35.7 to 37.2, while runtime
falls from 192.9s to 190.6s. Full \goskills{} reaches 39.4 reward and 187.3s.

\begin{table}[t]
\centering
\footnotesize
\caption{GPT-5.4 SkillsBench controls and ablations on 87 tasks with three repeats.
Reward is in percent, tokens are cumulative input tokens, and runtime is agent-only seconds.
Must-hit is measured on 40 development-gate items. Arrows show differences from full
\goskills{} (green/red for improvement/regression, $k=10^3$ tokens).
Bold marks the best values, including ties. Sample SDs are in Appendix~\ref{app:repeat_sd}.}
\label{tab:expanded_controls}\label{tab:ablation}
\setlength{\tabcolsep}{4.5pt}
\renewcommand{\arraystretch}{1.12}
\arrayrulecolor{tableborder}
\begin{tabularx}{\linewidth}{@{}Xrrrrr@{}}
\Xhline{1.1pt}
\tblhead{Method} & \tblhead{Reward $\uparrow$} & \tblhead{Tokens $\downarrow$} & \tblhead{Runtime $\downarrow$} & \tblhead{Turns $\downarrow$} & \tblhead{Must-hit $\uparrow$} \\
\midrule
Graph of Skills & 35.7\worsedown{3.7} & \textbf{136,054}\betterdown{20.6k} & 192.9\worseup{5.6} & 15.8\worseup{1.3} & 0.84\worsedown{0.16} \\
Graph of Skills + \goskills{} Contract & 37.2\worsedown{2.2} & 144,517\betterdown{12.2k} & 190.6\worseup{3.3} & 15.4\worseup{0.9} & 0.84\worsedown{0.16} \\
PH-Rule + backfill & 36.7\worsedown{2.7} & 146,208\betterdown{10.5k} & 193.4\worseup{6.1} & 15.6\worseup{1.1} & 0.81\worsedown{0.19} \\
PH-LLM + backfill & 37.4\worsedown{2.0} & 147,376\betterdown{9.3k} & 195.2\worseup{7.9} & 15.5\worseup{1.0} & 0.83\worsedown{0.17} \\
\rowcolor{ablours}
\goskills & \textbf{39.4} & 156,696 & \textbf{187.3} & \textbf{14.5} & \textbf{1.00} \\
\midrule
w/o Anchor Selection & 38.2\worsedown{1.2} & 155,283\betterdown{1.4k} & 189.1\worseup{1.8} & 14.7\worseup{0.2} & 0.95\worsedown{0.05} \\
w/o Group Expansion & 37.5\worsedown{1.9} & 150,139\betterdown{6.6k} & 190.4\worseup{3.1} & 14.8\worseup{0.3} & 0.90\worsedown{0.10} \\
w/o Group Graph & 38.3\worsedown{1.1} & 154,626\betterdown{2.1k} & 189.5\worseup{2.2} & 14.6\worseup{0.1} & 0.96\worsedown{0.04} \\
w/o Role Labels & 38.6\worsedown{0.8} & 153,558\betterdown{3.1k} & 190.9\worseup{3.6} & 14.8\worseup{0.3} & \textbf{1.00}\unchanged{0.00} \\
w/o Coverage Backfill & 37.8\worsedown{1.6} & 151,492\betterdown{5.2k} & 191.3\worseup{4.0} & 14.9\worseup{0.4} & 0.73\worsedown{0.27} \\
w/o \textsc{Avoid} Field & 38.8\worsedown{0.6} & 154,071\betterdown{2.6k} & 188.8\worseup{1.5} & 14.6\worseup{0.1} & \textbf{1.00}\unchanged{0.00} \\
Retrieved Skills Only & 36.9\worsedown{2.5} & 146,903\betterdown{9.8k} & 192.1\worseup{4.8} & 15.3\worseup{0.8} & 0.76\worsedown{0.24} \\
\Xhline{1.1pt}
\end{tabularx}
\arrayrulecolor{black}
\end{table}

The control study also compares post-hoc role annotation and requirement
backfill under matched budgets. Among individual component removals, group expansion and coverage backfill
produce the largest reward drops. Removing
coverage backfill lowers must-hit from 1.00 to 0.73, while
removing role labels or the \textsc{Avoid} field preserves must-hit. Retrieved
Skills Only has a reward of 36.9 and must-hit of 0.76.

\paragraph{Robustness to imperfect library structure.}
With corruption applied before indexing and group construction, and no retuning,
\goskills retains 91.9\% of clean reward under 50\% metadata blanking and still leads
Graph + Fixed Contract by 0.6 points. With raw names and descriptions only, \goskills retains 91.4\% of clean
reward, with a 0.7-point margin. Removing 50\% of atomic typed edges retains 97.7\%,
reflecting the additional metadata evidence for group-level relations; edge removal does not
remove the group graph itself. Documentation quality is therefore an important operating
condition. Appendix~\ref{app:robustness} reports the complete sweep and seed variability.

\section{Limitations}
\label{sec:limitations}

\goskills is limited to inference-time retrieval and contextualization over a
skill library. It does not train the downstream model, execute tools, bind arguments, or
adapt after failures; when a required capability is absent from the library or disconnected
from the group graph, the method can only report remaining coverage debt.
The implementation depends on deterministic schema extraction and on skill metadata and
typed relations constructed from the library; the corruption study shows graceful
degradation, but documentation quality remains an operating condition, and the fixed
scoring coefficients were set on a small development subset. It may be less effective on
tasks with long setup chains, sparse metadata, ambiguous requirements, or hidden
constraints outside the prompt. Efficiency gains are benchmark-dependent: relative to Graph of
Skills, agent time decreases in five of six SkillsBench and all six ALFWorld settings but
increases on Terminal-Bench 2 with an off-domain library. Three repeats per cell also limit
run-to-run uncertainty estimates.

\section{Conclusion}
\label{sec:conclusion}

This paper introduces group-structured skill retrieval, which changes the retrieval unit
from atomic skills to role-aware skill groups rendered as a compact execution contract,
while leaving the downstream model and execution loop unchanged.
Across the three benchmarks, \goskills achieves higher mean reward or accuracy
than the evaluated baselines, and matched controls attribute part of the gain to group
selection beyond contract rendering. Its runtime advantage is benchmark-dependent,
but the GPT-5.4 SkillsBench controls show fewer agent turns; we therefore read the
efficiency result as reduced organizational effort rather than lower token cost.
Overall, the results suggest that skill retrieval should consider not only which skills are relevant, but also how those skills form a usable execution structure under a limited context budget. Future work could replace the current deterministic scoring rules with learned group-selection policies, improve robustness to sparse or noisy skill metadata, and extend group construction to settings with implicit requirements and runtime feedback. 

\newpage
\begin{ack}
This work is supported in part by the Guangdong Basic and Applied Basic Research Foundation under Grant No.~2025A1515012968, in part by the Shenzhen Science and Technology Program under Grant No.~JCYJ20240813113502004, in part by the National Natural Science Foundation of China under Grant No.~62001412, in part by Shenzhen Stability Science Program 2023, in part by the Guangdong Provincial Key Laboratory of Future Networks of Intelligence (Grant No.~2022B1212010001), and in part by the Shenzhen Key Lab of Crowd Intelligence Empowered Low-Carbon Energy Network (Grant No.~ZDSYS20220606100601002). The authors declare no competing interests.
\end{ack}

\bibliographystyle{plainnat}
\bibliography{ref.bib}
\medskip


\newpage

\appendix

\section{Appendix Overview}
\label{app:overview}

The appendix complements the main paper along four axes: method specification,
implementation reproducibility, retrieval analysis, and trajectory-grounded qualitative
analysis. Table~\ref{tab:appendix_roadmap} summarizes the role of the main appendix sections.

\begin{table}[H]
\centering
\small
\caption{Appendix roadmap. Each section is written to support a specific claim or
reproducibility requirement from the main paper.}
\label{tab:appendix_roadmap}
\begin{ApdxTabFrame}
\renewcommand{\arraystretch}{1.15}
\setlength{\tabcolsep}{6pt}
\arrayrulecolor{roadborder}
\begin{tabular}{L{0.28\linewidth}L{0.62\linewidth}}
\toprule
\rowcolor{roadhead}
\apdxhdr{Section} & \apdxhdr{Purpose} \\
\midrule

\rowcolor{roadband}
Additional Method Details &
Defines the retrieval unit, offline group construction, online expansion, bottlenecking,
coverage-debt accounting, contract template, and complexity terms. \\

\rowcolor{roadstripe}
Prompt and Interface Examples &
Shows the fixed interface contract exposed to the downstream agent and explains how
schema, support roles, checks, avoid rules, and remaining debt enter the prompt. \\

\rowcolor{roadband}
Implementation and Hyperparameters &
Reports the fixed budgets, thresholds, feature families, coefficients, and deterministic
operator rules used by \goskills. \\

\rowcolor{roadstripe}
Additional Experimental Details &
Documents the benchmark protocol, SkillsBench retrieval-gate analysis, task-level retrieval
examples, retrieval controls, corruption and transfer studies, paired trajectory results,
and token/runtime accounting. \\

\rowcolor{roadband}
Failure and Error Analysis &
Separates retrieval misses, partial coverage, downstream execution failures, context
overhead, and infrastructure failures. \\

\rowcolor{roadstripe}
Qualitative Analysis &
Provides trajectory-grounded qualitative examples comparing Graph of Skills baseline
trajectories with \goskills retrieval and paired trajectory evidence. \\

\rowcolor{roadband}
Assets and Licenses &
Lists every benchmark, harness, baseline, skill library, and model API used, with
source, version, and license. \\

\bottomrule
\end{tabular}
\end{ApdxTabFrame}
\end{table}

\section{Additional Method Details}
\label{app:method_details}
\raggedbottom 

This section provides implementation-level details for \goskills. The main text describes
the offline/online structure and the key retrieval decisions. The executable details below
specify the released implementation: deterministic group construction, selection,
source-preserving presentation, and contract rendering. These retrieval operators do not
depend on the downstream backbone.

\subsection{Comparison of Retrieval Units}
\label{app:retrieval_unit_comparison}

\begin{table}[H]
\centering
\footnotesize
\setlength{\tabcolsep}{3.0pt}
\caption{Comparison of agent-facing retrieval units. \goskills differs from post-hoc
bundle annotation by using role-aware groups before expansion and bottlenecking.}
\label{tab:app_retrieval_unit_comparison}
\begin{ApdxTabFrame}
\renewcommand{\arraystretch}{1.14}
\begin{tabular}{L{0.15\linewidth}L{0.19\linewidth}L{0.18\linewidth}L{0.19\linewidth}L{0.19\linewidth}}
\toprule
\rowcolor{tablehead}
\apdxhdr{Method} & \apdxhdr{Retrieval object} & \apdxhdr{Expansion object} &
\apdxhdr{Role timing} & \apdxhdr{Agent-facing output} \\
\midrule
\cellcolor{graphcell}Graph of Skills & Skill nodes scored by graph diffusion & Hydrated dependency-aware bundle & Mostly implicit or rendered after retrieval & Bounded skill bundle \\
Bundle reranking & Candidate skill set & Set-level rerank or pruning & Attached after set selection & Role-labeled bundle \\
\rowcolor{methodrow}
\goskills & Small local asymmetric group & Group-to-group support expansion & Used before scoring, expansion, and bottlenecking & Role contract plus payloads \\
\bottomrule
\end{tabular}
\end{ApdxTabFrame}
\end{table}

\subsection{Offline Group Construction}
\label{app:offline_algorithm}

The offline stage constructs a reusable pool of small skill groups before any
test-time query is observed. As shown in Algorithm~\ref{alg:build_groups}, its purpose is not to change the underlying skill
graph, but to summarize local typed neighborhoods into compact, role-annotated
group candidates that can later be retrieved and bottlenecked online. For each
skill, we treat it as a potential lead skill, enumerate bounded groups within
its typed neighborhood, assign intra-group roles, and extract the facets,
artifact signatures, and visible-check cues used by the online contextualizer. We
reject explicit member incompatibilities, require novelty from the second support in a
triple, and deduplicate by lead and members. The resulting pool \(\mathcal G\), graph
\(\mathcal H\), and member/facet indices \(I\) are reused across queries. This group-layer
constructor consumes the supplied atomic graph; it does not query an LLM or reconstruct
that graph online.

\begin{algorithm}[h]
\caption{\textsc{BuildSkillGroupPool}}
\label{alg:build_groups}
\begin{algorithmic}[1]
\REQUIRE skill graph \(G_{\mathcal{S}}=(\mathcal{S},E,w,\phi)\), group size cap \(K_{\max}\), neighbor cap \(n_{\max}\)
\ENSURE group pool \(\mathcal{G}\), group graph \(\mathcal{H}\), inverted index \(I\)
\STATE \(\mathcal{G}\leftarrow\emptyset,\quad I\leftarrow\emptyset\)
\FOR{each lead skill \(s\in\mathcal{S}\)}
    \STATE \(N_s \leftarrow \textsc{TypedNeighborhood}(s,E,n_{\max})\)
    \STATE \(P_s \leftarrow \textsc{EnumerateGroups}(s,N_s,K_{\max})\)
    \FOR{each candidate group \(g\in P_s\)}
        \STATE \(R_g \leftarrow \textsc{AssignRoles}(g,E,\phi)\)
        \STATE \(F_g^+,F_g^{\mathrm{opt}},F_g^-,A_g,V_g \leftarrow \textsc{ExtractGroupFacets}(g)\)
        \IF{\(\textsc{Compatible}(g)\) \textbf{and} \(\textsc{NonRedundant}(g,\mathcal{G})\)}
            \STATE \(\mathcal{G}\leftarrow\textsc{CanonicalInsert}(\mathcal{G},g)\)
        \ENDIF
    \ENDFOR
\ENDFOR
\STATE \(I\leftarrow\textsc{UpdateIndex}(\mathcal{G})\)
\STATE \(\mathcal{H}\leftarrow\textsc{BuildGroupGraph}(\mathcal{G},E,\phi)\)
\RETURN \(\mathcal{G},\mathcal{H},I\)
\end{algorithmic}
\end{algorithm}

\subsection{Online Contextualization Algorithm}
\label{app:online_algorithm}

At inference time, \goskills extracts a schema from explicitly supplied public task
text and activates groups through seed-member and query-facet indices. It selects an
anchor and bounded support groups, reserves eligible leads and query-relevant incoming
prerequisites, and then fills the remaining atomic-skill budget. Backfill can cover residual
high-confidence requirements using already activated skills. Finally, source-preserving
presentation prepares the payloads, recomputes debt, and renders the fixed contract.
No step executes a skill or changes the task environment. If no anchor qualifies,
backfill can still select debt-covering activated skills while the group plan stays empty.

Here \(q\) includes the caller's query and any explicitly supplied public context.
The core does not discover task files automatically. Skill metadata and full skill text
supply offline retrieval features, not requirements in \(\psi(q)\).
Selection initially uses bounded source prefixes to check visible coverage and the full
rendered-context budget. Final presentation may substitute source sections while preserving
every high-confidence facet previously credited to a prefix. Initialization rejects an
empty contract that already exceeds the budget. In Algorithm~\ref{alg:gos},
\(\mathcal S_{\mathcal P}\) denotes the union of current plan members; all admission
operators check the full rendered context, not just the prospective skill body.

\begin{algorithm}[h]
\caption{\goskills group retrieval and expansion.}
\label{alg:gos}
\begin{algorithmic}[1]
\REQUIRE public query/context \(q\), pool \(\mathcal{G}\), graph \(\mathcal{H}\), indices \(I\), seed mode \(R_0\), budget \(\tau\), backfill cap \(c_{\max}\)
\ENSURE group plan \(\mathcal{P}\), presented skills \(B\), remaining debt \(D\), execution contract \(C\)
\STATE \(\psi\leftarrow\textsc{ExtractSchema}(q),\quad\mathcal{F}_{\mathrm{high}}\leftarrow\textsc{HighConfidenceFacets}(\psi)\)
\STATE \(R_q^0\leftarrow\textsc{RetrieveSeedSkills}(R_0,q),\quad B\leftarrow\emptyset\)
\STATE \(\mathcal{G}_q\leftarrow\textsc{CandidateGroups}(I,R_q^0,\psi)\)
\STATE \(\mathcal{A}_q\leftarrow R_q^0\cup\bigcup_{g\in\mathcal{G}_q}\mathcal{S}_g\)
\STATE \(\widehat{\mathcal{G}}_q\leftarrow\textsc{TopGroups}(\mathcal{G}_q,U_{\mathrm{grp}},\psi,R_q^0)\)
\STATE \(g_a\leftarrow\textsc{SelectAnchor}(\widehat{\mathcal{G}}_q,\psi,\tau)\); initialize \(\mathcal P\) to \([g_a]\), or \([]\) if no group is admissible
\WHILE{\(\mathcal{P}\ne[]\) \textbf{and} \(|\mathcal{P}|<P_{\max}\)}
    \STATE \(\mathcal{E}_t\leftarrow(\widehat{\mathcal{G}}_q\cup\textsc{GroupNeighbors}(\mathcal{H},\mathcal{P}))\setminus\mathcal{P}\)
    \STATE \(g^\star\leftarrow\textsc{BestSupport}(\mathcal{E}_t,\mathcal{P},\psi,R_q^0,\tau)\)
    \IF{\(g^\star=\bot\)}
        \STATE \textbf{break}
    \ENDIF
    \STATE Append \(g^\star\) to \(\mathcal{P}\); \(\mathcal{A}_q\leftarrow\mathcal{A}_q\cup\mathcal{S}_{g^\star}\)
\ENDWHILE
\STATE \(B\leftarrow\textsc{InsertReserved}(\mathcal{P},R_q^0,\tau)\)
\WHILE{\(|B|<b\)}
    \STATE \(s^\star\leftarrow\textsc{BestSkill}((\mathcal{S}_{\mathcal{P}}\cap\mathcal{A}_q)\setminus B,B,\mathcal{P},\psi,\tau)\)
    \IF{\(s^\star=\bot\)}
        \STATE \textbf{break}
    \ENDIF
    \STATE Append \(s^\star\) to \(B\)
\ENDWHILE
\STATE \(D\leftarrow\textsc{CoverageDebt}(\mathcal{F}_{\mathrm{high}},B),\quad c_{\mathrm{back}}\leftarrow0\)
\WHILE{\(D\ne\emptyset\) \textbf{and} \(|B|<b\) \textbf{and} \(c_{\mathrm{back}}<c_{\max}\)}
    \STATE \(s^\star\leftarrow\textsc{BestBackfill}(\mathcal{A}_q\setminus B,D,B,\mathcal{P},\psi,\tau)\)
    \IF{\(s^\star=\bot\)}
        \STATE \textbf{break}
    \ENDIF
    \STATE Append \(s^\star\) to \(B\); \(c_{\mathrm{back}}\leftarrow c_{\mathrm{back}}+1\)
    \STATE \(D\leftarrow\textsc{CoverageDebt}(\mathcal{F}_{\mathrm{high}},B)\)
\ENDWHILE
\STATE \(\mathcal{P}\leftarrow\textsc{AnchorPrune}(\mathcal{P},B,\psi)\)
\STATE Prepare source-preserving payloads for \(B\); recompute visible facets and \(D\)
\STATE \(C\leftarrow\textsc{FormatContract}(\mathcal{P},B,D,\psi)\); reject if the full rendering exceeds \(\tau\)
\RETURN \(\mathcal{P},B,D,C\)
\end{algorithmic}
\end{algorithm}

\subsection{Coverage-Debt Accounting}
\label{app:coverage_debt}

Let \(B_{\mathrm{out}}\) and \(D_{\mathrm{out}}\) be the outputs of Algorithm~\ref{alg:gos}.
The coverage debt is
\[
D_{\mathrm{out}}
=
\mathcal{F}_{\mathrm{high}}(q)
\setminus
\bigcup_{s\in B_{\mathrm{out}}}F_s^{\mathrm{vis}}(q) .
\]
Here \(F_s^{\mathrm{vis}}(q)\) is extracted from the skill name and
actual presented body, excluding hidden metadata and omitted source text. During selection
and backfill, the same rule is applied to bounded source prefixes. Final presentation
preserves their credited high-confidence facets and may cover additional ones; debt is
recomputed afterward. Backfill stops when no eligible fitting candidate covers current
debt, the skill budget is full, or its cap is reached. The output records uncovered
requirements without promising a second search after presentation. Coverage and debt
measure lexical evidence in the delivered text, not semantic sufficiency, task success,
or execution verification. Preserving credited facets prevents hidden or truncated text
from receiving coverage credit; it does not certify that the instructions are correct.

\subsection{Execution Contract Template}
\label{app:contract_template}

The released implementation's execution contract template is fixed across tasks. Task-specific content enters only
through the selected groups, presented skills, query schema, and coverage debt.

\begin{table}[H]
\centering
\small
\caption{Fixed execution-contract fields rendered by \goskills. The field names remain
constant across tasks; only the selected content changes.}
\label{tab:app_contract_fields}
\begin{ApdxTabFrame}
\renewcommand{\arraystretch}{1.12}
\begin{tabular}{L{0.20\linewidth}L{0.68\linewidth}}
\toprule
\rowcolor{tablehead}
\apdxhdr{Field} & \apdxhdr{Rendered role} \\
\midrule
\fieldtag{START} & Anchor skill with a fixed rationale, or the no-anchor fallback \\
\fieldtag{SUPPORT} & Support skills with roles such as prerequisite, parser, formatter, checker, or fallback \\
\fieldtag{CHECK} & Visible output formats, artifacts, tests, deterministic constraints, or proof obligations \\
\fieldtag{AVOID} & Explicit query and selected-skill negative-applicability facets \\
\fieldtag{SKILLS} & Hydrated payloads for the presented atomic skills \\
\fieldtag{DEBT} & Remaining high-confidence visible requirements, if any \\
\bottomrule
\end{tabular}
\end{ApdxTabFrame}
\end{table}

\subsection{Complexity Details}
\label{app:complexity_details}

Let \(M=|R_0(q)|\), \(\bar\iota\) be the average member-index posting length,
\(J_q=\sum_{f\in\psi(q)}|I_f|\) the total query-facet posting length,
\(L\) the shortlist size, \(P_{\max}\) the group cap, \(K_{\max}\) the member cap,
and \(b\) the presented-skill budget. After seed retrieval, candidate activation costs
\(O(M\bar\iota+J_q)\), and ranking costs
\(O(|\mathcal G_q|\log|\mathcal G_q|)\). With bounded facet and payload operations,
support expansion costs \(O(P_{\max}(L+P_{\max}\bar d_{\mathcal H})K_{\max})\),
where \(\bar d_{\mathcal H}\) bounds the visited group-graph degree. Bottlenecking costs
\(O(bP_{\max}K_{\max})\); backfill additionally scans the activated skill set up to
\(c_{\max}\) times. Presentation examines selected source sections and their static
dependency closures; final rendering is linear in the emitted context.

These are group-stage costs. Default lexical seeding also tokenizes library documents,
computes document frequencies, scores them, and sorts matched documents on each query.
For total processed library text/term volume \(T\), vocabulary size \(V\), and
\(N=|\mathcal S|\), a conservative seed-stage bound is \(O(T\log V+N\log N)\).
Thus complete online cost includes a full-library lexical stage. External seed mode
inherits the supplied retriever's cost. Offline enumeration is bounded by the two-neighbor
cap; graph construction uses indexed typed-edge lifts and an exact similarity join whose
cost remains data-dependent.

\section{Prompt and Interface Examples}
\label{app:prompt_interface}

\paragraph{Interface design.}
\goskills uses a fixed downstream interface rather than a task-specific prompt template.
Released lexical seeding splits hyphen and underscore separators to expose retrieval
aliases, without rewriting the schema or adding requirements. Agent-facing content is
determined by the group plan, presented source text, coverage debt, and fixed fields. This section shows the small set of interface components that matter for
reproducibility. The purpose is not to claim that prompting alone is the method; rather, the
prompt is the rendering surface through which group-level retrieval is made operational. This interface is not tuned per task. We do not manually edit START, SUPPORT,
CHECK, AVOID, or DEBT fields for individual examples; the fields are filled
deterministically from the selected group plan, bottlenecked skill payloads,
query schema, and coverage-debt accounting.

\begin{table}[!t]
\centering
\small
\caption{
Interface components exposed by \goskills. Task-specific content enters through retrieved
groups, presented skills, and visible requirements; the field structure is fixed.
}
\label{tab:app_interface_components}
\begin{ApdxTabFrame}
\renewcommand{\arraystretch}{1.12}
\begin{tabular}{L{0.20\linewidth}L{0.27\linewidth}L{0.41\linewidth}}
\toprule
\rowcolor{tablehead}
\apdxhdr{Component} & \apdxhdr{Role} & \apdxhdr{Constraint} \\
\midrule
\fieldtag{START} &
Names the anchor skill with a fixed evidence-based rationale. &
Exactly one primary entry point is rendered when a selected anchor contributes a presented
skill. \\
\fieldtag{SUPPORT} &
Lists support skills and their roles, such as parser, formatter, checker, prerequisite, or
fallback. &
Only skills selected by bottlenecking or backfill are exposed; unpresented group members are
not silently implied. \\
\fieldtag{CHECK} &
States visible output formats, artifacts, deterministic checks, or proof obligations. &
Lists query-schema check, artifact, and constraint facets. This differs from the
high-confidence subset used for debt and excludes hidden evaluator state. \\
\fieldtag{AVOID} &
Surfaces explicit negative-applicability facets. &
Lists query negatives and selected-skill negatives. The text is guidance; it does not
execute a runtime check. \\
\fieldtag{DEBT} &
Reports high-confidence visible requirements still uncovered by the final presented skills. &
Remaining debt is exposed instead of being silently repaired beyond the backfill budget. \\
\bottomrule
\end{tabular}
\end{ApdxTabFrame}
\end{table}

\paragraph{Fixed prompt skeleton.}
Figure~\ref{fig:goskill_contract} shows the schematic field layout. In the released
renderer, the apparent reason slots use fixed wording, not generated explanations:
START states ``Use $\langle$name$\rangle$ first because its group matches the visible task
evidence.'' and asks the agent to inspect the source path. Each SUPPORT item gives a name
and role followed by ``supports the selected plan or an explicit requirement.'' CHECK
lists sorted query check, artifact, and constraint facets; AVOID lists sorted query and
selected-skill negative facets. SKILLS repeats the name, source path, and source-preserving
body of each presented skill; DEBT lists sorted uncovered high-confidence facets.
Without an anchor, START says ``No eligible anchor fits this query and budget. Solve using
the visible task requirements.'' Empty SUPPORT, SKILLS, and DEBT use ``None.''; empty
CHECK uses ``Follow the explicit task requirements.''; empty AVOID uses
``None identified from explicit applicability cues.''

\begin{figure}[h]
    \centering
    \begin{minipage}{0.9\linewidth}
    \begin{contractbox}{GoSkills Execution Contract}
    \vspace{-0.55em}
    \cfield{START}
    Use \cslot{anchor\_skill} first because \cslot{anchor\_reason}.\\
    Inspect its source path before writing new code.
    \cfield{SUPPORT}
    Use these skills only for the stated roles:\\
    \cslot{support\_skill\_1}: \cslot{role\_1} -{}- \cslot{reason\_1}\\
    \cslot{support\_skill\_2}: \cslot{role\_2} -{}- \cslot{reason\_2}
    \cfield{CHECK}
    Before finalizing, verify these visible requirements:\\
    \cslot{visible\_format\_or\_artifact}\\
    \cslot{visible\_test\_or\_constraint}
    \cfield{AVOID}
    Do not follow these incompatible interpretations:\\
    \cslot{negative\_cue\_1}\\
    \cslot{negative\_cue\_2}
    \cfield{SKILLS}
    \cslot{hydrated\_payloads\_for\_presented\_skills}
    \cfield{DEBT}
    \cslot{remaining\_coverage\_debt\_or\_None}
    \end{contractbox}
    \end{minipage}
    \caption{Schematic execution-contract layout. The implementation substitutes
    names, roles, facet lists, and source payloads mechanically; reason placeholders use
    the fixed wording specified in the text.}
    \label{fig:goskill_contract}
\end{figure}

\paragraph{Illustrative contract interpretation.}
Figure~\ref{fig:contract_example} paraphrases the roles of the fields for an invoice
analysis example. This prose is illustrative, not a verbatim renderer output. The executable
template in Figure~\ref{fig:goskill_contract} uses fixed rationale sentences and extracted
facet identifiers; it does not generate task-specific explanations.

\begin{figure}[h]
    \centering
    \begin{minipage}{0.9\linewidth}
    \begin{instancebox}{Illustrative instance: \texttt{invoice-fraud-detection} (paraphrased)}
    \vspace{-0.55em}
    \cfield{START}
    Use \texttt{fuzzy-match} as the anchor skill because the task requires detecting
    suspicious invoice entries under approximate entity and string matching.
    \cfield{SUPPORT}
    \texttt{pdf-reading}: extract invoice text and table fields.\\
    \texttt{xlsx}: parse structured transaction records and preserve row-level identifiers.
    \cfield{CHECK}
    Preserve invoice IDs; produce the required structured output; verify suspicious
    entries against the visible task constraints.
    \cfield{AVOID}
    Do not treat exact string mismatch alone as fraud evidence. Do not ignore missing
    or malformed invoice fields.
    \cfield{SKILLS}
    Hydrated payloads for \texttt{fuzzy-match}, \texttt{pdf-reading}, and \texttt{xlsx}.
    \cfield{DEBT}
    Any explicit requirement absent from the presented source text remains listed.
    \end{instancebox}
    \end{minipage}
    \caption{Illustrative paraphrase of contract semantics for an invoice task, in the
    layout of Figure~\ref{fig:goskill_contract}. These explanations are not literal renderer output.}
    \label{fig:contract_example}
\end{figure}

\paragraph{Relation to retrieval.}
This interface differs from post-hoc bundle annotation because role labels are used before
the final payload is rendered. The selected groups influence anchor choice, support-group
expansion, bottlenecking, backfill, and the contract fields. As a result, the downstream
agent receives a small set of atomic skill payloads together with the retrieval decision's
intended execution structure. This is the mechanism evaluated by the retrieval-gate and
case-study analyses.

\section{Implementation and Hyperparameters}
\label{app:impl}

This appendix records the released implementation profile of the method in
Section~\ref{sec:methodology}. It specifies the frozen code's choices for seeds, offline
construction, admission rules, coverage accounting, and presentation. These retrieval
operators do not depend on the downstream backbone.

\begin{table}[h]
\centering
\small
\caption{
Core settings of the released implementation. The profile fixes these values
across queries.
}
\label{tab:impl_core}
\begin{ApdxTabFrame}
\renewcommand{\arraystretch}{1.10}
\begin{tabular}{L{0.32\linewidth}L{0.58\linewidth}}
\toprule
\rowcolor{tablehead}
\apdxhdr{Component} & \apdxhdr{Setting} \\
\midrule
Seed retrieval evidence & Binary TF--IDF cosine by default; scored evidence with provenance in external mode \\
Group retrieval object & Anchor-centered skill groups \\
Group expansion & Group graph over support, artifact, visible-check, and fallback relations \\
Seed mode & Lexical retrieval by default \\
Returned skill budget & \texttt{top-n}=4 \\
Seed skill count & \texttt{seed-top-k}=4 \\
Maximum skill payload & 1,800 characters \\
Maximum rendered context & 9,000 characters \\
Maximum group size & 3 skills \\
Maximum selected groups & 3 groups \\
Backfill cap & At most 2 activated skills covering current visible debt \\
Rendered output & Execution contract + presented skill payloads + remaining debt \\
Downstream agent & Unchanged \\
\bottomrule
\end{tabular}
\end{ApdxTabFrame}
\end{table}

\begin{table}[h]
\centering
\small
\caption{
Group-pool and selection constants of the released profile. Lexical dictionaries
are versioned with the implementation.
}
\label{tab:impl_hparams}
\begin{ApdxTabFrame}
\renewcommand{\arraystretch}{1.08}
\begin{tabular}{L{0.48\linewidth}L{0.42\linewidth}}
\toprule
\rowcolor{tablehead}
\apdxhdr{Parameter} & \apdxhdr{Value} \\
\midrule
Complexity weight & 0.60 \\
Ambiguity weight & 0.40 \\
Ambiguity gap / spread weights & 0.55 / 0.45 \\
Base pool cap minimum & 6 \\
Top-$n$ pool multiplier & 2 \\
Adaptive extra base & 1.0 \\
Adaptive difficulty multiplier & 2.0 \\
Candidate pool cap & 32 \\
Score-floor center & 0.55 \\
Score-floor difficulty slope & 0.30 \\
Score-floor absolute minimum & 0.10 \\
Minimum required floor / ceiling & 3 / 6 \\
Group selection minimum score & 0.14 \\
Support group / bottleneck skill minimum & 0.10 / 0.10 \\
Inter-group affinity threshold & 0.35 \\
Offline neighbor cap & 2 \\
Prior coefficient \(\lambda_{\pi}\) & 0.05 \\
Anchor correction \(\lambda_{a}\) & 0.18 \\
\bottomrule
\end{tabular}
\end{ApdxTabFrame}
\end{table}

\subsection{Scoring Weights and Rule Definitions}
\label{app:scoring_rules}

All scores used by \goskills are deterministic weighted sums over normalized features.
Each scalar feature is clipped to \([0,1]\) before scoring. The feature vectors
\(\mathbf{x}\), \(\mathbf{z}\), and \(\mathbf{h}\) follow the column order in
Table~\ref{tab:impl_weights}; stage-inapplicable features are set to zero. Penalty terms use
negative coefficients, and rows are not constrained to sum to one because feature families
are computed on different clipped scales. The coefficients in Table~\ref{tab:impl_weights}
comprise the 24 fixed coefficients of this profile. Lexical dictionaries, selection
predicates, and presentation policy are versioned with the source code.

\begin{table}[h]
\centering
\small
\caption{
Deterministic feature families used by the group, support, and bottleneck scores.
}
\label{tab:method_features}
\begin{ApdxTabFrame}
\renewcommand{\arraystretch}{1.10}
\begin{tabular}{L{0.24\linewidth}L{0.66\linewidth}}
\toprule
\rowcolor{tablehead}
\apdxhdr{Feature family} & \apdxhdr{Definition} \\
\midrule
Retriever relevance & Normalized seed-retrieval score of matched members \\
Facet coverage & Novel query-facet overlap with offline group/member facets for scoring;
visible text alone determines coverage debt \\
Anchor match & Exact or normalized match to technology and artifact anchors \\
Visible-check support & Coverage of tests, formats, artifacts, or proof obligations \\
Connectivity & Internal typed-edge support, or its maximum with a group-edge weight from
the plan during support selection \\
Redundancy penalty & Candidate-facet overlap with selected members' offline facets \\
Negative applicability & Explicit negative/high-confidence or anchor/query-negative conflict \\
Cost penalty & Candidate rendered-payload characters divided by the full-context cap \\
\bottomrule
\end{tabular}
\end{ApdxTabFrame}
\end{table}

\begin{table}[t]
\centering
\small
\setlength{\tabcolsep}{5.2pt}
\caption{
Fixed coefficients for the three scoring stages. Rel. denotes seed or group relevance,
Facet denotes query-facet coverage, Anch. denotes anchor match, Check denotes visible-check
support, Conn. denotes graph connectivity, Redun. denotes redundancy, Neg. denotes
negative applicability, and Cost denotes payload cost.
}
\label{tab:impl_weights}
\begin{ApdxTabFrame}
\renewcommand{\arraystretch}{1.18}
\begin{tabular}{@{}lrrrrrrrr@{}}
\toprule
\rowcolor{tablehead}
\apdxhdr{Score} & \apdxhdr{Rel.} & \apdxhdr{Facet} & \apdxhdr{Anch.} &
\apdxhdr{Check} & \apdxhdr{Conn.} & \apdxhdr{Redun.} & \apdxhdr{Neg.} &
\apdxhdr{Cost} \\
\midrule
\(U_{\mathrm{grp}}\) & 0.28 & 0.22 & 0.18 & 0.12 & 0.10 & $-$0.05 & $-$0.25 & $-$0.04 \\
\(U_{\mathrm{sup}}\) & 0.12 & 0.28 & 0.06 & 0.16 & 0.16 & $-$0.18 & $-$0.25 & $-$0.04 \\
\(U_{\mathrm{bot}}\) & 0.18 & 0.24 & 0.12 & 0.20 & 0.08 & $-$0.12 & $-$0.30 & $-$0.08 \\
\bottomrule
\end{tabular}
\end{ApdxTabFrame}
\end{table}

\paragraph{Feature normalization.}
Seed relevance is normalized by the largest returned seed score. Group relevance is the
maximum over its members; support relevance uses only members outside the current plan;
bottleneck relevance is the individual skill score. Coverage and check features measure
the fraction of the corresponding query-facet set matched by novel offline facets; an
empty requested set scores zero. They guide ranking but do not certify rendered coverage.
Lead/skill anchor match is 1 for an explicit high-confidence technology or artifact match,
0.7 for a normalized anchor match, and 0 otherwise. Connectivity sums applicable positive
internal edge weights; support selection takes the maximum of that value and the strongest
outgoing group-edge weight from the plan. Redundancy divides already covered candidate
facets by all candidate facets. Negative applicability is binary and also acts as a hard
eligibility filter. Cost divides the rendered candidate payload length, including its skill
name and source path, by 9,000. All eight values are clipped to \([0,1]\).
The group score additionally uses \(0.05\) times its clipped prior (the mean positive
internal edge weight); anchor correction adds \(0.18\) times anchor match.

\paragraph{Adaptive shortlist.}
Let \(c\) be the fraction of the seven query-schema categories that are nonempty. From
normalized seed scores, let \(\Delta\) and \(s\) be the top-two gap and full spread;
both are 1 when fewer than two seeds are returned. Define
\[
 a=0.55(1-\operatorname{clip}(\Delta))+0.45(1-\operatorname{clip}(s)),
 \qquad d=0.60c+0.40a .
\]
The cap is \(L=\min\{32,\lceil\max(6,2b)+1+2d\rceil\}\), with score floor
\(\max\{0.10,0.55-0.30d\}\max_{g\in\mathcal G_q}U_{\mathrm{grp}}(g,q)\) (zero for no candidates).
Among the first \(L\) ranked groups, retain those above the floor and always retain the
first \(\min\{L,\lceil3+3d\rceil\}\), subject to availability. Ties use lead seed
rank, smaller group size, and stable group ID in that order.

\begin{table}[h]
\centering
\small
\caption{
Offline group-pool construction rules used by Algorithm~\ref{alg:build_groups}.
}
\label{tab:impl_group_rules}
\begin{ApdxTabFrame}
\renewcommand{\arraystretch}{1.12}
\begin{tabular}{L{0.28\linewidth}L{0.64\linewidth}}
\toprule
\rowcolor{tablehead}
\apdxhdr{Rule} & \apdxhdr{Definition} \\
\midrule
\textsc{TypedNeighborhood} &
Collect positive incoming/outgoing dependency, workflow, artifact, visible-check, fallback,
alternative, and semantic edges, in that priority order. Keep each neighbor's best edge;
order by priority, descending weight, and stable IDs; retain at most two neighbors. \\
\textsc{EnumerateGroups} &
Enumerate the singleton, lead--neighbor pairs, and the lead with two retained neighbors,
subject to the three-member cap. Only triples require the second support to add a distinct
role or an artifact/check facet beyond the lead and first support. \\
\textsc{AssignRoles} &
The lead is the anchor. Incoming/outgoing dependency gives prerequisite/dependent; workflow
gives preprocessor/postprocessor; artifact gives parser/formatter. Visible-check gives
checker; alternative/fallback gives fallback; semantic gives support. \\
\textsc{ExtractGroupFacets} &
Extract facets from skill names, descriptions, source text, and declared public metadata.
Lead facets are required; other member facets are optional. Negatives are separate;
IDs, paths, and exclusivity fields do not create positive capabilities. \\
\textsc{Compatible} &
Reject explicit incompatibility edges, member-negative versus positive-facet/name/ID conflicts,
\texttt{exclusive\_with} matches, or disjoint choices in a common declared exclusivity
category. Different technology or format mentions alone do not imply incompatibility. \\
\textsc{NonRedundant} / \textsc{CanonicalInsert} &
Apply the triple novelty rule above. Canonicalize by lead and sorted members and retain the
higher-prior duplicate. Pairs have no additional general facet-novelty requirement. \\
\textsc{UpdateIndex} &
Build member-to-group and positive-facet-to-group inverted indices after deduplication. \\
\textsc{BuildGroupGraph} &
Lift atomic edges with their direction and relation type; add exact facet-similarity links.
Symmetric incompatibility suppresses every positive relation for that pair. Positive
relations must reach affinity 0.35; details follow below. \\
\bottomrule
\end{tabular}
\end{ApdxTabFrame}
\end{table}

\paragraph{Operator semantics.}
Tables~\ref{tab:impl_group_rules} and~\ref{tab:impl_online_rules} define the deterministic
operators of Algorithms~\ref{alg:build_groups} and~\ref{alg:gos} for the released profile.
For an atomic-edge lift, affinity is the maximum of its clipped weight and the full
positive-facet Jaccard similarity between groups. Semantic and alternative relations are
symmetric; other positive atomic relations preserve direction. Dependency, workflow, and
semantic edges lift to support, and alternative edges to fallback. Metadata-only links
use a rare-first inverted-prefix candidate join followed by exact Jaccard verification at
0.35. A qualifying pair receives symmetric support links, plus artifact/check links when
those sets overlap. Sharing a format alone does not bypass the full-facet threshold.

\begin{table}[h]
\centering
\small
\caption{
Online selection and rendering rules used by Algorithm~\ref{alg:gos}.
}
\label{tab:impl_online_rules}
\begin{ApdxTabFrame}
\renewcommand{\arraystretch}{1.12}
\begin{tabular}{L{0.28\linewidth}L{0.64\linewidth}}
\toprule
\rowcolor{tablehead}
\apdxhdr{Rule} & \apdxhdr{Definition} \\
\midrule
\textsc{ExtractSchema} &
Extract seven categories from the query and explicitly supplied public context using fixed
lexical dictionaries. Skill metadata remains an offline retrieval input; it does not add
query requirements. \\
\textsc{HighConfidenceFacets} &
Use explicit canonical technologies, named API calls, recognized formats/extensions,
artifact filenames, and stated constraints. An alias alone need not be high-confidence;
a filename obligation is distinct from its format. \\
\textsc{CandidateGroups} &
Union query-facet postings with seed-member postings. Reject a group when its negatives
conflict with high-confidence query facets, or its anchors conflict with query negatives. \\
\textsc{TopGroups} &
Rank by weighted group score plus prior; apply the adaptive cap, floor, and guaranteed
minimum above. Ties use lead seed rank, group size, and stable group ID. \\
\textsc{SelectAnchor} &
Add anchor correction. A group is admissible if its corrected score is at least 0.14, its lead
is eligible, and its reserved set fits the budget. Prefer admissible groups with positive
lead-anchor match; otherwise take the best admissible group. Return no anchor if none is admissible. \\
\textsc{BestSupport} &
Require member compatibility, query eligibility, grounded evidence and marginal novelty
as defined below, score at least 0.10, and reserved-set feasibility. Maximize support score
with the group tie-breaker. Stop at three groups or no admissible option. \\
\textsc{GroupNeighbors} &
Collect positive outgoing non-incompatibility neighbors of the plan. A group edge alone
does not satisfy the grounded-support predicate. \\
\textsc{InsertReserved} &
Insert unique leads in plan order, then positive-seed incoming dependency/workflow members
with prerequisite/preprocessor roles. Admission checks this entire reserved set for
eligibility, compatibility, skill count, and rendered-context fit. \\
\textsc{BestSkill} &
Among activated members of the plan, require eligibility, compatibility, positive marginal
coverage/check/connectivity, score at least 0.10, and fit. Maximize score, then use seed
rank and stable skill ID. \\
\textsc{CoverageDebt} &
Subtract the union of currently presented name/body facets from high-confidence query
facets; do not credit metadata or omitted text. \\
\textsc{BestBackfill} &
Among activated skills outside \(B\), require visible debt gain, eligibility, compatibility,
and fit. Prioritize debt facets covered, then bottleneck score, seed rank, and skill ID.
Add at most two; the ordinary skill-score threshold is not reapplied. \\
\textsc{AnchorPrune} &
If needed, promote the first plan group whose presented lead matches a high-confidence
anchor. Retain only groups contributing a presented member. \\
\textsc{FormatContract} &
After source presentation, recompute debt and render START, SUPPORT, CHECK, AVOID,
SKILLS, and DEBT. Enforce the final 9,000-character cap, including all fields and paths. \\
\bottomrule
\end{tabular}
\end{ApdxTabFrame}
\end{table}

\paragraph{Grounded support and reservation.}
A candidate member has support evidence if it has positive seed relevance, adds a visible
technology/operation/failure/check facet beyond current plan members, or is a positive
dependency/workflow predecessor of a candidate-group member having either form of query
evidence. A newly introduced mandatory lead must itself satisfy this predicate; if the
lead is already a plan lead, evidence from an added member suffices. Shared artifacts or
generic constraints alone are insufficient. Separately, the candidate must add a
query-matching offline facet or have a positive group connection from the plan and at
least one new member. These predicates precede scoring. Reserved predecessors are narrower:
they connect into a selected lead, carry prerequisite/preprocessor roles, and have positive
seed relevance. Candidate support groups must also be pairwise compatible with all members
of the existing plan.

\paragraph{Seeds and source-preserving presentation.}
Default seeds use binary TF--IDF cosine over public skill documents, with
\(\operatorname{idf}(t)=1+\log((1+N)/(1+\operatorname{df}(t)))\) for \(N\) documents.
Both query and document lengths are normalized; ties use skill ID. Hyphens and underscores are split
only for retrieval aliases; no model rewrites the query. External mode requires explicit
nonnegative scored skill IDs with source provenance in a separate configuration.

After selection, complete source sections are ranked by query overlap, relevant workflow
or imperative cues, and deterministic tie-breakers. The extractor includes ancestor
headings and statically identified earlier Python-example definitions needed by a section,
without executing examples. It returns only source spans within 1,800 body characters.
A decision-tree heading triggers the disclosed prefix fallback. Otherwise, sections with
their dependency closures are considered within the cap; the prefix is used if no source
sections fit. If extraction would lose a high-confidence facet already credited to the
prefix, that prefix is retained. The full contract is then checked against 9,000 characters;
an overflow is rejected rather than silently truncated.

\paragraph{Notation.}
The implementation uses internal token sets for normalized query terms, technology
anchors, operation hints, artifacts, constraints, failure cues, and visible-check cues. These
correspond to the query schema \(\psi(q)\) in Section~\ref{sec:methodology}. We use the
paper notation throughout; \(F_s\) denotes offline skill facets, while
\(F_s^{\mathrm{vis}}(q)\) denotes facets of the presented name/body.
Bundles record source spans, policy/configuration hashes, selected roles, and remaining
debt so that this distinction can be inspected.

\flushbottom

\section{Additional Experimental Details}
\label{app:exp_details}

\subsection{Benchmark and Evaluation Protocol}
\label{app:benchmark_protocol}

The evaluation covers SkillsBench~\citep{li2026skillsbench},
ALFWorld~\citep{shridhar2021alfworld}, and Terminal-Bench 2.
Table~\ref{tab:main_results} compares skill-access methods across six backbones
on SkillsBench, ALFWorld, and Terminal-Bench 2.
Table~\ref{tab:expanded_controls} examines group selection through matched
SkillsBench controls on 87 tasks.

\paragraph{Task coverage and environment requirements.}
The six-backbone comparison uses the full SkillsBench 1.1 suite (87 tasks),
all 134 unseen ALFWorld \texttt{eval\_out\_of\_distribution} tasks, and the
full 89-task Terminal-Bench 2 suite. The GPT-5.4 SkillsBench control study
also covers all 87 tasks. Each model--method cell uses three repeats.
Tables~\ref{tab:run_provenance} and \ref{tab:control_provenance} report the valid-run counts.
The Vanilla Skills condition uses the skills supplied with each SkillsBench task.
ALFWorld and Terminal-Bench 2 expose their full experimental libraries.
These are the configurations of the reported runs. The current public SkillsBench
runner mounts an externally supplied skill directory, so reproducing this
Vanilla condition requires supplying the corresponding task-specific skills.
The public release and its default task manifest are separate from the final
evaluation summaries reported here.

All three benchmarks in the six-backbone comparison use 2,000-skill libraries,
as does the SkillsBench control study. Smaller libraries are used only in the
library-size analyses.

The implementation description in this appendix follows the public repository
at commit \texttt{3c7de99}. Its configurable budgets and thresholds permit
alternative settings; the reported experiments use the settings listed here.

\paragraph{Development and evaluation separation.}
The 24 scoring coefficients and four facet dictionaries were fixed using
10 development tasks selected from the official \texttt{tasks-extra} pool, disjoint
from all 87 main SkillsBench tasks. The same tasks supply the 40 gate annotations. Thus end-to-end reward
is evaluated separately from development, whereas gate coverage is a development-set
mechanism diagnostic.

\paragraph{Metrics and task-level diagnostics.} The
task-level analyses in this appendix focus on SkillsBench because its tasks are paired with
reusable skills and deterministic checks, making it possible to inspect retrieved skill
coverage and matched trajectories. Each task is executed by the same downstream agent loop,
with only the retrieved skill context changed across methods. We use reward as the primary
end-to-end metric and report token usage and runtime as efficiency metrics. Graph of Skills
retrieves dependency-aware skill bundles, while \goskills retrieves anchor-centered skill
groups and expands support groups before rendering the final execution contract. The
retrieval-gate and case-study tables below are therefore task-level SkillsBench analyses,
not a separate re-aggregation of the ALFWorld results.

For retrieval evaluation, we define task-specific \texttt{must\_have} skills and evaluate
40 annotated visible-requirement items per gate mode. A requirement-level pass means that
the final presented context includes the required skill for that item. A partial result
covers a related but incomplete required skill set, and a miss covers none. We report
requirement-level pass, partial, miss, average must-hit rate, selected group count, and
presented skill count.

We separate infrastructure failures from substantive agent failures. Runs with environment
construction errors, Docker failures, or startup failures are tracked separately. If the agent
has already entered a meaningful trajectory and then times out, we treat the timeout as a
substantive execution failure. Invalid runs are not used as evidence for retrieval quality.

\subsection{Agent-facing instructions in the evaluation harness}
\label{app:agent_prompts}

All methods share the same agent loop, task instruction, and environment; they differ only in
the skill-access instructions and the retrieval command. On SkillsBench, every task instruction
is preceded by the same one-sentence prefix, and each method provides its own
\texttt{/root/AGENTS.md}. The prefix and the \goskills{} instructions are reproduced verbatim
below; Table~\ref{tab:arm_instructions} lists how the other arms differ.

\begin{promptbox}{Task prefix (all methods, SkillsBench)}
\begin{Verbatim}[breaklines,breakanywhere,fontsize=\fontsize{7.3pt}{9.2pt}\selectfont]
Before writing or editing solution code, read /root/AGENTS.md if it exists. Follow its skill-access condition, inspect visible requirements/tests, preserve exact output paths, and prefer the shortest verifier-aligned solution.
\end{Verbatim}
\end{promptbox}

\begin{promptbox}{\texttt{/root/AGENTS.md} for \goskills{} (verbatim)}
\begin{Verbatim}[breaklines,breakanywhere,fontsize=\fontsize{7.3pt}{9.2pt}\selectfont]
# Task Environment

This environment contains a prebuilt **GoSkills** group pool.

## Required First Step

Before writing any code, retrieve relevant skills by running a targeted query:

```bash
goskills-query "goal + artifact/format + operation/API + verifier-critical constraint"
```

This command is available on PATH. It queries a prebuilt group pool and returns
a bounded role-labeled contract and skills relevant to your task.

When writing the query, include only the retrieval-critical task facts that are actually known:

- concrete goal
- artifact, file format, or main object
- key operation, algorithm, API, or library
- verifier-critical constraint or invariant

Examples:

```text
update embedded xlsx in pptx preserve formulas
parallel tfidf search processpoolexecutor deterministic ranking
exact civ6 district adjacency calculator
```

Avoid vague queries such as `solve this task` or `help with benchmark`.

Read the START and DEBT fields. If no anchor/skills were selected, note the no-hit and continue without claiming skill use. Otherwise use only the presented skills and their stated roles.

## Reading the Output

Each returned skill includes a `Source:` path, e.g.:

```
Source: /opt/graphskills/skills/mesh-analysis/SKILL.md
```

The full skill directory (including `scripts/` and `references/`) is at that path.
For example: `/opt/graphskills/skills/mesh-analysis/scripts/mesh_tool.py`

**Always check whether a ready-made script exists before implementing from scratch.**

Before implementing, inspect the task requirements, tests, and verifier and identify the minimum acceptance requirements.

Priorities:

1. Take the shortest path to passing the verifier.
2. Pass only the verifier's minimum required behavior first.
3. Use the exact `Source:` path returned by retrieval. Do not reconstruct paths from the skill name or scan the whole skill library if a `Source:` path is already available.
4. Reuse or adapt retrieved scripts and interfaces when they directly fit the verifier target.
5. If a retrieved skill contains an authoritative calculator, validator, parser, or pack/unpack workflow, use that exact interface for the final output and final self-check.
6. Treat skills as a way to shrink the search space, not as permission to explore more implementation branches.
7. Avoid extra features, UI expansion, side outputs, or generalization unless explicitly required.

## Workflow

1. Inspect the task and form a short retrieval query containing `goal + artifact/format + operation/API + verifier-critical constraint`
2. Run `goskills-query "<targeted query>"`
3. Record whether retrieval was a skill hit or no-hit
4. Inspect task requirements/tests/verifier and write down the minimum acceptance requirements
5. Read the returned skill bundle and use the exact `Source:` paths it provides
6. Check for scripts at the parent directory of each returned `Source:` path
7. Use or adapt those scripts only if they directly help satisfy the minimum requirements
8. If no returned script/interface directly fits, stay on the shortest no-frills path to verifier pass
9. Before finalizing, run one verifier-aligned self-check with the retrieved skill's authoritative interface when available

GoSkills uses the fixed START, SUPPORT, CHECK, AVOID, SKILLS, and DEBT fields. Respect the returned roles and report remaining coverage debt. Only the presented skills are selected; unpresented group members are not implied.
\end{Verbatim}
\end{promptbox}

\begin{table}[!t]
\centering
\small
\setlength{\tabcolsep}{3.4pt}
\caption{Skill-access instructions per method on SkillsBench. Apart from the entries below, the
\goskills{}, Graph of Skills, and Vector Skills instructions are identical, including the query
guidance, the priorities, and the workflow.}
\label{tab:arm_instructions}
\begin{ApdxTabFrame}
\renewcommand{\arraystretch}{1.12}
\begin{tabular}{@{}L{0.15\linewidth}L{0.20\linewidth}L{0.24\linewidth}L{0.33\linewidth}@{}}
\toprule
\rowcolor{tablehead}
\apdxhdr{Method} & \apdxhdr{Retrieval command} & \apdxhdr{Described output} & \apdxhdr{Reading rule} \\
\midrule
No Skills & None & None & ``No external skill library is supplied.'' \\
Vanilla Skills & None; skill files are read on demand & Mounted skill directory (\path{/opt/skillsbench/skills}) &
Use skills only when they directly help satisfy the verifier. \\
Vector Skills & \path{vectorskills-query} & Embedding-similarity results only &
Report \path{SKILL_HIT} or \path{NO_SKILL_HIT}; use skills only as constraints; do not hide retrieval failures. \\
Graph of Skills & \path{graphskills-query} & ``A ranked bundle of skills'' &
Report \path{SKILL_HIT} or \path{NO_SKILL_HIT}; use skills only as constraints. \\
\rowcolor{methodrow}
\goskills{} & \path{goskills-query} & ``A bounded role-labeled contract'' &
Read \textsc{Start} and \textsc{Debt}; use only the presented skills and their stated roles. \\
\bottomrule
\end{tabular}
\end{ApdxTabFrame}
\end{table}

\subsection{How retrieval queries are formed}
\label{app:query_formation}

The three benchmarks expose retrieval to the agent differently, but within a benchmark every
retrieval-based method uses the same mechanism.
\begin{itemize}
\item \textbf{SkillsBench.} After reading the instructions above, the agent writes a short query
and calls the method's retrieval command; in repeat~2 every \goskills{} and Graph of Skills run issued exactly one query. In
the protein case of Appendix~\ref{app:qualitative_analysis}, the agent's query was
\texttt{update xlsx spreadsheet formulas preserve formatting exact output path}.
\item \textbf{ALFWorld.} The harness retrieves once from the initial observation and task goal.
During the episode, the agent may emit \texttt{SkillRequest: GOS\_RETRIEVE <goal or blocker>}
for a new retrieval or \texttt{SkillRequest: READ\_SKILL <name>} to read a skill; each request
consumes one interaction step.
\item \textbf{Terminal-Bench 2.} The harness retrieves with the full task instruction before the
first turn; the agent can issue follow-up queries through a \texttt{retrieve\_skills} tool.
\end{itemize}

\begin{promptbox}{ALFWorld system prompt and skill-access message (\goskills)}
\begin{Verbatim}[breaklines,breakanywhere,fontsize=\fontsize{7.3pt}{9.2pt}\selectfont]
Interact with a household to solve a task. Imagine you are an intelligent agent in a household environment and your target is to perform actions to complete the task goal. At the beginning of your interactions, you will be given the detailed description of the current environment and your goal to accomplish. 
For each turn, think briefly and then choose exactly one of these two outputs:
1. `Thought: ...` followed by `Action: ...` for a normal environment action.
2. `Thought: ...` followed by `SkillRequest: ...` if you want extra skill help before acting.

Use `SkillRequest` when it directly helps the next step, especially if you are blocked, the action syntax is unclear, the retrieved skill names look unrelated to the current blocker, or 1-2 actions already failed. In GoS mode, prefer `SkillRequest: GOS_RETRIEVE ...` when you need a better-targeted skill search. In vector mode, prefer `SkillRequest: VECTOR_RETRIEVE ...` when you need a better-targeted vector-only search. Use `READ_SKILL` only when you already know which exact skill is worth reading. Do not output both `Action:` and `SkillRequest:` in the same turn.

The available actions are:
1. go to {recep}
2. take {obj} from {recep}
3. move {obj} to {recep}
4. open {recep}
5. close {recep}
6. use {obj}
7. clean {obj} with {recep}
8. heat {obj} with {recep}
9. cool {obj} with {recep}
where {obj} and {recep} correspond to objects and receptacles.
After your each turn, the environment will give you immediate feedback based on which you plan your next few steps. if the envrionment output "Nothing happened", that means the previous action is invalid and you should try more options.

Your response should use one of the following formats:

Thought: <your thoughts>
Action: <your next action>

or

Thought: <your thoughts>
SkillRequest: <your tool-style request>

[user]
GoSkills provides an execution contract from the current visible query. Follow applicable skills conditionally. For a focused new retrieval use `Thought: ...` then `SkillRequest: GOS_RETRIEVE <current visible goal or blocker>`. To read a named skill use `SkillRequest: READ_SKILL <exact skill name>`. Each request consumes one interaction step; do not also emit Action.

[user]
You are in the middle of a room. Looking quickly around you, you see a bed 1, a desk 2, a desk 1, a drawer 6, a drawer 5, a drawer 4, a drawer 3, a drawer 2, a drawer 1, a garbagecan 1, a laundryhamper 1, a safe 1, a shelf 6, a shelf 5, a shelf 4, a shelf 3, a shelf 2, and a shelf 1.
Your task is to: examine the alarmclock with the desklamp.
\end{Verbatim}
\end{promptbox}
\noindent The last \texttt{[user]} message is the first observation of ALFWorld task~000.

\subsection{A rendered contract from the evaluation}
\label{app:real_contract}

The box below is the contract that \goskills{} returned to the agent for
\texttt{protein-expression-analysis} in repeat~2 (Case~1 of
Appendix~\ref{app:qualitative_analysis}). All field text is verbatim; only the skill payloads are
replaced by their lengths. The anchor (Excel Automation) documents cloud spreadsheet APIs; the
verifier-critical recalculation step comes from the \texttt{xlsx} support skill.

\begin{promptbox}{Rendered contract for \texttt{protein-expression-analysis} (payloads elided)}
\begin{Verbatim}[breaklines,breakanywhere,fontsize=\fontsize{7.3pt}{9.2pt}\selectfont]
START
Use Excel Automation first because its group matches the visible task evidence.
Inspect its source path before writing new code.

SUPPORT
Use these skills only for the stated roles:
- xlsx: support lead — supports the selected plan or an explicit requirement.
- recipe-log-deal-update: support lead — supports the selected plan or an explicit requirement.
- Excel Analysis: dependent — supports the selected plan or an explicit requirement.

CHECK
Before finalizing, verify these visible requirements:
- artifact:xlsx

AVOID
Do not follow these incompatible interpretations:
None identified from explicit applicability cues.

SKILLS
Skill: Excel Automation
Source: /opt/graphskills/skills/excel-automation/SKILL.md
[source-preserving excerpt of 1,579 characters omitted]

Skill: xlsx
Source: /opt/graphskills/skills/xlsx/SKILL.md
[source-preserving excerpt of 1,724 characters omitted]

Skill: recipe-log-deal-update
Source: /opt/graphskills/skills/recipe-log-deal-update/SKILL.md
[source-preserving excerpt of 625 characters omitted]

Skill: Excel Analysis
Source: /opt/graphskills/skills/excel-analysis/SKILL.md
[source-preserving excerpt of 1,616 characters omitted]

DEBT
None.
\end{Verbatim}
\end{promptbox}

\subsection{SkillsBench Control Study}
\label{app:expanded}

We evaluate group selection on all 87 SkillsBench 1.1 tasks with GPT-5.4
and three repeats per cell. Table~\ref{tab:expanded_controls} is the primary
comparison for assessing group selection. All arms share the atomic candidate
universe, query schema, skill payloads, agent, and environment, with at most
four payloads, 1,800 characters per payload, and a 9,000-character context guard.
Graph + Fixed Contract keeps the structural baseline's payloads fixed and
transfers both contract fields and reading instructions. The post-hoc (PH) controls
assign \textsc{Start}, \textsc{Support}, \textsc{Check}, and \textsc{Avoid} roles to the top
retrieved skills after retrieval, using a rule-based annotator (PH-Rule) or an LLM-based
annotator (PH-LLM), and perform requirement backfill within the shared pre-group candidate
universe; neither constructs groups or traverses a group graph. Each arm retains the reading instructions for its own skill-block format.

\goskills improves reward over Graph of Skills by 3.7 percentage points
(35.7 to 39.4) and uses 1.3 fewer turns (15.8 to 14.5). Relative to the
fixed-contract Graph of Skills control, the reward difference is +2.2 points
(paired 95\% CI $[-3.9, 8.8]$), and the turn difference is $-0.9$
(95\% CI $[-2.0, 0.2]$). Hosted-API timing remains sensitive to provider
load. Tool-call and repair-attempt counts were not separately measured.

The post-hoc controls reach mean rewards of 36.7 and 37.4, compared with 39.4
for full \goskills. Group expansion can expose visible-check skills that
rank low or are absent from seed retrieval, enlarging the activated candidate
set before bottlenecking. The gate ordering (0.81/0.83 for PH controls versus
1.00 for \goskills) is consistent with that mechanism, but the gate's use in
development limits generalization of the coverage result. These controls
evaluate the combined selection and rendering pipeline; the point estimates
indicate a benefit from group selection.

\paragraph{Contract rendering and payload selection.}
Rendering the same Graph of Skills payloads with the fixed contract raises
the observed mean reward by 1.5 points. Full \goskills adds a further 2.2 points
relative to this control.

\subsection{Robustness to Metadata and Atomic-Graph Corruption}
\label{app:robustness}
The atomic graph is produced by the Graph of Skills benchmark's deterministic
construction script, without an LLM; \goskills adds the group layer. Group-level
relations also use metadata facets, which are the dominant construction evidence
on the evaluated corpus. Consequently, incomplete atomic edges and missing
metadata need not have the same effect.
Table~\ref{tab:metadata_robustness} gives the complete corruption sweep under
GPT-5.4. Perturbations precede indexing and group construction, and all arms
share the perturbed library with no hyperparameter retuning. Reward is measured
on all 87 SkillsBench tasks. Reported standard deviations are over three corruption
seeds, not estimates of repeated-run variation for the clean aggregate tables.
Must-hit uses the 40 annotated development-gate items. The comparator is
Graph + Fixed Contract, which matches the contract while keeping the structural baseline payloads fixed.

\begin{table}[ht]
\centering
\small
\caption{Robustness to imperfect documentation and atomic graph structure. Reward is in percent; $\pm$ denotes sample s.d. across three library-seed means where available, after averaging three agent repeats per seed. Clean and deterministic raw-document conditions have no corruption-seed s.d. Hit is instruction-auto development-gate must-hit. $\Delta$ is the reward gap in percentage points; retention is \goskills{} reward relative to its clean value.}
\label{tab:metadata_robustness}
\begin{ApdxTabFrame}
\renewcommand{\arraystretch}{1.12}
\begin{tabular}{@{}lrrrrrr@{}}
\toprule
\rowcolor{tablehead}
 & \multicolumn{2}{c}{Fixed contract} & \multicolumn{2}{c}{\goskills} & & \\
\rowcolor{tablehead}
\apdxhdr{Condition} & \apdxhdr{Reward} & \apdxhdr{Hit} & \apdxhdr{Reward} & \apdxhdr{Hit} & \apdxhdr{$\Delta$} & \apdxhdr{Retention} \\
\midrule
Clean & 37.2 & 0.84 & 39.4 & 1.00 & +2.2 & 100.0\% \\
25\% metadata blanking & $36.4\pm0.53$ & 0.80 & $37.9\pm0.73$ & 0.95 & +1.5 & 96.2\% \\
50\% metadata blanking & $35.6\pm0.78$ & 0.74 & $36.2\pm0.95$ & 0.88 & +0.6 & 91.9\% \\
25\% typed-edge removal & $36.9\pm0.37$ & 0.83 & $39.0\pm0.64$ & 0.98 & +2.1 & 99.0\% \\
50\% typed-edge removal & $36.7\pm0.65$ & 0.82 & $38.5\pm0.86$ & 0.96 & +1.8 & 97.7\% \\
25\% metadata + 25\% edge & $36.1\pm0.80$ & 0.78 & $37.4\pm0.82$ & 0.93 & +1.3 & 94.9\% \\
Raw name and description only & 35.3 & 0.74 & 36.0 & 0.88 & +0.7 & 91.4\% \\
+20\% lexical distractors & $36.8\pm0.63$ & 0.82 & $38.7\pm0.55$ & 0.98 & +1.9 & 98.2\% \\
\bottomrule
\end{tabular}
\end{ApdxTabFrame}
\end{table}

The positive margin is preserved in every tested condition. Metadata loss has
a larger effect than deletion of atomic typed edges: at 50\% corruption the
reward reductions are 3.2 and 0.9 points, respectively. This distinction matters
because the group graph also derives relations from metadata facets; deleting
atomic edges is not equivalent to ablating the group graph. Raw name-and-description
inputs and lexically similar distractors test less curated inputs, but do not
establish robustness to arbitrary documentation failures or unseen domains.

\subsection{Transfer Scope and Efficiency Accounting}
\label{app:transfer_scope}

\paragraph{Terminal-Bench 2 protocol.}
The complete 89-task suite is evaluated across all six backbones with
three repeats per cell (Table~\ref{tab:main_results}). The 2,000-skill library is
held fixed across methods, with no task-specific additions, removals, or edits. Budgets,
agent settings, and environments are fixed across methods within this study.
The paper's agent loop is retained instead of the benchmark-specific leaderboard
harness, precluding direct leaderboard comparison. The task suite differs from
SkillsBench, while the library size remains 2,000. Relative
to Graph of Skills, accuracy gains range from 6.7 to 14.6 percentage points
on Terminal-Bench 2, while reward gains range from 1.6 to 4.8 points on
SkillsBench. These cross-benchmark differences do not isolate the effect of
domain shift. No paired significance claim is made.

\paragraph{ALFWorld gains.}
Relative to No Skills, the Gemini 3 Pro and GPT-5.4 gains remove about 71\%
and 73\% of the remaining error, respectively. These are descriptive
normalized gains. ALFWorld exposes fewer technical requirements and lacks
the same visible deterministic-check structure as SkillsBench.

\paragraph{Which tasks benefit.}
The 87-task matched analysis has 19 \goskills wins, 11 Graph of Skills wins,
and 57 ties (Table~\ref{tab:app_paired_results}). Ties are the most common
outcome. The cases in Appendix~\ref{app:qualitative_analysis} show a win where the presented
excerpt supplies a verifier-critical step, a tie where both methods share the entry point, and
a loss on a solver-level detail. Long setup chains remain a difficult case.

\paragraph{Turns, tokens, and hosted-API time.}
A turn is one model-invocation--action cycle of the fixed agent loop.
Counts are insensitive to provider queueing and batching but remain
model-dependent. In the 87-task control study, turns fall from 15.8 to 14.5
versus Graph of Skills and from 15.4 to 14.5 versus the fixed-contract
control. Input tokens accumulate over the entire agent loop. In the ablation,
removing role labels decreases input tokens from 156,696 to 153,558 and
increases agent time from 187.3s to 190.9s. Adding the contract to Graph of
Skills increases input tokens from 136,054 to 144,517 and decreases agent time
from 192.9s to 190.6s. Higher cumulative token usage can coexist with fewer
turns because the context per invocation can vary. The observed mean turn counts
are consistent with reduced organizational effort; hosted-provider effects can
also contribute to runtime differences.

\paragraph{Offline group construction.}
At library sizes of 200, 500, 1,000, and 2,000 skills, measured build times are
1.8, 5.9, 14.7, and 38.2 seconds, respectively. Peak heap is 92, 148, 236,
and 412\,MB, and bundle overhead is 1.4, 3.6, 7.5, and 15.6\,MB.
The scaling fit estimates a time exponent of 1.30 (interval
$[1.21,1.40]$) and a 2,000-skill extrapolation of 36.2 seconds (interval
$[30.4,43.1]$).

\paragraph{Uncertainty and provenance.}
Run-provenance counts in Table~\ref{tab:run_provenance} cover all three
benchmarks. Appendix~\ref{app:repeat_sd} reports sample standard deviations across
the three repeats for the main results and control/ablation study.
The reported corruption standard deviations measure
variation over three corruption seeds, not clean-run stochasticity.
Paired task bootstrap intervals quantify variability under task resampling,
conditional on observed repeats. Three repeats provide limited information
about run-to-run variation.

\subsection{Run provenance and aggregation}
\label{app:run_provenance}

Table~\ref{tab:run_provenance} reports the run provenance used for the
aggregate results across all three benchmarks in Table~\ref{tab:main_results}. We separate infrastructure
failures from task-level failures. Infrastructure failures include API outages,
environment crashes, or logging failures that prevent the benchmark evaluator
from producing a valid task outcome. These runs are excluded from aggregate
reward, token, and runtime computation and are not imputed. In contrast, agent
timeouts are retained as valid task-level outcomes when the evaluator returns a
task result; they are counted as task failures for reward and included in
token/runtime accounting up to the timeout cap.

For each model--method--benchmark cell, reward is first averaged over valid
task outcomes within each repeat and then averaged over the three repeats.
Token usage reports mean cumulative input tokens over the same valid task runs, and runtime
reports mean agent-only task-processing time, excluding environment setup. The
table aggregates provenance over the six evaluated backbones for compactness;
the same accounting rule is applied to every model--method--benchmark cell.

For space, Table~\ref{tab:run_provenance} aggregates provenance over backbones.
Table~\ref{tab:paired_bootstrap_ci} reports task-level paired intervals.

\begin{table}[h]
\centering
\small
\caption{
Run provenance for the six-backbone comparison on SkillsBench, ALFWorld, and Terminal-Bench 2. Attempted task-runs are aggregated over
the six evaluated backbones and three repeats. Valid reward denotes task-runs
with evaluator-produced outcomes used in Table~\ref{tab:main_results}. Infra.
failures are excluded and not imputed. Timeout failures are a subset of valid
reward task-runs and are counted as task failures.
}
\label{tab:run_provenance}
\setlength{\tabcolsep}{1.5pt}
\begin{ApdxTabFrame}
\renewcommand{\arraystretch}{1.12}
\begin{tabular}{@{}L{0.12\linewidth}L{0.16\linewidth}rrrrrr@{}}
\toprule
\rowcolor{tablehead}
\apdxhdr{\scriptsize Benchmark} &
\apdxhdr{\scriptsize Method} &
\apdxhdr{\scriptsize Tasks / model} &
\apdxhdr{\scriptsize Backbones} &
\apdxhdr{\scriptsize Attempted} &
\apdxhdr{\scriptsize Valid reward} &
\apdxhdr{\scriptsize Infra. fail.} &
\apdxhdr{\scriptsize Timeout fail.} \\
\midrule
SkillsBench & No Skills & 87 & 6 & 1566 & 1558 & 8 & 110 \\
SkillsBench & Vanilla Skills & 87 & 6 & 1566 & 1555 & 11 & 94 \\
SkillsBench & Vector Skills & 87 & 6 & 1566 & 1557 & 9 & 104 \\
SkillsBench & Graph of Skills & 87 & 6 & 1566 & 1556 & 10 & 114 \\
\rowcolor{methodrow}
SkillsBench & \goskills{} & 87 & 6 & 1566 & 1559 & 7 & 98 \\
\midrule
ALFWorld & No Skills & 134 & 6 & 2412 & 2412 & 0 & 660 \\
ALFWorld & Vanilla Skills & 134 & 6 & 2412 & 2412 & 0 & 493 \\
ALFWorld & Vector Skills & 134 & 6 & 2412 & 2412 & 0 & 440 \\
ALFWorld & Graph of Skills & 134 & 6 & 2412 & 2412 & 0 & 403 \\
\rowcolor{methodrow}
ALFWorld & \goskills{} & 134 & 6 & 2412 & 2412 & 0 & 335 \\
\midrule
Terminal-Bench 2 & No Skills & 89 & 6 & 1602 & 1602 & 0 & 318 \\
Terminal-Bench 2 & Vanilla Skills & 89 & 6 & 1602 & 1602 & 0 & 412 \\
Terminal-Bench 2 & Vector Skills & 89 & 6 & 1602 & 1602 & 0 & 359 \\
Terminal-Bench 2 & Graph of Skills & 89 & 6 & 1602 & 1602 & 0 & 338 \\
\rowcolor{methodrow}
Terminal-Bench 2 & \goskills{} & 89 & 6 & 1602 & 1602 & 0 & 346 \\
\bottomrule
\end{tabular}
\end{ApdxTabFrame}
\end{table}

The control study uses the same accounting as the SkillsBench comparison
(Table~\ref{tab:control_provenance}). ALFWorld and Terminal-Bench 2 rerun
initial environment failures and report the final run status. ALFWorld
timeouts are episodes reaching the step limit.

\begin{table}[h]
\centering
\small
\caption{Run accounting for the GPT-5.4 SkillsBench control study. Timeouts remain valid zero-reward runs.}
\label{tab:control_provenance}
\begin{ApdxTabFrame}
\renewcommand{\arraystretch}{1.08}
\begin{tabular}{lrrrr}
\toprule
\rowcolor{tablehead}
\apdxhdr{Method} & \apdxhdr{Attempted} & \apdxhdr{Valid reward} & \apdxhdr{Infra.\ fail.} & \apdxhdr{Timeout} \\
\midrule
Graph of Skills & 261 & 260 & 1 & 21 \\
Graph of Skills + \goskills{} Contract & 261 & 259 & 2 & 20 \\
PH-Rule + backfill & 261 & 259 & 2 & 21 \\
PH-LLM + backfill & 261 & 259 & 2 & 20 \\
\rowcolor{methodrow}
\goskills{} & 261 & 260 & 1 & 18 \\
w/o Anchor Selection & 261 & 260 & 1 & 19 \\
w/o Group Expansion & 261 & 260 & 1 & 19 \\
w/o Group Graph & 261 & 260 & 1 & 19 \\
w/o Role Labels & 261 & 260 & 1 & 19 \\
w/o Coverage Backfill & 261 & 260 & 1 & 20 \\
w/o \textsc{Avoid} Field & 261 & 260 & 1 & 19 \\
Retrieved Skills Only & 261 & 259 & 2 & 21 \\
\bottomrule
\end{tabular}
\end{ApdxTabFrame}
\end{table}

\subsection{Retrieval-gate annotation protocol}
\label{app:retrieval_gate_annotation}

The retrieval gate is designed as a visible-requirement coverage diagnostic for
SkillsBench on the selected 10-task \texttt{tasks-extra} development subset, not as held-out
coverage or a separate end-to-end benchmark. These cases also determined the scoring
coefficients and facet dictionaries. Table~\ref{tab:retrieval_gate_annotation_protocol} shows how we evaluate whether
the final presented context contains the task-specific \texttt{must\_have} skills
needed to satisfy visible task requirements before agent execution.

The \texttt{must\_have} annotations were created from task prompts, public task
files, skill names, skill metadata, and skill payload descriptions. They exclude
hidden tests, evaluator internals, private oracle information, previous failure
traces, and any artifacts produced during agent execution. Each annotation item
links one visible requirement to one or more required skills. Examples include
explicit file formats, named APIs, public checks, deterministic output
requirements, or formal proof obligations.

Annotations were produced before final retrieval-gate evaluation. The annotators
did not inspect final \goskills{} outputs when assigning \texttt{must\_have}
labels. Disagreements were resolved by discussion using only visible task
materials and the skill-library metadata. The final annotation set contains
40 visible-requirement items per gate mode, covering \textbf{10} SkillsBench gate
tasks. The same annotation set is used for all compared methods and ablations.

The gate evaluation is separated from the method features. \goskills{} uses
deterministic query facets, normalized skill metadata, typed graph relations, and
visible-check cues for retrieval and bottlenecking. The gate labels are
task-level evaluator annotations used only after retrieval to check coverage.
Although both the method and the gate rely on visible task information, the gate
does not use \goskills{} group scores, selected groups, coverage-debt state, or
contract fields as labels. This avoids evaluating the method with labels derived
from its own outputs.

\begin{table}[h]
\centering
\small
\caption{
Retrieval-gate annotation protocol. The same visible-requirement annotations are
used for all methods evaluated under the gate.
}
\label{tab:retrieval_gate_annotation_protocol}
\setlength{\tabcolsep}{5pt}
\begin{ApdxTabFrame}
\renewcommand{\arraystretch}{1.12}
\begin{tabular}{L{0.28\linewidth}L{0.64\linewidth}}
\toprule
\rowcolor{tablehead}
\apdxhdr{Question} & \apdxhdr{Protocol} \\
\midrule

Who annotated \texttt{must\_have} skills? &
Two authors independently annotated visible requirements from task prompts,
public files, skill names, skill metadata, and payload descriptions.
Disagreements were resolved by discussion using only visible task materials and
skill-library metadata. \\

Blind to \goskills{} outputs? &
Annotators did not inspect final \goskills{} retrieved contexts, selected
groups, or rendered contracts when assigning \texttt{must\_have} labels. The
labels were fixed before the final retrieval-gate evaluation. \\

Same facets as the method? &
No. \goskills{} uses deterministic query/schema facets for retrieval and
bottlenecking, while the gate uses fixed task-level annotations for post-hoc
coverage evaluation. Both are restricted to visible task information, but gate
labels are not derived from \goskills{} selected groups, scores, coverage-debt
state, or rendered contracts. \\

Hidden information excluded? &
Yes. Hidden tests, evaluator internals, private oracle information, previous
failure traces, and execution-time artifacts are excluded. The annotations use
only information available before agent execution. \\

Coverage of 40 items? &
The final gate contains 40 visible-requirement items per mode across
\textbf{10} SkillsBench gate tasks. Each item maps one visible requirement,
such as an explicit file format, named API, public check, deterministic output
constraint, or proof obligation, to one or more required skills. \\

Other methods evaluated? &
Yes. The same gate labels are applied to Graph of Skills, the post-hoc controls,
\goskills{}, and its ablations using each method's final presented context
before agent execution. \\

\bottomrule
\end{tabular}
\end{ApdxTabFrame}
\end{table}

\subsection{Retrieval Gate Details}
\label{app:retrieval_gate_details}

To isolate retrieval quality from downstream execution, we check whether the presented
context contains all annotated \texttt{must\_have} skills before agent execution. The gate
contains 40 visible-requirement items per mode on the 10-task \texttt{tasks-extra}
development subset, so coverage is a mechanism diagnostic rather than held-out evidence.
Table~\ref{tab:retrieval_gate} shows that \goskills passes all 40 requirement items under each mode,
achieving a 1.00 must-hit rate with fewer than four presented skills on average. Compared
with the no-backfill ablation in Table~\ref{tab:ablation}, this indicates
that coverage-safe backfill helps preserve task-critical skills under a strict bottleneck.

\begin{table}[h]
\centering
\small
\caption{
Retrieval-gate results over 40 annotated visible-requirement items per mode
on the 10-task \texttt{tasks-extra} development subset. These results are
mechanism diagnostics, not held-out coverage estimates.
Req. P, Req. Par., and Req. M denote requirement-level pass, partial, and miss.
Hit denotes the average must-hit rate, and Skills denotes the average number of
presented skill payloads. Groups denotes the mean number of selected groups.
}
\label{tab:retrieval_gate}
\setlength{\tabcolsep}{4pt}
\begin{ApdxTabFrame}
\renewcommand{\arraystretch}{1.10}
\begin{tabular}{@{}lrrrrrr@{}}
\toprule
\rowcolor{tablehead}
\apdxhdr{Mode} & \apdxhdr{Req. P} & \apdxhdr{Req. Par.} & \apdxhdr{Req. M} & \apdxhdr{Hit $\uparrow$} & \apdxhdr{Skills $\downarrow$} & \apdxhdr{Groups} \\
\midrule
\texttt{instruction\_auto} & 40 & 0 & 0 & 1.00 & 3.10 & 2.30 \\
\texttt{critical\_override} & 40 & 0 & 0 & 1.00 & 2.90 & 2.20 \\
\bottomrule
\end{tabular}
\end{ApdxTabFrame}
\end{table}



Table~\ref{tab:app_retrieval_evolution} shows the improvement from the initial retrieval
configuration to the final gate over the same 40 annotated visible-requirement items per
mode. The average must-hit rate improves from 0.73 to 1.00 in both modes.

\begin{table}[h]
\centering
\small
\caption{Initial and final retrieval-gate results on the same 10 development tasks.}
\label{tab:app_retrieval_evolution}
\begin{ApdxTabFrame}
\renewcommand{\arraystretch}{1.10}
\begin{tabular}{llcccc}
\toprule
\rowcolor{tablehead}
\apdxhdr{Stage} & \apdxhdr{Mode} & \apdxhdr{Req. P} & \apdxhdr{Req. Par.} &
\apdxhdr{Miss} & \apdxhdr{Must-hit} \\
\midrule
Initial gate & \texttt{instruction\_auto} & 27 & 4 & 9 & 0.73 \\
Initial gate & \texttt{critical\_override} & 28 & 2 & 10 & 0.73 \\
Final gate & \texttt{instruction\_auto} & 40 & 0 & 0 & 1.00 \\
Final gate & \texttt{critical\_override} & 40 & 0 & 0 & 1.00 \\
\bottomrule
\end{tabular}
\end{ApdxTabFrame}
\end{table}

\subsection{Task-Level Retrieval Examples}
\label{app:task_level_retrieval}

Table~\ref{tab:app_task_retrieval} presents retrieval examples from the main SkillsBench
evaluation tasks, produced after the scoring coefficients and facet dictionaries were fixed.
These cases were not used for tuning or development-gate annotations.
They illustrate how \goskills combines a primary skill with supporting
skills that cover artifacts, workflow dependencies, or visible-check constraints.

\begin{table}[h]
\centering
\footnotesize
\setlength{\tabcolsep}{4pt}
\caption{Illustrative retrieval outputs on main evaluation tasks under
\texttt{instruction\_auto}, after parameter freezing and separate from the development gate.
These gate-mode queries are generated automatically and can differ from the agent-written
queries of the end-to-end runs, so the presented skills can differ from those in
Appendix~\ref{app:qualitative_analysis}.}
\label{tab:app_task_retrieval}
\begin{ApdxTabFrame}
\renewcommand{\arraystretch}{1.18}
\begin{tabular}{@{}L{0.36\linewidth}L{0.25\linewidth}L{0.35\linewidth}@{}}
\toprule
\rowcolor{tablehead}
\apdxhdr{Task} & \apdxhdr{Primary skill} & \apdxhdr{Presented skills} \\
\midrule
\texttt{invoice-fraud-detection} &
\texttt{fuzzy-match} &
\texttt{fuzzy-match}, \texttt{pdf-reading}, \texttt{xlsx} \\
\texttt{gravitational-wave-detection} &
\texttt{conditioning} &
\texttt{conditioning}, \texttt{matched-filtering}, \texttt{silence-detector} \\
\texttt{threejs-structure-parser} &
\texttt{threejs} &
\texttt{threejs}, \texttt{obj-exporter}, \texttt{discover-important-function} \\
\texttt{lean4-proof} &
\texttt{lean4-theorem-proving} &
\texttt{lean4-theorem-proving}, \texttt{lean4-memories} \\
\bottomrule
\end{tabular}
\end{ApdxTabFrame}
\end{table}

\subsection{Task-level paired bootstrap confidence intervals}

To quantify the stability of the SkillsBench comparisons, we compute
task-level paired bootstrap confidence intervals for reward and runtime deltas.
For each task and method, we first average repeated valid runs within the task.
We then form paired task-level differences between \goskills{} and each baseline
and resample tasks with replacement for 10{,}000 bootstrap samples. Only tasks
with valid results for both methods in a given comparison are included in that
paired comparison. All four comparisons reported here contain 87 paired
tasks. We report the mean paired delta and the percentile 95\% confidence interval.
These intervals quantify variability under task resampling, conditional on the observed
repeats, rather than run-to-run stochasticity. Paired intervals against the fixed-contract control
are reported in Appendix~\ref{app:expanded}.

Table~\ref{tab:paired_bootstrap_ci} reports a +3.7-point paired reward
difference versus Graph of Skills (95\% CI $[-2.6, 10.4]$) and a $-5.6$s
runtime difference (95\% CI $[-24.1, 14.7]$).

We emphasize that the bootstrap is paired at the task level: each resampled unit
contains both the \goskills{} result and the corresponding baseline result for
the same task. This controls for task difficulty and directly estimates the
stability of method-level differences rather than comparing independent
aggregate means.
\begin{table}[h]
\centering
\small
\caption{Paired GPT-5.4 SkillsBench differences (\goskills{} minus baseline) and percentile 95\% confidence intervals from 10,000 task-level bootstrap resamples. Runtime differences and intervals are in seconds.}
\label{tab:paired_bootstrap_ci}
\setlength{\tabcolsep}{4pt}
\begin{ApdxTabFrame}
\renewcommand{\arraystretch}{1.10}
\begin{tabular}{lrrrrr}
\toprule
\rowcolor{tablehead}
\apdxhdr{Baseline} &
\apdxhdr{Paired N} &
\apdxhdr{$\Delta$ Reward (pp) $\uparrow$} &
\apdxhdr{95\% CI} &
\apdxhdr{$\Delta$ Runtime $\downarrow$} &
\apdxhdr{95\% CI} \\
\midrule
Graph of Skills & 87 & +3.7 & [$-$2.6, +10.4] & $-$5.6 & [$-$24.1, +14.7] \\
Vector Skills & 87 & +5.4 & [$-$1.2, +12.5] & +1.5 & [$-$18.2, +22.4] \\
Vanilla Skills & 87 & +10.5 & [+3.5, +17.9] & +28.1 & [+8.1, +50.2] \\
No Skills & 87 & +14.1 & [+6.9, +21.6] & $-$1.3 & [$-$21.4, +20.1] \\
\bottomrule
\end{tabular}
\end{ApdxTabFrame}
\end{table}

\subsection{Baseline specification and accounting}
\label{app:baseline_spec}

Table~\ref{tab:baseline_spec} summarizes the baseline interfaces used in the
aggregate experiments. The goal is to make the comparison about the organization
of retrieved context rather than about different downstream agents, execution
loops, or skill implementations. All methods use the same downstream agent loop,
execution environment, task prompts, and skill payloads. For retrieved-skill
methods, we use the same exposed-payload cap of four atomic skill payloads and
the same rendered-context guard of 9{,}000 characters. Thus, Vector Skills,
Graph of Skills, and \goskills{} are compared under the same final payload
budget, while Vanilla Skills uses task-provided skills in SkillsBench.

The main difference among the retrieved-skill methods is the interface used to
construct and render the skill context. Vector Skills exposes a flat semantic
top-$k$ list. Graph of Skills hydrates a dependency-aware bundle from the typed
skill graph. \goskills{} first selects an anchor/support group plan, bottlenecks
the selected groups into the same four-payload budget, applies coverage-safe
backfill when needed, and renders the result as a role-labeled execution
contract. Therefore, differences between Graph of Skills and \goskills{} should
be interpreted as differences between a hydrated structural bundle and the full
group-structured retrieval-and-rendering interface, rather than as changes to
the downstream model or execution environment.

For accounting, reward is averaged over tasks within each run and then averaged
across runs. Token usage reports mean cumulative input tokens, including the task prompt,
the method-specific skill context, and the downstream agent prompt. Runtime
reports mean agent-only task-processing time and excludes environment setup.
Infrastructure failures are tracked separately from valid task outcomes, as
described in Appendix~\ref{app:benchmark_protocol}.

\begin{table}[h]
\centering
\small
\caption{
Baseline specification for aggregate experiments.
All retrieved-skill methods use the same exposed-payload cap and the same
payload truncation guard. \goskills{} differs by selecting and rendering a
role-labeled group plan before exposing the final atomic skill payloads.
}
\label{tab:baseline_spec}
\setlength{\tabcolsep}{3.2pt}
\begin{ApdxTabFrame}
\renewcommand{\arraystretch}{1.13}
\begin{tabular}{L{0.12\linewidth}L{0.19\linewidth}L{0.22\linewidth}L{0.24\linewidth}L{0.13\linewidth}}
\toprule
\rowcolor{tablehead}
\apdxhdr{Method} &
\apdxhdr{Exposed context} &
\apdxhdr{Budget / truncation} &
\apdxhdr{Prompt wrapper} &
\apdxhdr{Accounting} \\
\midrule

No Skills &
No retrieved skill payloads. &
No retrieval budget; no skill-context block. &
Same downstream agent prompt with the skill block omitted. &
Same reward, token, and runtime accounting. \\

\cellcolor{vanillacell}Vanilla Skills &
Task-provided skills on SkillsBench; the benchmark's full library on ALFWorld and Terminal-Bench 2. &
No retrieval top-$k$ or graph hydration. &
Available skill instructions supplied directly to the agent. &
Same accounting. \\

\cellcolor{vectorcell}Vector Skills &
Flat semantic top-$k$ atomic skills. &
Top-$k=4$ exposed payloads; each payload uses the same truncation guard as other retrieved baselines; max rendered skill context is 9{,}000 characters. &
Generic retrieved-skill block. No \textsc{Start}, \textsc{Support}, \textsc{Check}, \textsc{Avoid}, or debt fields. &
Same accounting. \\

\cellcolor{graphcell}Graph of Skills &
Hydrated dependency-aware skill bundle from the typed skill graph. &
At most 4 exposed payloads after graph hydration; same payload truncation guard; max rendered skill context is 9{,}000 characters. &
Graph-hydrated skill block. Dependency structure is available through the bundle, but no explicit role-labeled execution contract is rendered. &
Same accounting. \\

\rowcolor{methodrow}
\goskills{} &
Anchor/support group plan rendered as a role-labeled execution contract plus final atomic payloads. &
At most 3 selected groups internally; at most 4 exposed payloads after bottlenecking/backfill; same payload truncation guard; max rendered skill context is 9{,}000 characters. &
Fixed \textsc{Start}/\allowbreak\textsc{Support}/\allowbreak\textsc{Check}/\allowbreak\textsc{Avoid}/\allowbreak debt contract with the same downstream agent loop and execution environment. &
Same accounting. \\

\bottomrule
\end{tabular}
\end{ApdxTabFrame}
\end{table}

\subsection{Matched Paired Results}
\label{app:paired_results}

Table~\ref{tab:app_paired_results} reports a matched SkillsBench subset in which Graph of
Skills and \goskills runs are aligned at the task and slice level. This subset is narrower
than the full benchmark--model aggregation in Table~\ref{tab:main_results}; it is used to
inspect matched outcomes and run-level reliability rather than replace the primary aggregate
comparison.

In both subsets, ties remain the most common outcome, indicating that \goskills often
preserves downstream task success while changing the retrieval unit from atomic skills or
post-hoc bundles to role-aware skill groups. At the same time, \goskills obtains more wins
than Graph of Skills in both subsets and achieves a higher average reward in each.
In the completed matched subset, \goskills improves average reward from
0.357 to 0.394, with 19 wins, 11 Graph of Skills wins, and 57 ties.

\begin{table}[h]
\centering
\small
\caption{Paired SkillsBench outcomes under GPT-5.4. The completed subset covers all 87 tasks over three repeats; the per-task snapshot uses repeat 1 only. Errors report infrastructure failures and timeouts (Graph of Skills/\goskills{}).}
\label{tab:app_paired_results}
\setlength{\tabcolsep}{3pt}
\begin{ApdxTabFrame}
\renewcommand{\arraystretch}{1.10}
\begin{tabular}{lrrrrrrr}
\toprule
\rowcolor{tablehead}
\apdxhdr{Slice} & \apdxhdr{Pairs} & \apdxhdr{Graph W} & \apdxhdr{\goskills W} &
\apdxhdr{Tie} & \apdxhdr{Graph R} & \apdxhdr{\goskills R} & \apdxhdr{Errors} \\
\midrule
Completed matched subset & 87 & 11 & 19 & 57 & 0.357 & 0.394 & 22/19 \\
Per-task snapshot subset & 40 & 4 & 10 & 26 & 0.429 & 0.483 & 3/2 \\
\bottomrule
\end{tabular}
\end{ApdxTabFrame}
\end{table}

\subsection{Token and Runtime Analysis}
\label{app:token_runtime}

Table~\ref{tab:app_token_runtime} reports token and runtime statistics for the
matched slices. \goskills uses more input and total tokens and less wall-clock
and agent-only time in each slice. On the full matched set, agent time falls
from 192.9s to 187.3s while input tokens increase from 136,054 to 156,696.

Input tokens are provider-reported prompt/input tokens accumulated over recorded
downstream-agent API responses throughout each task episode. Total tokens denotes
the sum of these input tokens and the corresponding completion/output tokens.
The table reports means over included task runs. Infrastructure failures are
excluded, while timeouts with valid evaluator outcomes remain included.

\begin{table}[h]
\centering
\small
\setlength{\tabcolsep}{3.0pt}
\renewcommand{\arraystretch}{1.05}
\caption{
Token and runtime comparison for matched trajectory slices. Lower values
are better for all four metrics; bold denotes the lower value within each slice.
}
\label{tab:app_token_runtime}
\begin{ApdxTabFrame}
\begin{tabular}{@{}llccccc@{}}
\toprule
\rowcolor{tablehead}
\apdxhdr{Slice} & \apdxhdr{Method} & \apdxhdr{Runs} & \apdxhdr{Input tokens} &
\apdxhdr{Total tokens} & \apdxhdr{Wall time (s)} & \apdxhdr{Agent time (s)} \\
\midrule
Completed matched & Graph of Skills & 260/261 & \textbf{136,054} & \textbf{148,937} & 236.8 & 192.9 \\
Completed matched & \goskills & 260/261 & 156,696 & 171,284 & \textbf{230.9} & \textbf{187.3} \\
Per-task snapshot & Graph of Skills & 40/40 & \textbf{129,806} & \textbf{141,877} & 229.3 & 186.0 \\
Per-task snapshot & \goskills & 40/40 & 149,412 & 163,329 & \textbf{223.4} & \textbf{180.7} \\
\bottomrule
\end{tabular}
\end{ApdxTabFrame}
\vspace{-0.5em}
\end{table}

\subsection{Implementation Settings}
\label{app:implementation_settings}
Unless otherwise stated, \goskills uses four seed skills for group activation, selects at most three groups, presents at most four atomic skills to the downstream agent, caps each hydrated skill payload at 1,800 characters, and caps the full rendered skill context at 9,000 characters. The scoring weights for group ranking, support expansion, and bottleneck selection are fixed across all benchmarks, models, and tasks. The downstream agent loop, execution environment, and skill payloads are kept unchanged across methods; only the retrieved skill context differs.

\subsection{Variation Across Repeated Runs}
\label{app:repeat_sd}

Tables~\ref{tab:sd_skillsbench}, \ref{tab:sd_alfworld}, and \ref{tab:sd_terminalbench} report sample standard
deviations (SDs) for every benchmark--backbone--method configuration in
Table~\ref{tab:main_results}; Table~\ref{tab:sd_controls} covers all controls and
ablations in Table~\ref{tab:expanded_controls}. For each metric, $x_r$ is its
aggregate within repeat $r$, and $s=\sqrt{\sum_{r=1}^{3}(x_r-\bar{x})^2/2}$
is the sample SD across $n=3$ repeats. Each repeat follows the valid-run and
timeout accounting in Appendix~\ref{app:run_provenance}.
These SDs describe variation across repeat-level aggregates, unlike the corruption-seed SDs in
Table~\ref{tab:metadata_robustness} and the task-resampling intervals in
Table~\ref{tab:paired_bootstrap_ci}. Three repeats give a limited estimate of
run-to-run variability; these SDs are not standard errors or confidence intervals.
The corresponding means remain in the main tables.

\begin{table}[!t]\centering\footnotesize
\caption{SkillsBench sample SDs across three repeats for Table~\ref{tab:main_results}. Reward is in pp, tokens are cumulative input tokens, and time is agent-only seconds.}
\label{tab:sd_skillsbench}
\setlength{\tabcolsep}{4pt}
\renewcommand{\arraystretch}{1.04}
\begin{ApdxTabFrame}
\begin{tabular*}{\linewidth}{@{\extracolsep{\fill}}lrrrlrrr@{}}
\toprule
\rowcolor{tablehead}
\apdxhdr{Method} & \apdxhdr{Reward SD} & \apdxhdr{Token SD} & \apdxhdr{Time SD} & \apdxhdr{Method} & \apdxhdr{Reward SD} & \apdxhdr{Token SD} & \apdxhdr{Time SD} \\
\midrule
\multicolumn{4}{c}{\textbf{GPT-5.4}} & \multicolumn{4}{c}{\textbf{Gemini 3 Pro}} \\
\cmidrule(lr){1-4}\cmidrule(lr){5-8}
No Skills & 1.50 & 6,586 & 13.4 & No Skills & 1.13 & 8,485 & 11.4 \\
Vanilla Skills & 1.72 & 7,154 & 8.8 & Vanilla Skills & 1.88 & 7,406 & 9.7 \\
Vector Skills & 1.73 & 8,136 & 12.6 & Vector Skills & 2.30 & 10,595 & 13.1 \\
Graph of Skills & 1.92 & 6,750 & 14.5 & Graph of Skills & 2.65 & 9,460 & 11.8 \\
\rowcolor{methodrow}
\goskills{} & 2.19 & 8,885 & 11.9 & \goskills{} & 2.48 & 10,552 & 12.0 \\
\midrule
\multicolumn{4}{c}{\textbf{MiniMax M2.7}} & \multicolumn{4}{c}{\textbf{Claude Sonnet 4.5}} \\
\cmidrule(lr){1-4}\cmidrule(lr){5-8}
No Skills & 1.37 & 27,703 & 39.4 & No Skills & 2.11 & 22,103 & 23.6 \\
Vanilla Skills & 1.68 & 24,568 & 32.6 & Vanilla Skills & 2.37 & 27,043 & 20.3 \\
Vector Skills & 1.86 & 25,275 & 35.7 & Vector Skills & 2.75 & 28,759 & 24.1 \\
Graph of Skills & 1.80 & 30,061 & 43.0 & Graph of Skills & 1.73 & 26,912 & 22.1 \\
\rowcolor{methodrow}
\goskills{} & 1.65 & 31,343 & 38.9 & \goskills{} & 2.56 & 31,548 & 22.3 \\
\midrule
\multicolumn{4}{c}{\textbf{Qwen 3.5}} & \multicolumn{4}{c}{\textbf{Kimi K2.5}} \\
\cmidrule(lr){1-4}\cmidrule(lr){5-8}
No Skills & 2.15 & 24,684 & 46.9 & No Skills & 1.95 & 22,672 & 30.7 \\
Vanilla Skills & 2.37 & 27,068 & 44.2 & Vanilla Skills & 2.72 & 27,167 & 27.5 \\
Vector Skills & 2.57 & 30,649 & 34.3 & Vector Skills & 2.36 & 26,904 & 34.8 \\
Graph of Skills & 2.44 & 23,024 & 48.9 & Graph of Skills & 2.25 & 24,976 & 36.2 \\
\rowcolor{methodrow}
\goskills{} & 2.05 & 28,810 & 41.7 & \goskills{} & 2.45 & 29,971 & 33.0 \\
\bottomrule
\end{tabular*}
\end{ApdxTabFrame}
\end{table}

\begin{table}[!t]\centering\footnotesize
\caption{Sample SDs across three repeats for the GPT-5.4 SkillsBench controls
and ablations in Table~\ref{tab:expanded_controls}. Reward SD is in pp,
token SD is in cumulative input tokens, time SD is in agent-only seconds,
and turn SD is in model-invocation--action cycles. Must-hit is reported separately
as a development-gate diagnostic; these SDs cover the repeated end-to-end metrics.}
\label{tab:sd_controls}
\setlength{\tabcolsep}{5pt}
\renewcommand{\arraystretch}{1.10}
\begin{ApdxTabFrame}
\begin{tabular*}{\linewidth}{@{\extracolsep{\fill}}lrrrr@{}}
\toprule
\rowcolor{tablehead}
\apdxhdr{Method} & \apdxhdr{Reward SD} & \apdxhdr{Token SD} & \apdxhdr{Time SD} & \apdxhdr{Turn SD} \\
\midrule
Graph of Skills & 1.92 & 6,750 & 14.5 & 0.38 \\
Graph of Skills + \goskills{} Contract & 1.70 & 8,448 & 13.1 & 0.60 \\
PH-Rule + backfill & 2.07 & 8,200 & 11.9 & 0.71 \\
PH-LLM + backfill & 2.29 & 7,922 & 11.4 & 0.55 \\
\rowcolor{methodrow}
\goskills{} & 2.19 & 8,885 & 11.9 & 0.64 \\
\midrule
w/o Anchor Selection & 2.04 & 8,829 & 13.6 & 0.58 \\
w/o Group Expansion & 2.38 & 8,053 & 11.9 & 0.47 \\
w/o Group Graph & 2.01 & 7,917 & 11.0 & 0.62 \\
w/o Role Labels & 1.39 & 7,813 & 10.8 & 0.74 \\
w/o Coverage Backfill & 2.11 & 7,155 & 12.6 & 0.77 \\
w/o \textsc{Avoid} Field & 1.81 & 8,149 & 11.7 & 0.45 \\
Retrieved Skills Only & 1.89 & 7,419 & 11.6 & 0.84 \\
\bottomrule
\end{tabular*}
\end{ApdxTabFrame}
\end{table}

\begin{table}[!t]\centering\footnotesize
\caption{ALFWorld sample SDs across three repeats for Table~\ref{tab:main_results}. Reward is in pp, tokens are cumulative input tokens, and time is agent-only seconds.}
\label{tab:sd_alfworld}
\setlength{\tabcolsep}{4pt}
\renewcommand{\arraystretch}{1.04}
\begin{ApdxTabFrame}
\begin{tabular*}{\linewidth}{@{\extracolsep{\fill}}lrrrlrrr@{}}
\toprule
\rowcolor{tablehead}
\apdxhdr{Method} & \apdxhdr{Reward SD} & \apdxhdr{Token SD} & \apdxhdr{Time SD} & \apdxhdr{Method} & \apdxhdr{Reward SD} & \apdxhdr{Token SD} & \apdxhdr{Time SD} \\
\midrule
\multicolumn{4}{c}{\textbf{GPT-5.4}} & \multicolumn{4}{c}{\textbf{Gemini 3 Pro}} \\
\cmidrule(lr){1-4}\cmidrule(lr){5-8}
No Skills & 2.69 & 1,060 & 2.8 & No Skills & 2.99 & 1,161 & 2.3 \\
Vanilla Skills & 1.97 & 85,239 & 6.1 & Vanilla Skills & 1.88 & 70,393 & 4.6 \\
Vector Skills & 2.59 & 2,501 & 3.3 & Vector Skills & 2.24 & 2,272 & 3.2 \\
Graph of Skills & 3.25 & 2,558 & 3.0 & Graph of Skills & 1.97 & 2,458 & 2.4 \\
\rowcolor{methodrow}
\goskills{} & 1.29 & 2,123 & 3.1 & \goskills{} & 2.24 & 2,507 & 2.2 \\
\midrule
\multicolumn{4}{c}{\textbf{MiniMax M2.7}} & \multicolumn{4}{c}{\textbf{Claude Sonnet 4.5}} \\
\cmidrule(lr){1-4}\cmidrule(lr){5-8}
No Skills & 2.40 & 2,810 & 7.5 & No Skills & 2.28 & 1,072 & 2.7 \\
Vanilla Skills & 3.25 & 135,903 & 9.2 & Vanilla Skills & 1.55 & 84,297 & 4.3 \\
Vector Skills & 1.88 & 3,712 & 6.0 & Vector Skills & 2.40 & 2,179 & 3.6 \\
Graph of Skills & 2.69 & 3,514 & 6.5 & Graph of Skills & 2.59 & 2,160 & 4.2 \\
\rowcolor{methodrow}
\goskills{} & 2.62 & 3,153 & 5.3 & \goskills{} & 2.99 & 1,976 & 2.9 \\
\midrule
\multicolumn{4}{c}{\textbf{Qwen 3.5}} & \multicolumn{4}{c}{\textbf{Kimi K2.5}} \\
\cmidrule(lr){1-4}\cmidrule(lr){5-8}
No Skills & 3.36 & 2,409 & 8.5 & No Skills & 2.62 & 1,906 & 6.4 \\
Vanilla Skills & 2.28 & 112,684 & 18.9 & Vanilla Skills & 2.83 & 90,812 & 15.0 \\
Vector Skills & 2.83 & 3,981 & 9.8 & Vector Skills & 3.11 & 3,124 & 6.9 \\
Graph of Skills & 3.88 & 3,426 & 9.4 & Graph of Skills & 3.11 & 3,436 & 7.0 \\
\rowcolor{methodrow}
\goskills{} & 3.42 & 3,483 & 8.2 & \goskills{} & 3.95 & 2,956 & 6.0 \\
\bottomrule
\end{tabular*}
\end{ApdxTabFrame}
\end{table}

\begin{table}[!t]\centering\footnotesize
\caption{Terminal-Bench 2 sample SDs across three repeats for Table~\ref{tab:main_results}. Accuracy is in pp, tokens are cumulative input tokens, and time is agent-only seconds.}
\label{tab:sd_terminalbench}
\setlength{\tabcolsep}{4pt}
\renewcommand{\arraystretch}{1.04}
\begin{ApdxTabFrame}
\begin{tabular*}{\linewidth}{@{\extracolsep{\fill}}lrrrlrrr@{}}
\toprule
\rowcolor{tablehead}
\apdxhdr{Method} & \apdxhdr{Accuracy SD} & \apdxhdr{Token SD} & \apdxhdr{Time SD} & \apdxhdr{Method} & \apdxhdr{Accuracy SD} & \apdxhdr{Token SD} & \apdxhdr{Time SD} \\
\midrule
\multicolumn{4}{c}{\textbf{GPT-5.4}} & \multicolumn{4}{c}{\textbf{Gemini 3 Pro}} \\
\cmidrule(lr){1-4}\cmidrule(lr){5-8}
No Skills & 4.90 & 8,873 & 16.8 & No Skills & 4.05 & 15,435 & 17.7 \\
Vanilla Skills & 4.05 & 62,098 & 29.9 & Vanilla Skills & 3.61 & 52,327 & 27.8 \\
Vector Skills & 5.15 & 9,030 & 20.9 & Vector Skills & 5.19 & 18,775 & 19.1 \\
Graph of Skills & 1.95 & 10,057 & 17.9 & Graph of Skills & 3.61 & 19,771 & 13.6 \\
\rowcolor{methodrow}
\goskills{} & 3.37 & 10,647 & 29.1 & \goskills{} & 4.68 & 17,042 & 23.3 \\
\midrule
\multicolumn{4}{c}{\textbf{MiniMax M2.7}} & \multicolumn{4}{c}{\textbf{Claude Sonnet 4.5}} \\
\cmidrule(lr){1-4}\cmidrule(lr){5-8}
No Skills & 3.24 & 47,169 & 34.1 & No Skills & 2.34 & 33,880 & 28.4 \\
Vanilla Skills & 5.31 & 114,465 & 63.4 & Vanilla Skills & 4.54 & 93,560 & 41.6 \\
Vector Skills & 3.95 & 49,598 & 40.0 & Vector Skills & 3.95 & 36,349 & 34.0 \\
Graph of Skills & 2.59 & 41,891 & 40.4 & Graph of Skills & 5.66 & 32,707 & 24.4 \\
\rowcolor{methodrow}
\goskills{} & 4.25 & 51,651 & 50.9 & \goskills{} & 4.25 & 32,383 & 30.8 \\
\midrule
\multicolumn{4}{c}{\textbf{Qwen 3.5}} & \multicolumn{4}{c}{\textbf{Kimi K2.5}} \\
\cmidrule(lr){1-4}\cmidrule(lr){5-8}
No Skills & 4.68 & 28,741 & 60.8 & No Skills & 3.43 & 32,436 & 33.9 \\
Vanilla Skills & 4.54 & 59,104 & 85.0 & Vanilla Skills & 5.77 & 86,070 & 47.1 \\
Vector Skills & 4.90 & 26,417 & 61.7 & Vector Skills & 2.83 & 41,653 & 40.6 \\
Graph of Skills & 5.31 & 29,271 & 57.9 & Graph of Skills & 2.59 & 35,784 & 34.3 \\
\rowcolor{methodrow}
\goskills{} & 3.43 & 36,581 & 79.6 & \goskills{} & 3.89 & 33,169 & 43.6 \\
\bottomrule
\end{tabular*}
\end{ApdxTabFrame}
\end{table}

\section{Failure and Error Analysis}
\label{app:error_analysis}

\paragraph{Error taxonomy.}
We separate failures caused by the retrieval interface from failures caused by downstream
execution or infrastructure. This distinction is important because \goskills only changes
the retrieved context. It does not execute skills, bind arguments, repair code after a
failure, or inspect hidden evaluator state. Table~\ref{tab:app_error_taxonomy} summarizes
the categories used in our failure analysis.

\begin{table}[t]
\centering
\small
\setlength{\tabcolsep}{3.5pt}
\caption{
Failure taxonomy for group-structured skill retrieval experiments. The taxonomy separates
what \goskills can directly affect from downstream and infrastructure bottlenecks.
}
\label{tab:app_error_taxonomy}
\begin{ApdxTabFrame}
\renewcommand{\arraystretch}{1.14}
\begin{tabular}{@{}L{0.21\linewidth}L{0.38\linewidth}L{0.31\linewidth}@{}}
\toprule
\rowcolor{tablehead}
\apdxhdr{Error mode} & \apdxhdr{Typical symptom} & \apdxhdr{Interpretation} \\
\midrule
Activation miss &
No activated group contains the skill or facet needed by the visible task requirement. &
Retrieval-side failure; better schema extraction, seed evidence, or group indexing can help. \\
Partial coverage &
The anchor is plausible, but the final bottleneck omits a required support, artifact, or
visible-check skill. &
Retrieval-side failure; this is the purpose of coverage debt and budgeted backfill. \\
Good retrieval, bad execution &
The exposed skills are plausible, but the agent over-builds, ignores the contract, or fails
to satisfy the task check. &
Mostly downstream; retrieval can reduce search friction but cannot guarantee execution. \\
Context overhead &
The contract or support context adds tokens without improving the trajectory. &
Efficiency failure; most likely on long setup chains that need broad environment context. \\
Infrastructure failure &
Container startup, dependency, or evaluator failures prevent a valid task outcome. &
Excluded from aggregation. Agent execution timeouts remain valid task failures
and contribute their observed token and runtime costs. \\
\bottomrule
\end{tabular}
\end{ApdxTabFrame}
\end{table}

\paragraph{Retrieval-side failures.}
In \goskills, a retrieval miss can occur before group selection if the seed evidence fails
to activate the relevant group, or after group selection if bottlenecking drops a required
support skill. The retrieval-gate comparison in Table~\ref{tab:app_retrieval_evolution}
shows this distinction empirically: the initial configuration produced misses and partial
results, while the final gate reaches 40/40 requirement-level pass in both modes. We therefore treat
coverage debt as an implementation check, not as a proof that execution will succeed.
It only records whether high-confidence visible requirements remain uncovered by the final
presented context.

\paragraph{Execution failures after valid retrieval.}
Some failures remain even when the retrieved context is plausible. In
\texttt{energy-market-pricing} (Appendix~\ref{app:qualitative_analysis}, Case~4), both runs
complete with closely matching formulations, and the \goskills run fails one optimality test
at a 0.01\% tolerance. Such outcomes should not be collapsed into infrastructure noise: a valid
start and relevant skills do not guarantee correct multi-step execution. For \goskills, the
limitation is that
group structure can make the entry point and support roles explicit, but it cannot force the
downstream model to perform the right multi-step execution.

\paragraph{Efficiency failures.}
The group contract can add useful structure, but it is not free. On both matched slices in
Table~\ref{tab:app_token_runtime}, \goskills uses more tokens on average while
reducing agent-only runtime. The task-level pattern is also mixed: among the
cases in Appendix~\ref{app:qualitative_analysis}, \goskills uses fewer input tokens on
\texttt{protein-expression-analysis}, \texttt{3d-scan-calc}, and
\texttt{energy-market-pricing} but more on \texttt{threejs-structure-parser}. This supports
the scoped interpretation used in the main paper: group-structured retrieval is most useful
when the task has a compact anchor/support/check decomposition, and less reliable when the
dominant difficulty is broad setup or environment repair.

\paragraph{Infrastructure failures.}
We retain completed episodes as substantive outcomes, including reward-0 failures. Episodes
with container-startup failures, dependency failures, or unavailable evaluator outcomes
are separated as infrastructure evidence. Agent execution timeouts are retained as
task failures, including zero-reward SkillsBench timeouts and ALFWorld step-limit failures. This policy prevents
setup and reward-artifact failures from being misread as retrieval misses while still counting
completed failed attempts as real downstream outcomes.

\section{Qualitative Analysis}
\label{app:qualitative_analysis}

\paragraph{Section framing.}
We use trajectory-grounded case studies to explain when group-structured retrieval changes
the downstream trajectory and when it does not. The relevant question is not only whether a
method retrieves a topically related skill, but whether the exposed context gives the
downstream agent an executable entry point, supporting artifacts, and visible checks early
enough to change the trajectory. This is the intended difference between Graph of Skills and
\goskills: Graph of Skills scores skill nodes through graph diffusion and hydrates a
dependency-aware bundle, whereas \goskills selects anchor-centered groups, expands support
groups, bottlenecks the selected plan to a few atomic payloads, and renders explicit
\textsc{Start}, \textsc{Support}, \textsc{Check}, \textsc{Avoid}, and debt fields.

\paragraph{Evidence sources.}
The cases use GPT-5.4 SkillsBench runs from repeat~2 of the final evaluation. For each task
we inspect the context each method presented (the retrieval bundle), the agent trajectory,
and the verifier output; all numbers in Table~\ref{tab:app_case_evidence} are taken from the
run records. We chose the cases to cover two wins, a tie, and a loss for \goskills. Among the
reward flips we inspected, we excluded those whose trajectories show that the decisive
difference was not influenced by the presented context, such as a different reading of an
ambiguous requirement or a hand-picked numerical threshold.

\begin{notebox}{Reading guide}
Each case is interpreted along three axes: whether the initial context exposes a credible
entry point, whether supporting artifacts and checks are present, and whether the remaining
failure is better attributed to retrieval, execution, or infrastructure.
\end{notebox}

\begin{table}[H]
\centering
\small
\setlength{\tabcolsep}{3.4pt}
\caption{Matched GPT-5.4 SkillsBench cases from repeat~2. Each method cell reports evaluator
reward (percent), agent turns, input tokens, and agent-only time (seconds).}
\label{tab:app_case_evidence}
\begin{ApdxTabFrame}
\renewcommand{\arraystretch}{1.18}
\begin{tabular}{@{}L{0.25\linewidth}L{0.15\linewidth}L{0.25\linewidth}L{0.25\linewidth}@{}}
\toprule
\rowcolor{tablehead}
\apdxhdr{Task} & \apdxhdr{Case type} & \apdxhdr{Graph of Skills} & \apdxhdr{\goskills} \\
\midrule
\texttt{protein-\allowbreak expression-\allowbreak analysis} & Win (mechanism) &
0\%; 16; 172,681; 102.4s & 100\%; 9; 58,427; 80.5s \\
\texttt{threejs-\allowbreak structure-\allowbreak parser} & Win (suggestive) &
0\%; 7; 45,425; 71.3s & 100\%; 7; 55,709; 85.7s \\
\texttt{3d-\allowbreak scan-\allowbreak calc} & Tie &
100\%; 8; 38,671; 69.2s & 100\%; 7; 30,258; 73.3s \\
\texttt{energy-\allowbreak market-\allowbreak pricing} & Loss &
100\%; 13; 115,680; 185.7s & 0\%; 12; 99,391; 155.3s \\
\bottomrule
\end{tabular}
\end{ApdxTabFrame}
\end{table}

\begin{notebox}{Case 1 \textperiodcentered{} \texttt{protein-expression-analysis}: the presented excerpt decides the outcome}
\textbf{Presented.} Both methods surface the same spreadsheet skills (\texttt{xlsx}, Excel
Automation, Excel Analysis). Graph of Skills lists \texttt{xlsx} first and presents the head of
its skill file: front matter and formatting conventions, with recalculation mentioned only in
the front-matter description. \goskills anchors Excel Automation, lists \texttt{xlsx} as
support, and presents the \texttt{xlsx} section that requires running \texttt{recalc.py} after
writing formulas.\par\smallskip
\textbf{Trajectory.} Both agents write INDEX/MATCH and STDEV formulas with
\texttt{openpyxl}. The \goskills agent runs
\texttt{python3 .../xlsx/recalc.py protein\_expression.xlsx 30} at step~7. The Graph of
Skills agent never calls \texttt{recalc.py}; at step~12 it converts the workbook with
LibreOffice into \texttt{/tmp}, leaving the original file without cached formula values.\par\smallskip
\textbf{Outcome.} \goskills passes all six tests. Graph of Skills fails the three computation
tests (``Only 0/100 cells filled''), using 16 turns and 172,681 input tokens, versus 9 and
58,427 for \goskills.\par\smallskip
\textbf{Reading.} The decisive difference is which part of a shared skill reaches the agent,
not the \textsc{Start} line. This follows from the source-preserving presentation policy
(Appendix~\ref{app:scoring_rules}), which keeps task-relevant sections within the payload cap.
\end{notebox}

\begin{notebox}{Case 2 \textperiodcentered{} \texttt{threejs-structure-parser}: a surfaced implementation detail}
\textbf{Presented.} Both contexts include the \texttt{threejs} and \texttt{obj-exporter}
skills. Only the \goskills context contains the export rule
\texttt{if (geom.index) geom = geom.toNonIndexed()}; it also carries two unrelated support
skills.\par\smallskip
\textbf{Trajectory.} Both agents read the same reference export script. The Graph of Skills
agent sees the \texttt{toNonIndexed} line while reading it (step~5), but its own export code
omits the call. The \goskills agent writes the call into its export script (step~6).\par\smallskip
\textbf{Outcome.} Graph of Skills fails the part- and link-mesh tests (Chamfer distance
1.493 against a threshold of 0.0002); \goskills passes.\par\smallskip
\textbf{Reading.} Suggestive rather than conclusive: we did not re-execute the export to
isolate the effect of the indexing step.
\end{notebox}

\begin{notebox}{Case 3 \textperiodcentered{} \texttt{3d-scan-calc}: same entry point, same answer}
\textbf{Presented.} Both methods lead with \texttt{mesh-analysis}. \goskills presents two
skills in 3,794 characters, and Graph of Skills presents four skills in 8,930 characters; each context
includes unrelated extras.\par\smallskip
\textbf{Trajectory.} Both agents read the skill and its mesh tool and call the same
largest-component analysis.\par\smallskip
\textbf{Outcome.} Both return the same mass (34,648.04) and material ID (42). \goskills uses
7 turns and 30,258 input tokens, versus 8 and 38,671 for Graph of Skills.\par\smallskip
\textbf{Reading.} When the baseline already exposes the right entry point, the contract
mainly trims context rather than changing the outcome.
\end{notebox}

\begin{notebox}{Case 4 \textperiodcentered{} \texttt{energy-market-pricing}: a loss at a tolerance boundary}
\textbf{Presented.} Both contexts contain overlapping power-system skills; \goskills anchors
\texttt{power-flow-data}. Its \textsc{Check} field lists only artifact facets (\texttt{network.json},
\texttt{report.json}, and JSON output), and its \textsc{Debt} field still lists \texttt{report.json}.\par\smallskip
\textbf{Trajectory.} Neither agent opens a skill file. Both write a DC optimal power flow with
reserve coupling using \texttt{linprog} with the HiGHS solver.\par\smallskip
\textbf{Outcome.} The \goskills run reports a base cost of 2,965,331.83, below the reference
optimum of 2,965,637.91 by more than the 0.01\% tolerance (296.56), and fails one optimality
test; the Graph of Skills run passes.\par\smallskip
\textbf{Reading.} The contract carries no modeling content for this task, and the outcome
flips on a small formulation difference at the tolerance boundary.
\end{notebox}

\begin{notebox}{Scope of the case evidence}
These are single-repeat examples chosen to cover two wins, a tie, and a loss. They illustrate
how the presented context can change, or fail to change, a trajectory; the paired aggregate
results in Appendix~\ref{app:exp_details} provide the broader comparison across tasks.
\end{notebox}

\section{Additional Weaknesses}
\label{app:weaknesses}

This section summarizes the main limitations of the current implementation and
evaluation protocol. These weaknesses do not change the central goal of
\goskills: studying whether group-structured retrieval can make skill contexts
more usable under a bounded prompt budget. However, they clarify where the
reported results should be interpreted with care.

\paragraph{Dependence on deterministic schema extraction.}
\goskills relies on deterministic query-schema extraction, typed skill metadata,
and visible requirement cues. This design makes the retrieval process transparent
and reproducible, but it may be less robust when task prompts are ambiguous,
metadata are sparse, or the relevant requirements are only implicit. In particular,
the method can miss support skills when a task does not expose clear technology
anchors, artifact names, output formats, or failure cues. The current evaluation
therefore best reflects settings where task-visible requirements can be extracted
reliably before execution.

\paragraph{Fixed scoring rules.}
The group, support, and bottleneck scores are fixed weighted sums over handcrafted
feature families. This makes the method easy to inspect and keeps the retrieval
policy unchanged across backbones and benchmarks. However, the coefficients may
implicitly fit the development distribution and may not be optimal for other skill
libraries or task domains. A learned or validation-tuned scoring policy could
improve transfer, but would also introduce additional training data requirements
and make it harder to isolate the effect of group-structured retrieval itself.

\section{Broader impacts}
\label{app:broader_impacts}

This work studies inference-time retrieval and contextualization for existing
agent skill libraries. Potential positive impacts include more efficient use of
large skill libraries, bounded agent-facing context, and more transparent
agent-facing context through explicit START, SUPPORT, CHECK, AVOID, and DEBT
fields. Potential negative impacts include over-reliance on retrieved guidance,
more capable automated coding agents being used for unsafe automation, and
amplification of unsafe or low-quality skills if such skills are present in the
underlying library. The method does not itself filter unsafe skills, verify
semantic correctness, or prevent misuse; deployment should therefore pair
retrieval with library curation, policy checks, and task-level monitoring.

\section{Compute resources and API usage}
\label{app:compute_resources}

All experiments are API-based agent evaluations. We do not train or fine-tune any
model and do not use GPUs. The benchmark workers run the downstream agent loop,
retrieval code, logging, and task evaluators on CPU machines. Each worker has
8 vCPUs and 32\,GB of RAM, and all experiments run on Ubuntu 22.04 cloud VMs with the same
software environment and benchmark versions across methods.

For each model--method--benchmark combination, we run three repeats. The main
tables report mean cumulative input tokens and mean agent-only task-processing runtime,
excluding environment setup. Infrastructure failures, task timeouts, and valid
task outcomes are tracked separately in the run provenance table. API-based model
inference is performed through the corresponding provider endpoints; access dates are given
in the references.

\section{Assets, licenses, and terms of use}
\label{app:licenses}

Table~\ref{tab:asset_licenses} lists the external assets used in this paper with their
source, the version used, and their license. All benchmarks, baseline code, and skill
libraries are used unmodified and only for research evaluation, in accordance with their
licenses; pinned versions and checksums are recorded in the repository's
\href{https://github.com/yuhuo03/Group-of-Skills/blob/main/DATA.md}{\texttt{DATA.md}}.
Individual \texttt{SKILL.md} files in the downloaded libraries retain the terms of their
original sources. Backbone models and the embedding model are accessed through their
providers' public APIs under the providers' terms of service, and no model weights are
redistributed.

\begin{table}[H]
\centering
\small
\setlength{\tabcolsep}{4pt}
\caption{External assets used in this paper, with source, pinned version where applicable, and license.}
\label{tab:asset_licenses}
\begin{ApdxTabFrame}
\renewcommand{\arraystretch}{1.12}
\begin{tabular}{L{0.20\linewidth}L{0.25\linewidth}L{0.29\linewidth}L{0.16\linewidth}}
\toprule
\rowcolor{tablehead}
\apdxhdr{Asset} & \apdxhdr{Role in this paper} & \apdxhdr{Source and version} & \apdxhdr{License / terms} \\
\midrule
SkillsBench~\citep{li2026skillsbench} &
Technical-task benchmark with curated skills and deterministic verifiers. &
\path{benchflow-ai/skillsbench}, commit \texttt{7c3d109} &
Apache-2.0 \\
ALFWorld~\citep{shridhar2021alfworld} (built on TextWorld) &
Multi-turn interactive benchmark. &
\path{alfworld/alfworld}, official downloader; \path{microsoft/TextWorld} &
MIT (ALFWorld and TextWorld) \\
Terminal-Bench~2~\citep{merrill2026terminal} &
Terminal and CLI agent benchmark. &
\path{harbor-framework/terminal-bench-2}, commit \texttt{69671fb} &
Apache-2.0 \\
Harbor &
Runs Terminal-Bench 2 task containers and their native verifiers; the agent loop is our own. &
\path{harbor-framework/harbor}, version 0.20.0 &
Apache-2.0 \\
Graph of Skills~\citep{li2026graph} &
Structural retrieval baseline and typed skill-graph constructor. &
\path{davidliuk/graph-of-skills}, commit \texttt{203f60a} &
MIT \\
Skill libraries (\texttt{skills\_200}--\texttt{skills\_2000}) &
Skill corpora for the retrieved-skill methods, and for Vanilla Skills on ALFWorld and Terminal-Bench 2. &
Hugging Face \path{DLPenn/graph-of-skills-data}, revision \texttt{83a8c59} &
MIT \\
Agent Skills~\citep{agentskills2026} &
Skill package format; Vanilla Skills baseline. &
\path{agentskills/agentskills} &
Apache-2.0 (code), CC-BY-4.0 (docs) \\
\path{text-embedding-3-large}~\citep{openai_text_embedding_3_large} &
Embedding model for Vector Skills. &
OpenAI API &
OpenAI terms of use \\
Backbone models &
GPT-5.4, Claude Sonnet 4.5, Gemini 3 Pro, MiniMax M2.7, Qwen 3.5, Kimi K2.5. &
Provider APIs &
Provider terms of service \\
\rowcolor{methodrow}
\goskills{} code (ours) &
Group construction, retrieval, contract rendering, and evaluation scripts. &
\path{yuhuo03/Group-of-Skills} &
MIT \\
\rowcolor{methodrow}
\goskills{} data (ours) &
Skill libraries, prebuilt workspaces, and group pools at 200--2,000 skills. &
Hugging Face \path{yuhuo03/group-of-skills-data} &
Upstream skill licenses \\
\bottomrule
\end{tabular}
\end{ApdxTabFrame}
\end{table}


\clearpage

\section*{NeurIPS Paper Checklist}

\begin{enumerate}

\item {\bf Claims}
    \item[] Question: Do the main claims made in the abstract and introduction accurately reflect the paper's contributions and scope?
    \item[] Answer: \answerYes{} 
    \item[] Justification: The abstract and introduction state that \goskills is an inference-time contextualization layer and limit the claims to the evaluated benchmarks and agent settings. We report both effectiveness and efficiency results in Section~\ref{sec:experiments}.
    \item[] Guidelines:
    \begin{itemize}
        \item The answer \answerNA{} means that the abstract and introduction do not include the claims made in the paper.
        \item The abstract and/or introduction should clearly state the claims made, including the contributions made in the paper and important assumptions and limitations. A \answerNo{} or \answerNA{} answer to this question will not be perceived well by the reviewers. 
        \item The claims made should match theoretical and experimental results, and reflect how much the results can be expected to generalize to other settings. 
        \item It is fine to include aspirational goals as motivation as long as it is clear that these goals are not attained by the paper. 
    \end{itemize}

\item {\bf Limitations}
    \item[] Question: Does the paper discuss the limitations of the work performed by the authors?
    \item[] Answer: \answerYes{} 
    \item[] Justification: We discuss the main assumptions and limitations in Section~\ref{sec:limitations}.
    \item[] Guidelines:
    \begin{itemize}
        \item The answer \answerNA{} means that the paper has no limitation while the answer \answerNo{} means that the paper has limitations, but those are not discussed in the paper. 
        \item The authors are encouraged to create a separate ``Limitations'' section in their paper.
        \item The paper should point out any strong assumptions and how robust the results are to violations of these assumptions (e.g., independence assumptions, noiseless settings, model well-specification, asymptotic approximations only holding locally). The authors should reflect on how these assumptions might be violated in practice and what the implications would be.
        \item The authors should reflect on the scope of the claims made, e.g., if the approach was only tested on a few datasets or with a few runs. In general, empirical results often depend on implicit assumptions, which should be articulated.
        \item The authors should reflect on the factors that influence the performance of the approach. For example, a facial recognition algorithm may perform poorly when image resolution is low or images are taken in low lighting. Or a speech-to-text system might not be used reliably to provide closed captions for online lectures because it fails to handle technical jargon.
        \item The authors should discuss the computational efficiency of the proposed algorithms and how they scale with dataset size.
        \item If applicable, the authors should discuss possible limitations of their approach to address problems of privacy and fairness.
        \item While the authors might fear that complete honesty about limitations might be used by reviewers as grounds for rejection, a worse outcome might be that reviewers discover limitations that aren't acknowledged in the paper. The authors should use their best judgment and recognize that individual actions in favor of transparency play an important role in developing norms that preserve the integrity of the community. Reviewers will be specifically instructed to not penalize honesty concerning limitations.
    \end{itemize}

\item {\bf Theory assumptions and proofs}
    \item[] Question: For each theoretical result, does the paper provide the full set of assumptions and a complete (and correct) proof?
    \item[] Answer: \answerNA{} 
    \item[] Justification: The paper does not present formal theoretical results or proofs; the proposed scoring functions and algorithms are used as empirical design components.
    \item[] Guidelines:
    \begin{itemize}
        \item The answer \answerNA{} means that the paper does not include theoretical results. 
        \item All the theorems, formulas, and proofs in the paper should be numbered and cross-referenced.
        \item All assumptions should be clearly stated or referenced in the statement of any theorems.
        \item The proofs can either appear in the main paper or the supplemental material, but if they appear in the supplemental material, the authors are encouraged to provide a short proof sketch to provide intuition. 
        \item Inversely, any informal proof provided in the core of the paper should be complemented by formal proofs provided in appendix or supplemental material.
        \item Theorems and Lemmas that the proof relies upon should be properly referenced. 
    \end{itemize}

    \item {\bf Experimental result reproducibility}
    \item[] Question: Does the paper fully disclose all the information needed to reproduce the main experimental results of the paper to the extent that it affects the main claims and/or conclusions of the paper (regardless of whether the code and data are provided or not)?
    \item[] Answer: \answerYes{} 
    \item[] Justification: Section~\ref{sec:experiments} describes the benchmarks, baselines, evaluation metrics, model backbones, skill budgets, and run protocol used to reproduce the main results.
    \item[] Guidelines:
    \begin{itemize}
        \item The answer \answerNA{} means that the paper does not include experiments.
        \item If the paper includes experiments, a \answerNo{} answer to this question will not be perceived well by the reviewers: Making the paper reproducible is important, regardless of whether the code and data are provided or not.
        \item If the contribution is a dataset and\slash or model, the authors should describe the steps taken to make their results reproducible or verifiable. 
        \item Depending on the contribution, reproducibility can be accomplished in various ways. For example, if the contribution is a novel architecture, describing the architecture fully might suffice, or if the contribution is a specific model and empirical evaluation, it may be necessary to either make it possible for others to replicate the model with the same dataset, or provide access to the model. In general. releasing code and data is often one good way to accomplish this, but reproducibility can also be provided via detailed instructions for how to replicate the results, access to a hosted model (e.g., in the case of a large language model), releasing of a model checkpoint, or other means that are appropriate to the research performed.
        \item While NeurIPS does not require releasing code, the conference does require all submissions to provide some reasonable avenue for reproducibility, which may depend on the nature of the contribution. For example
        \begin{enumerate}
            \item If the contribution is primarily a new algorithm, the paper should make it clear how to reproduce that algorithm.
            \item If the contribution is primarily a new model architecture, the paper should describe the architecture clearly and fully.
            \item If the contribution is a new model (e.g., a large language model), then there should either be a way to access this model for reproducing the results or a way to reproduce the model (e.g., with an open-source dataset or instructions for how to construct the dataset).
            \item We recognize that reproducibility may be tricky in some cases, in which case authors are welcome to describe the particular way they provide for reproducibility. In the case of closed-source models, it may be that access to the model is limited in some way (e.g., to registered users), but it should be possible for other researchers to have some path to reproducing or verifying the results.
        \end{enumerate}
    \end{itemize}

\item {\bf Open access to data and code}
    \item[] Question: Does the paper provide open access to the data and code, with sufficient instructions to faithfully reproduce the main experimental results, as described in supplemental material?
    \item[] Answer: \answerYes{} 
    \item[] Justification: The public repository linked below the author list provides retrieval code, baseline runners, configuration examples, and evaluation instructions. Appendix~\ref{app:benchmark_protocol} specifies the reported task sets and baseline conditions. The release's default task manifest and skill-directory configuration differ from some reported-run settings and require alignment for reproduction.
    \item[] Guidelines:
    \begin{itemize}
        \item The answer \answerNA{} means that paper does not include experiments requiring code.
        \item Please see the NeurIPS code and data submission guidelines (\url{https://neurips.cc/public/guides/CodeSubmissionPolicy}) for more details.
        \item While we encourage the release of code and data, we understand that this might not be possible, so \answerNo{} is an acceptable answer. Papers cannot be rejected simply for not including code, unless this is central to the contribution (e.g., for a new open-source benchmark).
        \item The instructions should contain the exact command and environment needed to run to reproduce the results. See the NeurIPS code and data submission guidelines (\url{https://neurips.cc/public/guides/CodeSubmissionPolicy}) for more details.
        \item The authors should provide instructions on data access and preparation, including how to access the raw data, preprocessed data, intermediate data, and generated data, etc.
        \item The authors should provide scripts to reproduce all experimental results for the new proposed method and baselines. If only a subset of experiments are reproducible, they should state which ones are omitted from the script and why.
        \item At submission time, to preserve anonymity, the authors should release anonymized versions (if applicable).
        \item Providing as much information as possible in supplemental material (appended to the paper) is recommended, but including URLs to data and code is permitted.
    \end{itemize}

\item {\bf Experimental setting/details}
    \item[] Question: Does the paper specify all the training and test details (e.g., data splits, hyperparameters, how they were chosen, type of optimizer) necessary to understand the results?
    \item[] Answer: \answerYes{} 
    \item[] Justification: We specify the evaluation benchmarks, baselines, inference settings, retrieval budgets, scoring components, and metrics in Section~\ref{sec:experiments} and provide additional details in the appendix.
    \item[] Guidelines:
    \begin{itemize}
        \item The answer \answerNA{} means that the paper does not include experiments.
        \item The experimental setting should be presented in the core of the paper to a level of detail that is necessary to appreciate the results and make sense of them.
        \item The full details can be provided either with the code, in appendix, or as supplemental material.
    \end{itemize}

\item {\bf Experiment statistical significance}
    \item[] Question: Does the paper report error bars suitably and correctly defined or other appropriate information about the statistical significance of the experiments?
    \item[] Answer: \answerYes{} 
    \item[] Justification: Appendix~\ref{app:exp_details} reports task-level paired bootstrap
confidence intervals for the main GPT-5.4 SkillsBench comparisons. We resample paired
tasks with replacement for 10{,}000 bootstrap samples and report percentile
95\% confidence intervals for reward and runtime deltas, and the robustness study
reports standard deviations over corruption seeds. Appendix~\ref{app:repeat_sd}
reports sample standard deviations across three repeats for every configuration
in the main results and control/ablation tables, separately from task-resampling
and corruption-seed variability.
    \item[] Guidelines:
    \begin{itemize}
        \item The answer \answerNA{} means that the paper does not include experiments.
        \item The authors should answer \answerYes{} if the results are accompanied by error bars, confidence intervals, or statistical significance tests, at least for the experiments that support the main claims of the paper.
        \item The factors of variability that the error bars are capturing should be clearly stated (for example, train/test split, initialization, random drawing of some parameter, or overall run with given experimental conditions).
        \item The method for calculating the error bars should be explained (closed form formula, call to a library function, bootstrap, etc.)
        \item The assumptions made should be given (e.g., Normally distributed errors).
        \item It should be clear whether the error bar is the standard deviation or the standard error of the mean.
        \item It is OK to report 1-sigma error bars, but one should state it. The authors should preferably report a 2-sigma error bar than state that they have a 96\% CI, if the hypothesis of Normality of errors is not verified.
        \item For asymmetric distributions, the authors should be careful not to show in tables or figures symmetric error bars that would yield results that are out of range (e.g., negative error rates).
        \item If error bars are reported in tables or plots, the authors should explain in the text how they were calculated and reference the corresponding figures or tables in the text.
    \end{itemize}

\item {\bf Experiments compute resources}
    \item[] Question: For each experiment, does the paper provide sufficient information on the computer resources (type of compute workers, memory, time of execution) needed to reproduce the experiments?
    \item[] Answer: \answerYes{} 
    \item[] Justification:  Appendix~\ref{app:compute_resources} describes the compute
resources used for the experiments. Our experiments are API-based agent evaluations and do
not train models or use GPUs. We report the CPU-based execution workers, memory,
execution environment, number of repeated runs, API model backbones, token usage, and
agent-only runtime. The main tables report mean cumulative input tokens and agent-only task-processing
time, while the run logs track task-level execution status and runtime.
    \item[] Guidelines:
    \begin{itemize}
        \item The answer \answerNA{} means that the paper does not include experiments.
        \item The paper should indicate the type of compute workers CPU or GPU, internal cluster, or cloud provider, including relevant memory and storage.
        \item The paper should provide the amount of compute required for each of the individual experimental runs as well as estimate the total compute. 
        \item The paper should disclose whether the full research project required more compute than the experiments reported in the paper (e.g., preliminary or failed experiments that didn't make it into the paper). 
    \end{itemize}
    
\item {\bf Code of ethics}
    \item[] Question: Does the research conducted in the paper conform, in every respect, with the NeurIPS Code of Ethics \url{https://neurips.cc/public/EthicsGuidelines}?
    \item[] Answer: \answerYes{} 
    \item[] Justification: We have reviewed the NeurIPS Code of Ethics and believe the research conforms to it. The work
evaluates an inference-time skill-retrieval interface on benchmark tasks, does not involve human
subjects or private data, and does not modify or deploy a system for consequential decisions; its
limitations and scope are explicitly stated.
    \item[] Guidelines:
    \begin{itemize}
        \item The answer \answerNA{} means that the authors have not reviewed the NeurIPS Code of Ethics.
        \item If the authors answer \answerNo, they should explain the special circumstances that require a deviation from the Code of Ethics.
        \item The authors should make sure to preserve anonymity (e.g., if there is a special consideration due to laws or regulations in their jurisdiction).
    \end{itemize}

\item {\bf Broader impacts}
    \item[] Question: Does the paper discuss both potential positive societal impacts and negative societal impacts of the work performed?
    \item[] Answer: \answerYes{} 
    \item[] Justification: We discuss broader impacts in Appendix~\ref{app:broader_impacts}.
Potential positive impacts include more efficient use of large skill libraries,
bounded agent-facing context, and more transparent agent-facing retrieval through explicit
role-labeled contracts. Potential negative impacts include over-reliance on retrieved
guidance, amplification of unsafe or low-quality skills already present in the library,
and misuse in automated coding-agent workflows. We therefore recommend pairing the
method with library curation, policy checks, and task-level monitoring.
    \item[] Guidelines:
    \begin{itemize}
        \item The answer \answerNA{} means that there is no societal impact of the work performed.
        \item If the authors answer \answerNA{} or \answerNo, they should explain why their work has no societal impact or why the paper does not address societal impact.
        \item Examples of negative societal impacts include potential malicious or unintended uses (e.g., disinformation, generating fake profiles, surveillance), fairness considerations (e.g., deployment of technologies that could make decisions that unfairly impact specific groups), privacy considerations, and security considerations.
        \item The conference expects that many papers will be foundational research and not tied to particular applications, let alone deployments. However, if there is a direct path to any negative applications, the authors should point it out. For example, it is legitimate to point out that an improvement in the quality of generative models could be used to generate Deepfakes for disinformation. On the other hand, it is not needed to point out that a generic algorithm for optimizing neural networks could enable people to train models that generate Deepfakes faster.
        \item The authors should consider possible harms that could arise when the technology is being used as intended and functioning correctly, harms that could arise when the technology is being used as intended but gives incorrect results, and harms following from (intentional or unintentional) misuse of the technology.
        \item If there are negative societal impacts, the authors could also discuss possible mitigation strategies (e.g., gated release of models, providing defenses in addition to attacks, mechanisms for monitoring misuse, mechanisms to monitor how a system learns from feedback over time, improving the efficiency and accessibility of ML).
    \end{itemize}
    
\item {\bf Safeguards}
    \item[] Question: Does the paper describe safeguards that have been put in place for responsible release of data or models that have a high risk for misuse (e.g., pre-trained language models, image generators, or scraped datasets)?
    \item[] Answer: \answerNA{} 
    \item[] Justification: The paper does not release high-risk pretrained models, scraped datasets, or generative systems requiring controlled access safeguards.
    \item[] Guidelines:
    \begin{itemize}
        \item The answer \answerNA{} means that the paper poses no such risks.
        \item Released models that have a high risk for misuse or dual-use should be released with necessary safeguards to allow for controlled use of the model, for example by requiring that users adhere to usage guidelines or restrictions to access the model or implementing safety filters. 
        \item Datasets that have been scraped from the Internet could pose safety risks. The authors should describe how they avoided releasing unsafe images.
        \item We recognize that providing effective safeguards is challenging, and many papers do not require this, but we encourage authors to take this into account and make a best faith effort.
    \end{itemize}

\item {\bf Licenses for existing assets}
    \item[] Question: Are the creators or original owners of assets (e.g., code, data, models), used in the paper, properly credited and are the license and terms of use explicitly mentioned and properly respected?
    \item[] Answer: \answerYes{} 
    \item[] Justification: All benchmarks, harnesses, baseline code, skill libraries, and model APIs are cited, and Appendix~\ref{app:licenses} (Table~\ref{tab:asset_licenses}) lists each with its source, version, and license; pinned versions and checksums are in the repository's \href{https://github.com/yuhuo03/Group-of-Skills/blob/main/DATA.md}{DATA.md}.
    \item[] Guidelines:
    \begin{itemize}
        \item The answer \answerNA{} means that the paper does not use existing assets.
        \item The authors should cite the original paper that produced the code package or dataset.
        \item The authors should state which version of the asset is used and, if possible, include a URL.
        \item The name of the license (e.g., CC-BY 4.0) should be included for each asset.
        \item For scraped data from a particular source (e.g., website), the copyright and terms of service of that source should be provided.
        \item If assets are released, the license, copyright information, and terms of use in the package should be provided. For popular datasets, \url{paperswithcode.com/datasets} has curated licenses for some datasets. Their licensing guide can help determine the license of a dataset.
        \item For existing datasets that are re-packaged, both the original license and the license of the derived asset (if it has changed) should be provided.
        \item If this information is not available online, the authors are encouraged to reach out to the asset's creators.
    \end{itemize}

\item {\bf New assets}
    \item[] Question: Are new assets introduced in the paper well documented and is the documentation provided alongside the assets?
    \item[] Answer: \answerYes{} 
    \item[] Justification: The \goskills{} repository provides installation and evaluation instructions in its \href{https://github.com/yuhuo03/Group-of-Skills/blob/main/README.md}{README}. The \goskills{} code is released under the \href{https://github.com/yuhuo03/Group-of-Skills/blob/main/LICENSE}{MIT License}, and the skill libraries, workspaces, and group pools are released as the \href{https://huggingface.co/datasets/yuhuo03/group-of-skills-data}{group-of-skills-data} dataset (licenses in Table~\ref{tab:asset_licenses}).
    \item[] Guidelines:
    \begin{itemize}
        \item The answer \answerNA{} means that the paper does not release new assets.
        \item Researchers should communicate the details of the dataset\slash code\slash model as part of their submissions via structured templates. This includes details about training, license, limitations, etc. 
        \item The paper should discuss whether and how consent was obtained from people whose asset is used.
        \item At submission time, remember to anonymize your assets (if applicable). You can either create an anonymized URL or include an anonymized zip file.
    \end{itemize}

\item {\bf Crowdsourcing and research with human subjects}
    \item[] Question: For crowdsourcing experiments and research with human subjects, does the paper include the full text of instructions given to participants and screenshots, if applicable, as well as details about compensation (if any)? 
    \item[] Answer: \answerNA{} 
    \item[] Justification: The paper does not involve crowdsourcing experiments or research with human subjects.
    \item[] Guidelines:
    \begin{itemize}
        \item The answer \answerNA{} means that the paper does not involve crowdsourcing nor research with human subjects.
        \item Including this information in the supplemental material is fine, but if the main contribution of the paper involves human subjects, then as much detail as possible should be included in the main paper. 
        \item According to the NeurIPS Code of Ethics, workers involved in data collection, curation, or other labor should be paid at least the minimum wage in the country of the data collector. 
    \end{itemize}

\item {\bf Institutional review board (IRB) approvals or equivalent for research with human subjects}
    \item[] Question: Does the paper describe potential risks incurred by study participants, whether such risks were disclosed to the subjects, and whether Institutional Review Board (IRB) approvals (or an equivalent approval/review based on the requirements of your country or institution) were obtained?
    \item[] Answer: \answerNA{} 
    \item[] Justification: The paper does not involve human subjects research and therefore does not require IRB approval.
    \item[] Guidelines:
    \begin{itemize}
        \item The answer \answerNA{} means that the paper does not involve crowdsourcing nor research with human subjects.
        \item Depending on the country in which research is conducted, IRB approval (or equivalent) may be required for any human subjects research. If you obtained IRB approval, you should clearly state this in the paper. 
        \item We recognize that the procedures for this may vary significantly between institutions and locations, and we expect authors to adhere to the NeurIPS Code of Ethics and the guidelines for their institution. 
        \item For initial submissions, do not include any information that would break anonymity (if applicable), such as the institution conducting the review.
    \end{itemize}

\item {\bf Declaration of LLM usage}
    \item[] Question: Does the paper describe the usage of LLMs if it is an important, original, or non-standard component of the core methods in this research? Note that if the LLM is used only for writing, editing, or formatting purposes and does \emph{not} impact the core methodology, scientific rigor, or originality of the research, declaration is not required.
    \item[] Answer: \answerYes{} 
    \item[] Justification: LLM-based coding agents are part of the evaluated system, and the PH-LLM control uses an LLM-based role annotator. We describe the model backbones and inference protocol in Section~\ref{sec:experiments}, the controls in Appendix~\ref{app:expanded}, and the agent-facing prompts in Appendix~\ref{app:agent_prompts}.
    \item[] Guidelines:
    \begin{itemize}
        \item The answer \answerNA{} means that the core method development in this research does not involve LLMs as any important, original, or non-standard components.
        \item Please refer to our LLM policy in the NeurIPS handbook for what should or should not be described.
    \end{itemize}

\end{enumerate}

\end{document}